%% file: aamas_camera_ready.tex
\documentclass[sigconf]{aamas}

\usepackage{balance} 

\newcommand{\BLO}{BLO}

\usepackage{packages}
\usepackage{algorithm}
\usepackage{algorithmic}
\usepackage{microtype}
\usepackage{graphicx}
\usepackage{booktabs} 
\usepackage{camera_commands}
\usepackage{thm-restate}
\usepackage{multirow}
\usepackage{array}

\usepackage{hyperref}
\usepackage{cleveref}
\addtocontents{toc}{\protect\setcounter{tocdepth}{-10}}

\theoremstyle{plain}
\newtheorem{theorem}{Theorem}[section]

\newtheorem{lemma}[theorem]{Lemma}
\newtheorem{corollary}[theorem]{Corollary}
\theoremstyle{definition}
\newtheorem{definition}[theorem]{Definition}
\newtheorem{assumption}[theorem]{Assumption}
\theoremstyle{remark}
\newtheorem{remark}[theorem]{Remark}

\doi{NETG3616}

\makeatletter
\gdef\@copyrightpermission{
  \begin{minipage}{0.2\columnwidth}
   \href{https://creativecommons.org/licenses/by/4.0/}{\includegraphics[width=0.90\textwidth]{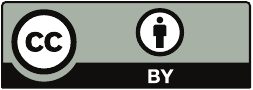}}
  \end{minipage}\hfill
  \begin{minipage}{0.8\columnwidth}
   \href{https://creativecommons.org/licenses/by/4.0/}{This work is licensed under a Creative Commons Attribution International 4.0 License.}
  \end{minipage}
  \vspace{5pt}
}
\makeatother

\setcopyright{ifaamas}
\acmConference[AAMAS '26]{Proc.\@ of the 25th International Conference
on Autonomous Agents and Multiagent Systems (AAMAS 2026)}{May 25 -- 29, 2026}
{Paphos, Cyprus}{C.~Amato, L.~Dennis, V.~Mascardi, J.~Thangarajah (eds.)}
\copyrightyear{2026}
\acmYear{2026}
\acmDOI{}
\acmPrice{}
\acmISBN{}

\acmSubmissionID{234}

\title[AAMAS-2026 Formatting Instructions]{Bilevel Policy Optimization with Nystr\"om Hypergradients}

\author{Arjun Prakash}
\authornote{Equal contribution.}
\affiliation{
  \institution{Brown University}
  \city{Providence, RI}
  \country{USA}}
\email{arjun\_prakash@brown.edu}

\author{Naicheng He}
\authornotemark[1]
\affiliation{
  \institution{Brown University}
  \city{Providence, RI}
  \country{USA}}
\email{naicheng\_he@brown.edu}

\author{Denizalp Goktas}
\affiliation{
  \institution{Simulacrum}
  \city{New York City, NY}
  \country{USA}}
\email{deni@smlcrm.com}

\author{Jacob Makar-Limanov}
\affiliation{
  \institution{Independent Researcher}
  \country{Boston, USA}}
\email{jacobmakarlimanov@gmail.com}

\author{Amy Greenwald}
\affiliation{
  \institution{Brown University}
  \city{Providence, RI}
  \country{USA}}
\email{amy\_greenwald@brown.edu}

\begin{abstract}
The dependency of the actor on the critic in actor-critic (AC) reinforcement learning means that AC can be characterized as a bilevel optimization (BLO) problem, also called a Stackelberg game. 
This characterization motivates two modifications to vanilla AC algorithms. 
First, the critic's update should be nested to learn a best response to the actor's policy. 
Second, the actor should update according to a hypergradient that accounts for changes in the critic. 
Computing this hypergradient involves finding an inverse Hessian vector product, a process that can be numerically unstable.
We thus propose a new algorithm, Bilevel Policy Optimization with Nystr\"om Hypergradients (\BLPO), which uses nesting to account for the nested structure of BLO, and leverages the Nyström method to compute the hypergradient.
Theoretically, we prove \BLPO{} converges to (a point that satisfies the necessary conditions for) a local strong Stackelberg equilibrium in polynomial time with high probability, 
assuming a linear parametrization of the critic's objective.
Empirically, we demonstrate that \BLPO{} performs on par with or better than PPO on a variety of discrete and continuous control tasks. 
\end{abstract}


\newcommand{\BibTeX}{\rm B\kern-.05em{\sc i\kern-.025em b}\kern-.08em\TeX}

\begin{document}


\pagestyle{fancy}
\fancyhead{}


\maketitle 


\input{sections/intro}
\input{sections/prelim}
\input{sections/nystrom}

\input{sections/conv}
\input{sections/RL}
\input{sections/experiments}

\input{sections/conc}

\input{sections/acknowledge}
\balance
\bibliography{references}
\bibliographystyle{ACM-Reference-Format} 

\ifcamera
\else
    \newpage

\appendix
\onecolumn
\input{appendix/content}
\input{appendix/Figures_and_Tables}
\input{appendix/Experimental_Details}
\input{appendix/additional_experiments}
\input{appendix/proofs}
\input{appendix/nystrom_details}
\input{appendix/pseudocode}
\input{appendix/toy}
\fi





\end{document}

%% file: sections/intro.tex
\section{Introduction}
\label{intro}

Bilevel optimization (BLO) is a class of hierarchical optimization problems with two objectives, an \emph{outer} one and an \emph{inner} one.
Crucially, the objective and the variables of the outer problem depend on the solution to the inner problem.
An active area of research, bilevel optimization has applications in myriad areas of machine learning and beyond. 
Examples include  reinforcement learning \cite{ChakrabortyBKWM24,zheng2022stackelberg}, hyperparameter optimization \cite{lorraine2020optimizing}, meta-learning \cite{rajeswaran2019meta}, energy markets \cite{afcsar2016energy}, agriculture \cite{bard1998determining}, and security \cite{sinha2018stackelberg}, to name a few.
Given functions $\outerobj: \R^{\outerdim} \times \R^{\innerdim} \to \R$ and $\innerobj: \R^{\innerdim} \to \R$, a(n unconstrained) bilevel opmization problem can be formulated as follows: 
%
\begin{equation}
    \min_{\outer \in \R^{\outerdim}} \hyperobj (\outer) \doteq \outerobj (\outer, \inner[][*] (\outer)) 
     \quad \text{s.t.} \quad
    \inner[][*] (\outer) \in \innerset[\outer][*] \doteq \argmin_{\inner \in \R^{\innerdim}} \innerobj[][\outer] (\inner).
    \label{eq:BLO} 
\end{equation} 


A solution to a \BLO{} comprises a pair $(\outer[][*], \inner[][*]) \in (\R^{\outerdim}, \R^{\innerdim})$ s.t.\ $\outer$ optimizes $\hyperobj (\outer)$ subject to the constraint that $\inner[][*]$ optimizes $\innerobj[][\outer] (\inner)$.
The notation $\innerobj[][\outer]$ 
allows the optimal value of $\innerobj$ (and the corresponding solution set $\innerset[\outer][*]$) to vary with $\outer$. 
In this paper, we assume $\outerobj$ and $\innerobj$ are differentiable functions, so they can be represented by neural networks, with $\outer$ and $\inner$ as the network parameters.

Bi-level optimization problems are sometimes referred to as two-player general-sum Stackelberg games \cite{dempe2018bilevel}, with the outer (respectively, inner) player as the Stackelberg leader (respectively, follower), in which case solutions are called Stackelberg equilibria (SE).
When the inner objective is strongly convex, solutions are guaranteed to exist under fairly general conditions \citep[Theorem 5.1]{dempe2018bilevel} and are called strong Stackelberg equilibria (SSE).
In such cases, a local solution to a BLO (i.e., a local SSE) is a point $\outer[][*]$ s.t.\ $\outerobj (\outer[][*])$ is a local minimum and $\inner[][*](\outer[][*])$ is a global minimum of $\innerobj[][\outereqbm] (\inner[][*])$.

Just as many optimization problems can be solved via gradient descent, BLOs, which comprise an outer optimization and an inner one, are naturally solved via \emph{nested\/} gradient descent. 
For each outer value $\outer$, the inner optimization is solved (e.g., by gradient descent) to find $\inner[][*] (\outer)$, which is then used to update $\outer$ by following the \emph{hypergradient\/} of $\outerobj$ at $\inner[][*] (\outer)$. 
The hypergradient comprises two components: the direct gradient, which captures the direct dependence of $\outerobj$ on $\outer$, and the implicit gradient, also called the best-response Jacobian, which accounts for how changes in $\outer$ affect $\inner[][*]$.

It is straightforward to calculate the direct gradient via auto-differentiation.
The implicit gradient, however, is more difficult to compute.
There are two common approaches \cite{lorraine2020optimizing}.
The first, called unrolling gradients, makes use of a modern auto-differentiation library like PyTorch \cite{paszke2019pytorch} or Jax \cite{jax2018github} to differentiate through 
the inner optimization algorithm, rather than through an optimal solution.
This method, however, has been shown to be empirically unstable \cite{scieur2022curse}.
Moreover, it has large memory requirements, as it requires storing a copy of $\inner$ at each gradient descent step en route to computing $\inner[][*]$.


The other popular approach is to leverage the implicit function theorem (IFT) to compute the implicit gradient 
\cite{lorraine2020optimizing}.
At face value, this method requires inverting the Hessian of $\innerobj[][\outer]$, which can be computationally intractable for large neural networks and numerically unstable for ill-conditioned Hessians. 
We tackle precisely this challenge in this paper, leveraging the Nystr\"om method \cite{drineas2005Nystrom} to find a low-rank approximation of the inverse Hessian vector product (\ihvp{}), which avoids materializing the full Hessian and thus can be empirically more stable than other methods \cite{hataya2023Nystrom}.

Our contribution is: we posit actor-critic, a classic reinforcement learning architecture~\cite{barto_AC}, as a BLO, for which we propose a nested gradient desent approach that computes hypergradients using Nystr\"om's method.
We call our reinforcement learning algorithm \emph{Bilevel Policy Optimization (\BLPO)}.
We show empirically that BLPO outperforms PPO \cite{ppo} on a variety of standard discrete and continuous control tasks, without imposing significant additional computational burden.
We also prove that \BLPO{} converges in polynomial time to a point satisfying the necessary conditions of a local SSE assuming a linear parameterization of the critic.%
\footnote{Supplementary material for this paper be found \color{magenta} \href{https://arxiv.org/abs/2505.11714}{here}.}

\subsection{Actor-Critic Methods}

A reinforcement learning (RL) agent learns to make decisions by interacting with its environment sequentially (i.e., choosing actions at each state it encounters), receiving rewards along the way, with the goal of maximizing its expected return, or long-term cumulative rewards \cite{sutton2018reinforcement}.
RL algorithms generally fall into two categories: policy-based methods, which directly optimize the parameters $\actorparams$ of a parameterized policy $\policy[\actorparams]$ that maps states to actions; and value-based methods, which first estimate the expected return of a state by a parameterized value function $\Vfunc[\criticparams]$, and then infer an optimal policy by selecting reward-maximizing actions at all states. 
Actor-critic (AC) algorithms combine these approaches by learning both a parameterized policy (the actor) and value function (the critic).
Many state-of-the-art RL algorithms are built on AC-like structures, such as Trust Region Policy Optimization \cite{trpo}, Proximal Policy Optimization \cite{ppo} and Deep Deterministic Policy Gradient \cite{ddpg}.

Formally, a discrete-time Markov decision process (MDP) is defined by the tuple $\mdp \doteq \langle \States, \Actions, \transition, \rewardfunc, \discount, \statedist \rangle$.
The letter $\States$ denotes a (possibly continuous) state space, and $\Actions$, a (possibly continuous) action space.
The initial state $\initstate$ is drawn from initial state distribution $\initstatedist$, a probability density over $\States$.
The transition dynamics $\transition: \States \times \Actions \times \States \to [0, 1]$ are specified by a conditional probability density $\transition \left[ \staterv[\timestep+1] \mid \state[\timestep], \action[\timestep] \right]$, which describes transitions to the next state $\staterv[\timestep + 1]$ from the current state $\state[\timestep]$ after taking action $\action[\timestep]$.
The function $\rewardfunc: \States \times \Actions \to \R$
specifies the reward for taking action $\action[\timestep]$ in state $\state[\timestep]$.
The return $\return[\traj] \doteq\sum_{\timestep=0}^{\infty} \discount^{\timestep} \reward (\state[\timestep], \action[\timestep])$ along trajectory $\traj = (\state[0], \action[0], \state[1], \action[1], \ldots)$ is defined as the discounted sum of the cumulative rewards, where $\discount \in [0,1]$ is the discount factor.

Given an MDP $\mdp$, a policy $\policy: \States \to \probset (\Actions)$ is a mapping from states to probability distributions over actions.
The initial state distribution $\initstatedist$, the transition dynamics, and a policy induce a discounted history distribution $\histdist[\policy][\initstatedist]$ over trajectories and a discounted occupancy distribution $\statedist[\policy][\initstatedist]$ over states.
When $\initstatedist$ is a Dirac delta function with its impulse defined at $\state$, then abusing notation, we write $\histdist[\policy][{\state}]$ and $\statedist[\policy][\state]$.

Given a policy $\policy[\actorparams]$, the value function $\Vfunc[\criticparams][\policy]: \States \to \R$ at a state $\state$ is defined as the expected return over trajectories originating at $\state$ under policy $\policy$: i.e.,
$\valfun[\criticparams][\policy] (\state) = 
\Ex_{\traj \sim \histdist[\policy][\state]} \left[ \return[\traj] \right]$.
%
%
Given a value function $\Vfunc[\criticparams]$, the actor aims to maximize the expected return $\actorobj (\actorparams, \criticparams)$ of the policy $\policy[\actorparams]$, i.e., 
$\actorobj (\actorparams, \criticparams) \doteq$
$\Ex_{\state \sim {\statedist[{\policy[\actorparams]}][\initstatedist]}} \left[ \Vfunc[\criticparams][{\policy[\actorparams]}] (\state) \right]$,
%
while the critic chooses parameters $\criticparams$ 
so as to minimize the (typically, squared) error in its value function representation, given the actor's policy $\policy[\actorparams]$, i.e., 
%
\begin{align}
\criticobj (\actorparams, \criticparams)
&\doteq \frac{1}{2} \Ex_{\state \sim \statedist[{\policy[\actorparams]}][\initstatedist]} \left[ \left( \Vfunc[][{\policy[\actorparams]}] (\state) - \Vfunc[\criticparams] (\state) \right)^2 \right].
\label{eq:critic-obj}
\end{align}
The actor's objective thus depends on the critic's value function, while the critic's objective depends on the actor's policy.

Celebrated AC algorithms like PPO \cite{ppo} and
SAC \cite{haarnoja2018soft} 
update the actor and critic simultaneously, meaning each updates the parameters of its network during iteration $t+1$, given the other's parameters at iteration $t$.
Simultaneous updating corresponds to a mutual better-response dynamic, which, in the event of convergence,
would ideally find a solution to the following simultaneous-move game:
\begin{equation}
    \argmin_{\actorparams \in \R^{\outerdim}} -\actorobj (\actorparams, \criticparams)
    \quad \quad \quad 
    \argmin_{\criticparams \in \R^{\innerdim}} \criticobj (\criticparams, \actorparams).
    \label{eq:AC-Nash}
\end{equation}

We make a different modeling choice, partially
inspired by \cite{zheng2022stackelberg}, which is to define the critic's loss function as a \emph{parameterized\/} function of the actor's policy.
That is, we take $\criticobj[][\actorparams] (\criticparams) \doteq \criticobj (\actorparams, \criticparams)$, and model the problem as the following BLO:
\begin{equation}
    \min_{\actorparams \in \R^{\outerdim}} \hyperobj (\actorparams) \doteq - \actorobj (\actorparams, \criticparams[][*] (\actorparams))
    \quad
    \text{s.t.} \quad \criticparams[][*] (\actorparams) \in \arg \min_{\criticparams \in \R^{\innerdim}} \criticobj[][\actorparams] (\criticparams).
    \label{eq:AC-BLO}
\end{equation}
This characterization (AC-BLO) of the AC framework highlights its asymmetric nature, emphasizing the actor's policy search as the primary objective, with the critic's value function representation as a secondary objective, whose \emph{raison d'\^etre\/} is merely to aid the actor in its search. 
Indeed, poor estimates of the value function are known to produce less-than-ideal policies \cite{TD3}.

Since the critic's objective is a squared loss, a linear representation of the value function renders this objective strongly convex, ensuring a unique solution to the critic's optimization problem.
The actor's optimization problem, however, is in general non-convex. 
Thus, assuming a linear parameterization for the critic, the solution to our AC-BLO is a local SSE, the necessary conditions of we prove can be approximated in polynomial time, with high probability.



\subsection{Contributions}
We propose a new policy-gradient based RL algorithm, Bilevel Policy Optimization with Nystr\"om Hypergradient (\blpo), inspired by our AC-BLO formulation, in which the actor (definitively%
\footnote{In past work, the actor has also been nested~\cite{ChakrabortyBKWM24}, at least some of the time~\cite{zheng2022stackelberg}.}) plays the role of the Stackelberg leader, while the critic plays the role of the follower.
The nested structure of AC-BLO motivates a nested approach to solving it; specifically, it motivates iterating between taking one step along the hypergradient of the actor's objective, followed by many steps along the gradient of the critic's.

The aforementioned intuitions are not entirely novel, as others before us have nested the critic's computation.
In practice, however, past AC-BLO formulations have not born fruit~\cite{zheng2022stackelberg}.
Our empirical studies reveal that the culprit is the instability of the requisite \ihvp{} computations.
In particular, iterative methods, such as conjugate gradient (CG)~\cite{hestenes1952methods}, are unstable because even regularized Hessians are poorly conditioned, and thus difficult to invert. 
As a result, we compute a low-rank approximation of the \ihvp{} via Nystr\"om's method.
This modification characterizes our new algorithm, \blpo.


Empirically, we show that \blpo{} outperforms PPO in a variety of discrete and continuous control tasks.
We also present a series of ablations to confirm that both nesting (with the actor as leader, and the critic as follower) and the Nystr\"om method are essential for these performance gains. 
Related, we show that using CG to approximate the \ihvp{} can lead to severe performance degradation. 

Under a linear parameterization of the critic's value function, the inner optimization of our proposed AC-BLO formulation is strongly convex.
Under this assumption, we prove that \BLPO{} converges in polynomial time to a point that satisfies the necessary conditions of local SSE, with high probability.
This latter caveat is necessary, as Nystr\"om's method is randomized \cite{drineas2005Nystrom}.
The rate we derive suggests a faster learning rate for the critic
($O \left( \nicefrac{1}{\condnum} \right)$), 
and a much slower one for the actor ($O \left( \nicefrac{1}{\condnum^3} \right)$), where $\condnum$ is the condition number of the inner objective function.
Moreover, since Nystr\"om's method is a constant-time operation, dependent only on the number of columns sampled to build a low-rank approximation, Nystr\"om's method eliminates a factor of $\order(\sqrt{\condnum})$ from the IHVP computation as compared to CG~\cite{ji2021bilevel}.

\vspace{-2.5mm}
\paragraph{Related Work on Game Theoretic AC}

Actor-critic as a BLO (i.e., a Stackelberg game) has been considered previously by the RL community.
\citet{zheng2022stackelberg, wen2021characterizing} attempted to solve various MDPs using variants of AC that incorporate a hypergradient.
Their methods are largely unstable, however, because they compute the hypergradient using CG, which can perform arbitrarily badly when the Hessian is ill-conditioned. 
\citet{ChakrabortyBKWM24} developed PARL, a hypergradient-based method for RLHF, which
also uses CG and also suffers from arbitrarily bad estimates of the hypergradient \cite{ding2024sail}.
\citet{Hong2023} assume a linear parametrization for an inner critic, and conclude that the critic should learn at a faster rate than the actor, a result corroborated here and by \citet{zhang2020provably}. 

\begin{figure}[t]
  \centering
  \begin{subfigure}[t]{0.48\columnwidth}
    \centering
    \includegraphics[width=\linewidth]{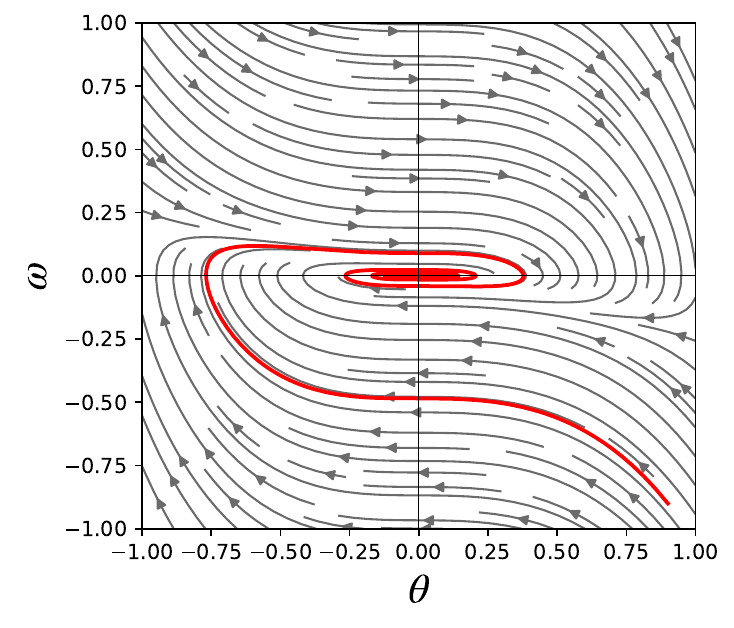}
    \subcaption{Simultaneous}
    \label{fig:simul}
  \end{subfigure}\hfill
  \begin{subfigure}[t]{0.48\columnwidth}
    \centering
    \includegraphics[width=\linewidth]{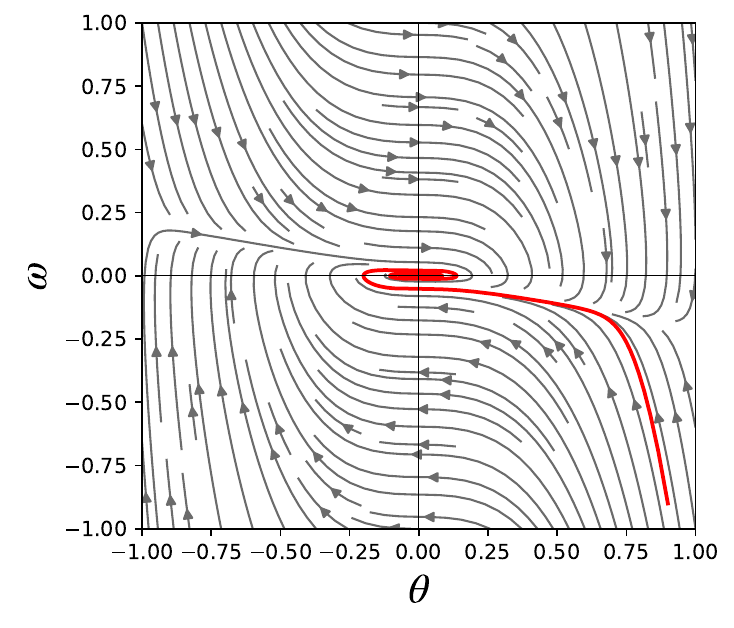}
    \subcaption{TTSA Simultaneous}
    \label{fig:simul_tts}
  \end{subfigure}

  \vspace{0.5em}

  \begin{subfigure}[t]{0.48\columnwidth}
    \centering
    \includegraphics[width=\linewidth]{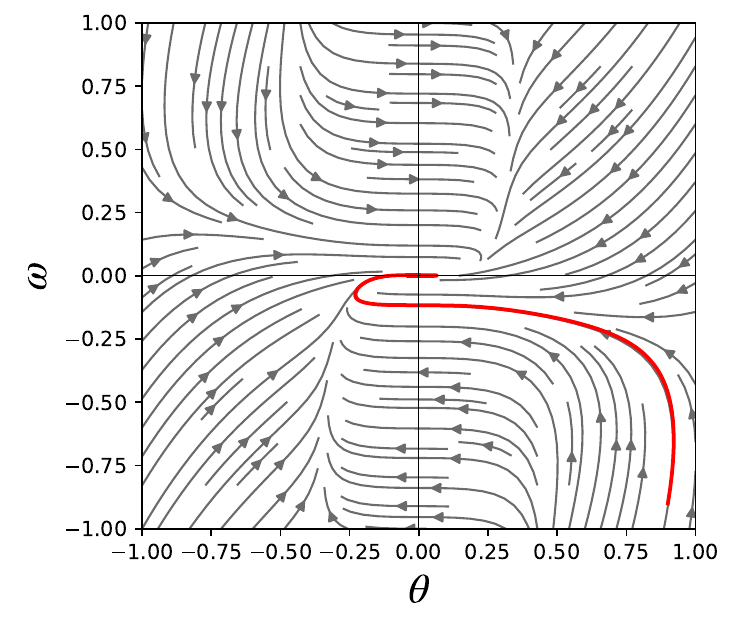}
    \subcaption{Regularized Stackelberg}
    \label{fig:stackelberg_reg}
  \end{subfigure}\hfill
  \begin{subfigure}[t]{0.48\columnwidth}
    \centering
\includegraphics[width=\linewidth]{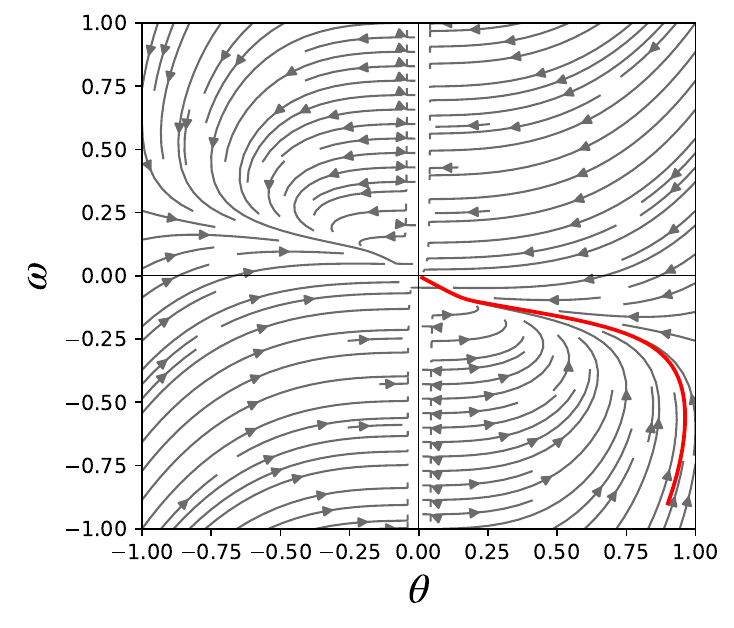}
    \subcaption{Stackelberg}
    \label{fig:stackelberg}
  \end{subfigure}
  \caption{We extend the one-step MDP problem from \citet{zheng2022stackelberg} to show the cycling behavior of simultaneous training dynamics (\Cref{fig:simul}), even with TTSA (\Cref{fig:simul_tts}). Stackelberg dynamics (\Cref{fig:stackelberg}) converge to the equilibrium, $(0,0)$, even when regularized (\Cref{fig:stackelberg_reg}).
  Further details can be found \color{magenta} \href{https://arxiv.org/abs/2505.11714}{here}.}
  \label{fig:toy}
  \vspace{-1.5em} 
\end{figure}

Regarding AC as a Nash (i.e., simultaneous-move) game,
\citet{castro2010convergent} analyze two-timescale stochastic approximate (TTSA) AC algorithms with linear function approximation, and find that simultaneous TTSA AC algorithms converge to a neighborhood around a local Nash equilibrium.
The size of this neighborhood can be decreased by speeding up the learning rate of the critic at the risk of increased instability. 
\citet{heusel2017gans} sharpens this analysis to show that TTSA AC with a faster critic converges to a local Nash equilibrium almost surely.
Note that these results concern Nash not Stackelberg equilibria, as these AC variants rely only on gradients, not on hypergradients.
The behavior of simultaneous vs.\ Stackelberg training dynamics in a simple single-step MDP is shown in \Cref{fig:toy}.


Assuming a linear parameterization for the critic, our 
formulation of AC is as an unconstrained non-convex (outer) strongly-convex (inner) BLO. 
\citet{ghadimi_approximation_2018} analyze an iterative approach to solving non-convex strongly-convex BLOs using hypergradients, the implicit function theorem (IFT), and the conjugate gradient method.
\citet{ji2021bilevel} sharpen this analysis using warm starts in the inner optimization.
\citet{hataya2023Nystrom} introduce the Nystr\"om method to estimate the hypergradient with a low-rank approximation, overcoming challenges associated with iterative methods \cite{lorraine2020optimizing}.
We combine ideas from all of the above to prove polynomial-time convergence, with high probability, to (a point that satisfies the necessary conditions of) a local SSE using hypergradients, the IFT, and Nyström's method, and we validate our approach experimentally.



%% file: sections/prelim.tex
\section{Mathematical Preliminaries}
\label{sec_app:prelims}

\paragraph{Notation} 

We use calligraphic uppercase letters to denote sets (e.g., $\calX$),
uppercase letters to denote matrices (e.g., $H$),
bold lowercase letters to denote vectors (e.g., $\hvp$), and
lowercase letters to denote scalar quantities (e.g., $x$). 
We denote an element of a matrix $H$ by the corresponding lowercase letter using the index as a subscript, e.g., $\hessentry[ij]$ denotes the element at the $i$th row of the $j$th column. 
Similarly, we denote the $j$th entry of a vector by the same lowercase letter with subscript $j$, e.g., $v_j$.
We use $\R$ to denote the set of real numbers. 
We use $\Vert \cdot \Vert_2$ to denote the Euclidean norm and $\Vert \cdot \Vert_{\onorm}$ to denote the operator norm. 
We denote a function's parameters by a subscript (e.g., $f_\x$).
We denote an iteration step by $\iter$ using superscripted brackets (e.g., $x^{(\iter)})$. 
We denote a total gradient by $\grad$ and a partial derivative by $\grad$ with a subscript, e.g., the partial derivative of $\outerobj (\outer, \inner)$ w.r.t.\ $\outer$ is written $\grad[\outer] \outerobj (\outer, \inner)$.


\paragraph{Concepts}

A function $f: \R^n \to \R$ is said to be $L$-Lipschtiz continuous w.r.t. $\Vert \cdot \Vert$ iff $\forall \outer, \outer[][\prime] \in \R^n$, $\Vert f(\outer) - f(\outer[][\prime]) \Vert\ \leq L\Vert \outer - \outer[][\prime]\Vert$. 
A function $f$ is $\strongconst$-convex if $f(\outer) \geq f(\outer[][\prime]) + \langle \grad[\outer]f(\outer[][\prime]), \outer - \outer[][\prime] \rangle + \nicefrac{\strongconst}{2}\Vert \outer - \outer[][\prime]\Vert^2$. 
For a symmetric matrix $A$, we write $0 \precsim A$ if all eigenvalues of $A$ are nonnegative, which implies $A$ is positive semidefinite. 
If all eigenvalues of are strictly positive, then $A$ is positive definite, and we write $0 \prec A$. Additionally, if $0 \precsim A - B$, then we write $B \precsim A$.
If a function $f$ has $L$-Lipschitz continuous gradients, and is twice differentiable, its Hessian satisfies $\grad[][2] f(\outer) \precsim L\I$.
If the function $f$ is also $\strongconst$-convex, its condition number is $\condnum = \nicefrac{L}{\strongconst}$ and $\strongconst \I \precsim \grad[][2] f(\outer) \precsim L\I$.
For $\epsilon > 0$, the Euclidean $\epsilon$-ball around $x$ is defined as $\ball[\epsilon](x) = \{y \in \R^d \mid \Vert y - x \Vert_2 \, \leq \epsilon\}$.

\begin{definition}
Let $h: \R^d \to \R$ be differentiable function. 
A point $\outereqbm \in \R^d$ is called an \mydef{$\epsilon$-first-order stationary point} of $h$ iff the gradient of $h$ at $\outereqbm$ does not exceed $\epsilon$, i.e., $\Vert \grad h(\outereqbm) \Vert^2 \, \leq \epsilon$, for 
$\epsilon > 0$.
\label{def:stationary-point}
\end{definition}


\begin{definition} 
Let $\outerobj: \R^{\outerdim} \times \R^{\innerdim} \to \R$ and $\innerobj: \R^{\innerdim} \to \R$ be continuously differentiable everywhere, and assume $\innerobj$ is $\strongconst$-strongly convex.
For $\epsilon > 0$, we say that $(\outereqbm, \innereqbm)$ is an \mydef{$\epsilon$-bilevel stationary point} iff it satisfies lower-level $\order (\sfrac{\epsilon}{\strongconst})$-optimality, i.e., $\Vert \innereqbm - \argmin_{\inner \in \innerset (\outereqbm)} \innerobj[][\outereqbm] (\inner) \Vert \, \leq \order (\sfrac{\epsilon}{\strongconst})$, and $\outereqbm$ is an \mydef{$\epsilon$-first-order stationary point} of $f$, i.e., $\Vert \grad \hyperobj (\outereqbm) \Vert^2 \, \leq \epsilon$.
\label{blo-stationary-point}
\end{definition}

We seek an efficient algorithm to compute a solution to \Cref{eq:BLO} in the form of an $\epsilon$-bilevel stationary point, which we can then apply to solve \Cref{eq:AC-BLO}.

%% file: sections/nystrom.tex
\section{BLO with Nystr\"om Hypergradients}
\label{sec:blo}

\paragraph{Implicit Function Theorem}

Recall the definition of a BLO presented in \Cref{eq:BLO}.
Given its hierarchical nature, the outer gradient must take into account how changes in the outer variable $\outer$ affect the inner variable $\inner$. 
This \emph{hypergradient} is given by
\begin{align}
\grad \outerobj (\outer, \inner[][*] (\outer)) = \grad[\outer] \outerobj (\outer, \inner) + \grad \inner[][*] (\outer)  \grad[\inner] \outerobj (\outer, \inner).
\end{align}


The main challenge in computing the hypergradient lies in computing the implicit gradient $\grad \inner[][*] (\outer)$.
Assuming $\grad \outerobj$ is continuously differentiable and $\grad[\inner \inner][2]\innerobj[][\outer]$ is well-behaved, if an optimizer of $\innerobj[][\outer]$ exists (which we can ensure, for example, by assuming strong convexity), then we can invoke the implicit function theorem (IFT) to evaluate $\grad \inner[][*] (\outer)$ without explicilty solving for $\inner[][*](\outer)$:

\begin{theorem}
Assuming $\grad[] \outerobj(\outer,\inner)$ is continuously differentiable and $\grad[\inner \inner][2] \innerobj[][\outer] (\inner)$ is invertible, if $\innerpoint (\outerpoint)$ is a first-order stationary point of $\innerobj[][\outer]$ in a neighborhood around $\outerpoint$, i.e., $\grad[\inner] \innerobj[][\outer] (\inner) \vert_{\outerpoint, \innerpoint (\outerpoint)} \ = 0$, for all $\outer \in \ball(\outerpoint)$, then there exists a unique continuous function $\outer \mapsto \inner[][*](\outer)$ s.t.\
$\grad[\inner] \innerobj[][\outer] (\inner) \vert_{\outer, \inner[][*] (\outer)} \ = 0$, for all $\outer \in \ball(\outerpoint)$.
Moreover,
\begin{align}
    \grad\inner[][*] (\outer) \bigg \vert_{\outerpoint} = - \grad[\outer \inner][2] \innerobj[][\outer] (\inner) (\grad[\inner \inner][2] \innerobj[][\outer] (\inner))^{-1} \bigg \vert_{\outerpoint, \inner[][*] (\outerpoint)}.
\end{align}
\end{theorem}

By applying the IFT, we can re-express the second term of the hypergradient as follows: 
\begin{align}
\grad \inner[][*] (\outer) \grad[\inner] \outerobj (\outer, \inner)
&= - \underbrace{\grad[\outer \inner][2] \innerobj[][\outer] (\inner) \underbrace{(\grad[\inner \inner][2] \innerobj[][\outer] (\inner))^{-1} \grad[\inner] \outerobj (\outer, \inner)}_{\text{inverse Hessian vector product}}}_{\text{Jacobian vector product}}.
\label{eq:IFT}
\end{align}
We then define the \mydef{inverse Hessian vector product (\IHVP)} $\hvp[][*] \doteq (\grad[{\inner \inner}][2] \innerobj[][\outer] (\inner))^{-1} \grad[\inner] \outerobj (\outer, \inner)$.

Our goal thus reduces to computing the inverse Hessian of the inner objective.
Since modern neural networks can be millions of parameters in size, storing and directly inverting their Hessians is impractical: an $m \times m$ Hessian would require $\order(m^3)$ operations \cite{layton2020numerical}.
Nonetheless, 
it is possible to compute the \IHVP{} via iterative methods, like conjugate gradient \cite{dagréou2024howtocompute}.

\vspace{-2.5mm}
\paragraph{Approximating the hypergradient}




Given the computational intractability of directly computing the \IHVP, many approaches have been proposed to instead approximate $\hvp$. 
One of the most popular is the conjugate gradient (CG) method \cite{hestenes1952methods}, which iteratively refines the solution to the linear system $\bm{A} \bm{x} = \bm{b}$ by updating in conjugate directions until convergence to $\bm{x} = \bm{A}^{-1}\bm{b}$.
In our application, $\bm{A} = \grad[\inner \inner][2] \innerobj[][\outer]$ and $\bm{b} = \grad[\inner] \outerobj$.

A standard way to improve the performance of CG is to use a preconditioner (PCG): choose a symmetric positive definite matrix $\bm{M}$, and then solve $A' \bm{\hat{x}} = b'$, where
$A' = \bm{M}^{-1/2} \bm{A} \bm{M}^{-1/2}$, $\bm{\hat{x}} = \bm{M}^{-1/2} \bm{x}$, and $b' = \bm{M}^{-1/2} b$.
If the preconditioner is chosen well, PCG reduces the effective condition number---for example, by rescaling the diagonal---yielding faster convergence.
Choosing a preconditioner can be an art, however \cite{martens2010deep}.

\begin{figure}
    \centering
    \includegraphics[width=\linewidth]{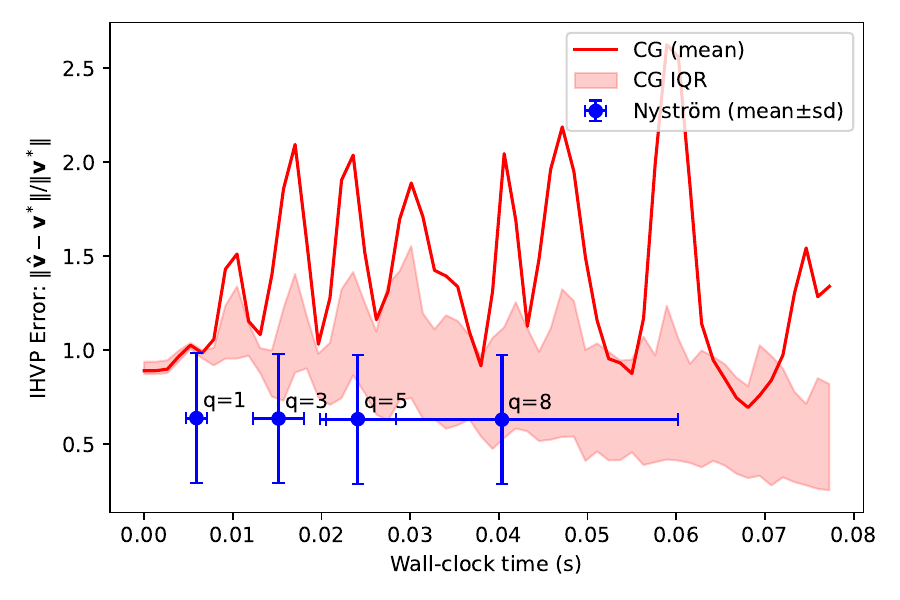}
    \caption{We calculate the IHVP error averaged over 100 small neural networks using both the Nyst\"om method and CG.
    The CG mean is consistently outside the interquartile range (IQR), which suggests at least occasional occurrence of massive errors.
    Further experimental details can be found \color{magenta} \href{https://arxiv.org/abs/2505.11714}{here}.}
    \label{fig:ill-cond}
\end{figure}

While CG overcomes the memory issue of storing the entire Hessian, it can take an arbitrary number of iterations to converge \cite{frangella2023randomized}.
Furthermore, and key to our main contribution, is the fact that CG requires a well-conditioned matrix to converge.
If the matrix is not well-conditioned, numerical errors can accumulate, and the approximation can become arbitrarily bad.
As shown in \Cref{fig:ill-cond}, ill conditioning can arise from the inherent non-convexity of deep neural networks and from overparameterization \cite{saarinen1993ill, paige1975solution}, which can cause the IHVP error to spike when using CG.

\paragraph{The Nystr\"om Method}
Proposed by \citet{hataya2023Nystrom}, the Nystr\"om method produces a low-rank approximation of the \IHVP{}.
Assume a $\Hessdim$-dimensional positive (semi)definite Hessian denoted by $\Hessian$ and a low-rank approximation of $\Hessian$ denoted by $\Hessian[\rank]$, for $\rank \ll \Hessdim$.
By selecting a set of $\randi$ indices of size $\rank$ at random, we obtain the Nystr\"om approximation
$\Hessian[\rank] = \Hessian[[:, \randi]] \Hessian[[\randi, \randi]][\dagger] \Hessian[[:, \randi]][\top]$,
where $\Hessian[[:, \randi]] \in \R^{\Hessdim \times \rank}$ is a matrix of select columns of $\Hessian$ at indices $\randi$;
$\Hessian[[\randi, \randi]] \in \R^{\rank \times \rank}$ is a matrix of select rows of $\Hessian[[:, \randi]]$ at indices $\randi$;
and $\Hessian[[\randi, \randi]][\dagger]= \bm{U}\bm{\Lambda}^{-1}\bm{U}^{\top}$ is the Moore-Penrose pseudoinverse of $\Hessian[[\randi, \randi]]$, where $\bm{U}$ are the eigenvectors and $\bm{\Lambda}$  are the eigenvalues of $\Hessian[[\randi, \randi]]$.
For some small regularization constant $\nystromconst > 0$, we define the $\nystromconst$-regularized \IHVP{} as $(\Hessian[\rank] + \nystromconst \I)^{-1}$, where $\I$ is the $p$-dimensional identity matrix.%
\footnote{Incorporating the small perturbation $\nystromconst \I$ into this \IHVP{} is equivalent to optimizing a proximal regularization of the inner objective: i.e., $\innerobj[][\outer] (\inner) + \nicefrac{\nystromconst}{2}\Vert \inner - \arg \min_\inner \innerobj[][\outer] (\inner) \Vert$ \cite{lorraine2020optimizing}.}
This $\nystromconst$-regularized \IHVP{} decomposes as follows:
$(\Hessian[\rank] + \nystromconst \I)^{-1} = (\Hessian[[:, \randi]] \Hessian[[\randi, \randi]][\dagger] \Hessian[[:, \randi]][\top] + \nystromconst \I) ^{-1}$.
Next, applying the Woodbury matrix identity allows us to obtain:
\begin{align*}
    (\Hessian[\rank] + \nystromconst \I)^{-1}  = \frac{1}{\nystromconst} \I - \frac{1}{\nystromconst^2}\Hessian[[:, \randi]] \bigg( \Hessian[[\randi:\randi]] + \frac{1}{\nystromconst} \Hessian[[:, \randi]][\top] \Hessian[[:, \randi]] \bigg)^{-1} \Hessian[[:, \randi]][\top].
\end{align*}
This final expression allows us to approximate the \IHVP{} $\hvp$ by $\esthvp = (\Hessian[\rank] + \nystromconst \I)^{-1} \grad[\inner] \outerobj (\outer, \innerest)$.
The regularization parameter $\nystromconst$ plays a crucial role in balancing numerical stability and the fidelity of curvature information. 
As discussed by \citet{vicol_implicit_2022}, for an eigenvalue $\lambda$ of the Hessian $H$, the effect of regularization is to modify the inverse eigenvalue as $(\lambda + \nystromconst)^{-1}$. 
When $\lambda \gg \nystromconst$, this term behaves as $\lambda^{-1}$, preserving curvature information. 
Conversely, when $\nystromconst \gg \lambda$, the approximation becomes insensitive to low-curvature directions, effectively ignoring them.  
In practice, $\nystromconst$ is selected empirically. 
One practical advantage of the Nystr\"om method is its ability to operate effectively with smaller $\nystromconst$.
For example, we use $\nystromconst = 50$, compared to $\nystromconst = 500$ in \citet{zheng2022stackelberg} for continuous control RL, and $\nystromconst = 10,000$ in \citet{fiez_implicit_2020} for a GAN-based bilevel problem, both of which use CG.
Notably, \citet{lorraine2020optimizing} observed that using the identity matrix, which is equivalent to choosing $\nystromconst \gg \firstlip$ and then rescaling the IHVP, matched the performance of approximate IHVPs computed via CG.
We find that the Nystr\"om method offers an attractive balance, where regularization remains modest without sacrificing stability. \Cref{alg:nystrom-blo} presents our bilevel optimization algorithm with Nystr\"om hypergradients.

\input{sections/algo}

%% file: sections/algo.tex
\begin{algorithm}
\caption{BLO with Nystr\"om Hypergradients}
\label{alg:nystrom-blo}
\begin{minipage}{1.1\textwidth}
\begin{algorithmic}
    \INPUT $\iters[\outer], \iters[\inner], \outer[][(0)], \inner[][(0)]$, and learning rates $\learnrate[\outer], \learnrate[\inner]$

    \OUTPUT $\left( \left\{ \outer[][][{\iter}] \right\}_{\iter=0}^{\iters[\outer]-1}, \left\{ \inner[][][{\iter}] \right\}_{\iter=0}^{\iters[\inner]-1} \right)$ 
    
    
    \FOR{$\iter = 0, \ldots, \iters[\outer] - 1$}
    
    \STATE $\temp[][][0] \gets \inner[][][\iter - 1]$ if $\iter > 0$ else $\inner[][][0]$
    \COMMENT{Warm starts}
    
    \FOR{$\inneriter = 0, \ldots, \iters[\inner] - 1$}
    
    \STATE $\temp[][][\inneriter+1] \gets \temp[][][\inneriter] - \learnrate[\inner] \grad[\inner] \innerobj[][{\outer[][][\iter]}] (\temp[][][\inneriter])$
    \ENDFOR

    \STATE $\inner[][][\iter] \gets \temp[][][{\iters[\inner]}]$

    \STATE calculate $\esthvp$ using the Nystr\"om method
    
    \STATE $\grad \esthyperobj(\outer[][][\iter]) \gets \grad[\outer] \outerobj (\outer[][][\iter], \inner[][][\iter]) - \grad[\outer \inner][2] \innerobj[][{\outer[][][\iter]}] (\inner[][][\iter]) \, \esthvp$
    
    \STATE $\outer[][][\iter+1] \gets \outer[][][\iter] - \learnrate[\outer] \grad \esthyperobj (\outer[][][\iter])$
    \ENDFOR
\end{algorithmic}
\end{minipage}
\end{algorithm}

%% file: sections/conv.tex
\section{Convergence of BLO-Nystr\"om}
\label{sec:conv}

We make several convexity and smoothness assumptions, which are standard in the BLO literature \cite{ghadimi_approximation_2018, ji2021bilevel}.
Our first assumption ensures that the inner optimization admits a unique solution $\innereqbm$ for a given $\outer$, and thus implies invertibility of the Hessian.

\begin{assumption}[Strong convexity]
\label{ass:strongly-convex}
The inner objective is $\strongconst$-strongly convex in the inner variable, i.e., $\inner \mapsto \innerobj[][\outer] (\inner)$ is $\strongconst$-strongly-convex. 
This assumption ensures the Hessian $\grad[\inner \inner][2] \innerobj[][\outer]$ is full rank.
\end{assumption}

The next three assumptions pertain to the smoothness of both the inner and outer objective functions, meaning they ensure that the first and second-order gradients are well behaved.
In particular, they imply the condition number $\condnum = \nicefrac{\firstlip}{\strongconst}$ for the inner objective $\innerobj[][\outer]$.
Moreover, Assumptions~\ref{ass:strongly-convex} and \ref{ass:first-order-lip} are sufficient for the inverse function theorem
to apply, because together they imply $\inner[][*](\outer)$ is a first-order stationary point.

\begin{assumption}[Lipschitz continuity]
\label{ass:first-order-lip}
Both the inner and outer objectives, $\outerobj$ and $\innerobj[][\outer]$, have $\firstlip$-Lipschitz bounded gradients: 
i.e., for all $\outer, \outer[][\prime] \in \R^{\outerdim}$, and $\inner, \inner[][\prime] \in \R^\innerdim$,
\begin{align*}
    &\Vert \grad \outerobj (\outer, \inner) - \grad \outerobj (\outer[][\prime], \inner[][\prime]) \Vert \ \leq \firstlip \Vert (\outer, \inner) - (\outer[][\prime], \inner[][\prime])\Vert, \\
    &\Vert \grad \innerobj[][\outer] (\inner) - \grad \innerobj[][\outerprime] (\inner[][\prime]) \Vert \ \leq \firstlip \Vert (\outer, \inner) - (\outer[][\prime], \inner[][\prime])\Vert.
\end{align*}
\end{assumption}

\begin{assumption}[Lipschitz smoothness]
\label{ass:second-order-lip}
The inner objective $\innerobj[][\outer]$ is also $\secondlip$-Lipschitz smooth, i.e., its second derivatives are bounded: for all $\outer, \outer[][\prime] \in \R^{\outerdim}$, and $\inner, \inner[][\prime] \in \R^\innerdim$,
\begin{align*}
    &\Vert \grad[\outer \inner] \innerobj[][\outer] (\inner) - \grad[\outer \inner] \innerobj[][\outerprime] (\inner[][\prime]) \Vert \ \leq \secondlip \Vert  (\outer, \inner) - (\outer[][\prime], \inner[][\prime]) \Vert, \\
    &\Vert \grad[\inner \inner][2] \innerobj[][\outer] (\inner) - \grad[\inner \inner][2] \innerobj[][\outerprime] (\inner[][\prime]) \Vert \ \leq \secondlip \Vert  (\outer, \inner) - (\outer[][\prime], \inner[][\prime]) \Vert.
\end{align*}
\end{assumption}

\begin{assumption}
\label{ass:grad-bound}
There exists a constant $\gradbound \in \R^{++}$ s.t.\ for all $\outer \in \R^{\outerdim}$, $\Vert \grad[\inner] \outerobj (\outer, \inner) \Vert \, \leq \gradbound$.
\end{assumption}


\citet{ghadimi_approximation_2018} were the first to provide convergence results for non-convex $\hyperobj$ and strongly-convex $\innerobj[][\outer]$. 
Their analysis was sharpened by \citet{ji2021bilevel}, whose analysis also relied on conjugate gradient to compute the IHVP.
Our analysis extends the literature by considering the Nystr\"om method, rather than CG, in this non-convex strongly-convex case.
Specifically, we characterize the irreducible error incurred by the random sampling of columns.

Our first result%
\footnote{The proof of all claims in this paper can be found in 
\color{magenta} \href{https://arxiv.org/abs/2505.11714}{here}.}
justifies the use of warm-starts in \Cref{alg:nystrom-blo}: if the value of the outer variable changes slightly, then the corresponding inner solutions do not stray too far from one another either:

\begin{restatable}[Warm Starts]{lemma}{warmstarts}
Under \Cref{ass:strongly-convex} and~\ref{ass:first-order-lip},
for all $\outer, \outerpoint \in \R^{\outerdim}$, $\inner \in \argmin_\both \innerobj[][\outer](\both)$, and $\innerpoint \in \argmin_\both \innerobj[][\outerpoint](\both)$,
    $\norm[\inner - \innerpoint] \leq \nicefrac{\firstlip} {\strongconst} \norm[\outer - \outerpoint]$.
\label{lem:warm-starts}
\end{restatable}

\begin{restatable}[Convergence]{theorem}{convergence}
Given $\strongconst, \firstlip, \secondlip, \gradbound$ as defined in Assumptions \ref{ass:first-order-lip} through \ref{ass:grad-bound}, a Nystr\"om regularization constant $\nystromconst$, and a hypergradient smoothness parameter $\hyperlip \in \order (\condnum^3)$.
Choose $\learnrate[\inner] = \nicefrac{1}{\firstlip}$, $\learnrate[\outer] = \nicefrac{1}{8\hyperlip}$, $\highprob \in (0, 1]$, $\iters[\inner] \in \order(\condnum)$ inner iterations, and $K \doteq \iters[\outer] \in \mathbb{N}$ outer iterations. 
Then, with probability at least $1 - \order (\highprob)$, the outputs 
of \Cref{alg:nystrom-blo} satisfy $\frac{1}{K} \sum^{K-1}_{\iter=0} \norm[{\hyperobj (\outer[][][\iter])}]^2$ $$\leq \frac{1}{K} \left( 64 \hyperlip (\hyperobj (\outer[][][0]) - \inf_{\outer} \hyperobj (\outer)) + 5 \norm[{\inner[][0][0] - \inner[][0]}]^2 \right) + 10 \firstlip^2\Psi^2,$$ where $\Psi$ is the irreducible error incurred by the Nystr\"om approximation. 
In order to obtain an $\epsilon$-stationary point, for each IHVP calculation, sample $\rank \geq \order \bigg( \frac{\truerank \log \left( \nicefrac{1}{\delta} \right)}{(\nicefrac{\epsilon}{2})^4} \bigg)$ columns of $\grad[\inner \inner][2] \innerobj[][\outer]$, where the indices chosen are sampled proportionally to the square of the values of the diagonal elements of $\grad[\inner \inner][2] \innerobj[][\outer]$.
Then, the requisite number of gradient computations are 
$\order (\frac{\condnum^3}{\nicefrac{\epsilon}{2}})$ for $\grad[\outer] \outerobj$;
$\order ({\frac{\condnum^4}{\nicefrac{\epsilon}{2}}})$ for $\grad[\inner] \innerobj[][\outer]$;   
$\order ({\frac{\condnum^3}{\nicefrac{\epsilon}{2}}})$ for the \IHVP{} $\esthvp$; and $\order ({\frac{\condnum^3}{\nicefrac{\epsilon}{2}}})$ for the Jacobian vector product $\grad[\outer \inner][2] \innerobj[][\outer] \esthvp$.
\label{thm:convergence}
\end{restatable}

\Cref{thm:convergence} states that \Cref{alg:nystrom-blo} can be used to solve a non-convex strongly-convex BLO.
By carefully selecting the learning rates $(\learnrate[\outer], \learnrate[\inner])$ and the number of inner and outer iterations $(\inneriters, \iters)$, and by column sampling according to eigenvalue magnitude, we derive an explicit convergence rate 
to an $\epsilon$-bilevel stationary point of \Cref{eq:BLO}.



%% file: sections/RL.tex
\section{Bilevel Policy Optimization (BLPO)}
\label{sec:algo}




Vanilla actor-critic algorithms run gradient descent for the actor and the critic  simultaneously, updating $\actorparams$ and $\criticparams$ much like a better-response dynamic:
\begin{equation*}
    \actorparams \gets \actorparams + \learnrate[\actorparams] \grad[\actorparams] \estactorobj (\actorparams, \criticparams)
    \quad \quad \quad
    \criticparams \gets \criticparams - \learnrate[\criticparams] \grad[\criticparams] \estcriticobj (\actorparams, \criticparams).
\end{equation*}
(Here, $\learnrate[\actorparams]$ and $\learnrate[\criticparams]$ are the actor's and critic's learning rates, respectively.)
%
%
Our approach, as already described, is \emph{not\/} to update these two sets of parameters simultaneously, but rather to solve the following BLO by descending on the hypergradient for the actor, and the gradient for critic:
\begin{equation*}
    \min_{\actorparams \in \R^{\outerdim}} - \actorobj (\actorparams, \criticparams[][*] (\actorparams)) \quad \text{s.t.} \quad
    \criticparams[][*] (\actorparams) \in \arg \min_{\criticparams \in \R^{\innerdim}} \criticobj[][\actorparams] (\criticparams). \nonumber
\end{equation*}

The main challenge that arises with our approach is computing the actor's hypergradient, namely:
\begin{align}
\grad[\actorparams] \actorobj (\actorparams, \criticparams[][*] (\actorparams))
&= \grad[\actorparams] \actorobj (\actorparams, \criticparams) + (\grad[\actorparams] (\criticparams[][*] (\actorparams)) \grad[\criticparams] \actorobj (\actorparams, \criticparams), \\
&= \grad[\actorparams] \actorobj (\actorparams, \criticparams) - \grad[\actorparams \criticparams][2] \criticobj[][\actorparams] (\criticparams) (\grad[\criticparams][2] \criticobj[][\actorparams] (\criticparams))^{-1} \grad[\criticparams] \actorobj (\actorparams, \criticparams).
\end{align}


There are four gradients involved in this formula. 
Three of them,
$\grad[\actorparams] \actorobj (\actorparams, \criticparams)$, $\grad[\criticparams] \actorobj (\actorparams, \criticparams)$, and $\grad[\criticparams] \criticobj[][\actorparams] (\criticparams)$, can be easily obtained through autodifferentiation.
The only potential difficulty lies in computing
$\grad[\actorparams] \criticobj[][\actorparams] (\criticparams)$, since $\criticobj$ is parameterized by $\actorparams$, and as such, is not directly a function of $\actorparams$.
In the next theorem, we invoke the policy gradient theorem to derive an analytical expression for $\grad[\actorparams] \criticobj[][\actorparams] (\criticparams)$, which we can use along with $\grad[\criticparams] \criticobj[][\actorparams] (\criticparams)$ to compute the mixed partial 
$\grad[\actorparams \criticparams][2] \criticobj[][\actorparams] (\criticparams)$.

\begin{restatable}[]{theorem}{surrogate}
The gradient of the critic's objective function with respect to the actor's parameters, $\grad[\actorparams] \criticobj[][\actorparams] (\criticparams)$, is:
\begin{align}
\Ex_{\state \sim \statedist[{\policy[\actorparams]}][\initstatedist]} \left[ \left( \valfun[][{\policy[\actorparams]}] (\state) - \valfun[\criticparams] (\state) \right) \bigg(  \Ex_{\substack{{\state'} \sim \statedist[{\policy[\actorparams]}][\state] \\ \action \sim \policy[\actorparams] (\state')}} \left[ \grad[\actorparams] \log \policy[\actorparams] (\action \mid \state') \, \Qfunc[][{\policy[\actorparams]}] (\state', \action) \right] \bigg) \right],
\label{eq:surrogate-gradient}
\end{align}
where $\policy[\actorparams] (\action \mid \state)$ denotes the probability of taking action $\action$ under policy $\policy$ at state $\state$.
\label{thm:surrogate-gradient}
\end{restatable}

\if{0}
\begin{proof}
\begin{align}
\grad[\actorparams] \criticobj[][\actorparams] (\criticparams)
&= \frac{1}{2} \grad[\actorparams] \Ex_{\substack{\state \sim \statedist[{\policy[\actorparams]}][\initstatedist]}}  \left[ \left(  \Vfunc[][{\policy[\actorparams]}] (\state) - \Vfunc[\criticparams] (\state) \right)^2 \right] \\
&= \Ex_{\state \sim \statedist[{\policy[\actorparams]}][\initstatedist]} \left[ \left( \valfun[][{\policy[\actorparams]}] (\state) - \valfun[\criticparams] (\state) \right) \grad[\actorparams] \left( \valfun[][{\policy[\actorparams]}] (\state) - \valfun[\criticparams] (\state) \right) \right] \\
&= \Ex_{\state \sim \statedist[{\policy[\actorparams]}][\initstatedist]} \left[ \left( \valfun[][{\policy[\actorparams]}] (\state) - \valfun[\criticparams] (\state) \right) \grad[\actorparams] \valfun[][{\policy[\actorparams]}] (\state) \right]
\end{align}

Now, applying the log-derivative trick yields:
\begin{align}
\grad[\actorparams] \valfun[][{\policy[\actorparams]}] (\state)
\propto& 
\Ex_{{\traj[][\state]} \sim {\histdist[{\policy[\actorparams]}][\state]}} \left[ \grad[\actorparams] \log \histdist[{\policy[\actorparams]}][\state] (\traj[][\state]) \, \return[{\traj[][\state]}] \right]
\end{align}

Finally, by the policy gradient theorem \cite{sutton1999policy},
\begin{align}
\Ex_{{\traj[][\state]} \sim {\histdist[{\policy[\actorparams]}][\state]}} \left[ \grad[\actorparams] \log \histdist[{\policy[\actorparams]}][\state] (\traj[][\state]) \, \return[{\traj[][\state]}] \right]
=& \Ex_{\substack{{\state'} \sim \statedist[{\policy[\actorparams]}][\state] \\ \action \sim \policy[\actorparams] (\state')}} \left[ \grad[\actorparams] \log \policy[\actorparams] (\action \mid \state') \, \Qfunc[][{\policy[\actorparams]}] (\state', \action) \right]
\end{align}
\end{proof}


\begin{corollary}
Given $\numsamples$ sampled trajectories  of length $\timesteps$, $\{ (\state[0][1], \action[0][1], \reward[0][1], \state[1][1]$, $\ldots$, $\reward[\timesteps-1][0])$, $\ldots$, $(\state[0][\numsamples], 
\action[0][\numsamples], \reward[0][\numsamples], \state[1][\numsamples], \ldots, \reward[\timesteps-1][\numsamples]) \}$,
the gradient of the critic's objective function with respect to the actor's parameters can be estimated by:

\arjun{why does the batch start at 1, and the trajectory start at 0?} \amy{you can change this if you care to, but doing so might introduce bugs.}
\begin{align}
    &\frac{1}{\numsamples T} \sum^{\numsamples}_{m=1} \sum^{\trajlength - 1}_{t=0} \left( \valfun[][{\policy[\actorparams]}] (\state[t][\sample]) - \valfun[\criticparams] (\state[t][\sample]) \right)
    \left( \frac{1}{T-t} \sum^{\trajlength - 1 -t}_{k=0} \grad[\actorparams] \log \policy[\actorparams] (\action[t+k][\sample] \mid \state[t+k][\sample]) \, \Qfunc[][{\policy[\actorparams]}] (\state[t+k][\sample], \action[t+k][\sample]) \right)
\end{align}
\end{corollary}
\fi

Notably, we use the Nystr\"om method to approximate the \IHVP{} $\esthvp[AC] = (\grad[\criticparams][2] \criticobj[][\actorparams] (\criticparams))^{-1} \grad[\criticparams] \actorobj (\actorparams, \criticparams)$.
We then compute the Jacobian vector product as $\grad[\actorparams \criticparams] \criticobj (\actorparams, \criticparams) \esthvp[AC]$, which we subtract from $\grad[\actorparams] \actorobj (\actorparams, \criticparams)$ to obtain the hypergradient.  
\BLPO{} is thus a special case of \Cref{alg:nystrom-blo} (BLO with Nystr\"om gradients) in which $\outerobj = \actorobj$ and $\innerobj = \criticobj$.

Next, we 
we present assumptions under which 
the critic's objective is strongly convex.

\begin{restatable}[]{theorem}{convexrl}
Given a policy $\policy[\actorparams]$, assume $\discount < 1$, and suppose the value function $\Vfunc[\criticparams]$ is linearly parameterized, i.e., $\Vfunc[\criticparams] (\state) = \criticparams[][\top]\featurevec(\state)$, for all $\state \in \States$, where $\featurevec: \States \to \R^\innerdim$ is a feature map.
Define $\fdiff(\state, \nextstate) = \featurevec (\state)- \discount \featurevec(\nextstate)$, for all $\state, \state' \in \States$ where $\state$ is the current state and $\nextstate$ is the successor state under $\policy[\actorparams]$.
Then, assuming the covariance matrix $\Ex_{\substack{\state \sim \statedist[{\policy[\actorparams]}][\initstatedist] \\ \nextstate \sim \transition \left[ \cdot \mid \state, \policy[\actorparams] (\state) \right]}} \left[ \fdiff(\state, \nextstate) \fdiff (\state, \nextstate)^{\top} \right]$ is positive definite, $\criticobj[][\actorparams] (\criticparams)$
is $\strongconst$-strongly-convex in $\criticparams$, where the strong-convexity constant $\strongconst$ is the smallest eigenvalue of the covariance matrix.
\label{thm:strong-convex-rl}
\end{restatable}

We have thus established conditions under which our actor-critic formulation, \Cref{eq:AC-BLO}, is indeed a non-convex strongly-convex BLO so that \Cref{thm:convergence} applies. 
As a result, we can design a reinforcement learning algorithm in the style of \Cref{alg:nystrom-blo}, which 
converges in polynomial time to an $\epsilon$-bilevel stationary point of \Cref{eq:AC-BLO}, with high probability.

\input{sections/blpo}



\paragraph{Advantage Functions and Estimators}

In practice, the actor and the critic's objective functions must be estimated.
To reduce the variance in these estimates, they are often computed using advantage functions.
%
%
%
At iteration $k$, when the critic's learned value function is $\valfun[{\criticparams[][(k)]}]$, we define the advantage function as 
$\adv[][(k)] (\state, \action) = \Qfunc[][{\policy[\actorparams]}] (\state, \action) - \valfun[{\criticparams[][(k)]}] (\state)$.
Now, since $\Vfunc[][{\policy[\actorparams]}] (\state) = \Qfunc[][{\policy[\actorparams]}] (\state, \policy[\actorparams] (\state))$,
%
\begin{align}
\criticobj[(k)][\actorparams] (\criticparams)
&= \Ex_{\substack{\state \sim \statedist[{\policy[\actorparams]}][\initstatedist] \\ \action \sim \policy[\actorparams] (\state)}} \left[ \frac{1}{2} \left( \underbrace{\Vfunc[{\criticparams[][(k)]}] (\state) + \adv[][(k)] (\state, \action)}_{\text{target}} - \Vfunc[\criticparams] (\state) \right)^2 \right].
\end{align}

Using this advantage function, we obtain the following corollary of \Cref{thm:surrogate-gradient}.

\begin{corollary}
At iteration $k$, the gradient of the critic's objective function with respect to the actor's parameters is given by: 
\begin{align}
\grad[\actorparams] \criticobj[(k)][\actorparams] (\criticparams[][(k)]) = &\Ex_{\state \sim \statedist[{\policy[\actorparams]}][\initstatedist]} \left[ \left( \valfun[{\criticparams[][(k)]}] (\state) + \adv[][(k)] (\state, \action) - \valfun[\criticparams] (\state) \right) \right. \times \\
&\left. \left( \Ex_{\substack{{\state'} \sim \statedist[{\policy[\actorparams]}][\state] \\ \action \sim \policy[\actorparams] (\state')}} \left[\grad[\actorparams] \log \policy[\actorparams] (\action \mid \state')  \, \adv[][(k)] (\state, \action) \right] \right) \right]. \nonumber
\end{align}
%
\label{cor:surrogate-gradient-advantage}
\end{corollary}

Using a straightforward estimator of this gradient, we obtain a further corollary:

\begin{corollary}
Given $\numsamples$ sampled trajectories of length $\timesteps$ $\{ \left( \state[0][1], \action[0][1], \reward[0][1], \state[1][1] \right.$, $\ldots$, $\left. \reward[\timesteps][0] \right)$, $\ldots$, $\left( \state[0][\numsamples], \action[0][\numsamples], \reward[0][\numsamples], \state[1][\numsamples], \ldots, \reward[\timesteps][\numsamples] \right) \}$, at iteration $k$, the gradient of the critic's objective function can be estimated by:
$\grad[\actorparams] \estcriticobj[(k)][\actorparams] (\criticparams) =
\frac{1}{\numsamples \timesteps} \sum_{\sample = 1}^{\numsamples} \sum_{\timestep = 0}^{\timesteps-1} \left( \valfun[{\criticparams[][(k)]}] \left( \state[\timestep][\sample] \right) + \adv[][(k)] \left( \state[\timestep][\sample], \action[\timestep][\sample] \right) - \valfun[\criticparams] \left( \state[\timestep][\sample] \right) \right) \times$ \\
$\left( \frac{1}{T-t} \sum^{\trajlength -1-t}_{k=0} \grad[\actorparams] \log \policy[\actorparams] \left( \action[t+k][\sample] \mid \state[t+k][\sample] \right) \adv[][(k)] \left( \state[t+k][\sample], \action[t+k][\sample] \right) \right)$.
\label{cor:surrogate-gradient}
\end{corollary}


We can likewise restate the (direct) gradient of the actor's objective using an advantage function and an estimator.
By the policy gradient theorem \cite{sutton2018reinforcement},
%
$\grad[\actorparams] \actorobj (\actorparams, \criticparams) = \Ex_{\substack{{\state} \sim \statedist[{\policy[\actorparams]}][\initstatedist] \\ \action \sim \policy[\actorparams] (\state)}} \left[ \grad[\actorparams] \log \policy[\actorparams] (\action \mid \state) \, \Qfunc[][{\policy[\actorparams]}] (\state, \action) \right]$.
%
%
This gradient can be stated equivalently in terms of $\Qfunc[][{\policy[\actorparams]}] (\state, \action)$
the TD-Residual $\reward[\timestep] + \Vfunc[][\policy[\actorparams]](\state[\timestep + 1]) - \Vfunc[][\policy[\actorparams]](\state[\timestep])$, 
or an advantage function $\adv[][] (\state, \action)$~\cite{gae}.
At iteration $k$, we employ the advantage function $\adv[][(k)] (\state, \action)$ so that
%
$\grad[\actorparams] \actorobj[(k)] (\actorparams, \criticparams) = \Ex_{\substack{\state \sim \statedist[{\policy[\actorparams]}][\initstatedist] \\ \action \sim \policy[\actorparams] (\state)}} \left[ \grad[\actorparams] \log \policy[\actorparams] (\action \mid \state) \, \adv[][(k)] (\state, \action) \right]$.
%
We thus arrive at the following gradient estimator:
\begin{align}
\grad[\actorparams] \estactorobj[(k)]  (\actorparams, \criticparams) \doteq
\frac{1}{\numsamples \timesteps} \sum_{\sample = 1}^{\numsamples} \sum_{\timestep = 0}^{\timesteps - 1} \grad[\actorparams] \log \policy[\actorparams] (\action[\timestep][\sample] \mid \state[\timestep][\sample]) \, \adv[][(k)] \left( \state[\timestep][\sample], \action[\timestep][\sample] \right).
\end{align}

The estimator
$\grad[\actorparams] \estcriticobj[(k)][\actorparams] (\criticparams)$ defined in
\Cref{cor:surrogate-gradient} is only unbiased if we take care to sample the inner $\adv[][(k)]$ from a new trajectory originating at each ${\state[\timestep+k]} \sim \statedist[{\policy[\actorparams]}][{\state[\timestep]}]$.
For reasons of data efficiency, our estimator is biased; that is, we use the same trajectories to compute the inner and the outer $\adv[][(k)]$s.
Our empirical results suggest that this biased estimator is sufficient for strong performance. 
We leave to future work a more thorough investigation of methods to mitigate this bias while preserving data efficiency.

Pseudocode for \BLPO{} with an advantage function and these estimators appears in \Cref{alg:BLPO-Nystrom}. 

%% file: sections/blpo.tex
\begin{algorithm}[t]
\caption{Bilevel Policy Optimization (BLPO)}
\label{alg:BLPO-Nystrom}
\begin{algorithmic}
    \INPUT number of outer (respectively, inner) iterations $\iters[\actorparams]$ (respectively, $\iters[\criticparams]$), length of rollouts $\timesteps$, number of epochs $\numepochs$, number of mini-batches $\numbatches$, learning rates $\learnrate[\actorparams]$ and $\learnrate[\criticparams]$

    \OUTPUT{the actor and critic networks, $\policy[\actorparams]$ and $\Vfunc[\criticparams]$}
    
    \STATE Initialize $\actorparams[][(0)]$, $\criticparams[][(0)]$
    
\FOR{$\iter = 0, 1, \ldots, \iters[\actorparams] - 1$}

    \STATE Collect rollouts of length $\timesteps$ using policy $\policy[{\actorparams[][(k)]}]$
    
    \STATE Use the value function $\valfun[{\criticparams[][(k)]}]$ to compute 
    $\adv[][(k)]$
    
    \FOR{all $\numepochs$ epochs}
        \FOR{all $\numbatches$ mini-batches} 

            
            \STATE $\temp[][][0] \gets \criticparams[][][\iter - 1]$ if $\iter > 0$ else $\criticparams[][][0]$

\renewcommand{\inneriter}{l}
    
            \FOR{$\inneriter = 0, 1, \ldots, \iters[\criticparams] - 1$} 
    

                

            \STATE
                $\criticparams[][(\inneriter + 1)] \gets \criticparams[][(\inneriter)] - \learnrate[\criticparams] \frac{1}{\nicefrac{\timesteps}{ \numbatches}} \grad[\criticparams] \sum_{t=0}^{\nicefrac{\timesteps}{ \numbatches}} \frac12 \left( \valfun[{\criticparams[][(\iter)]}] (\state[\timestep]) + \adv[][(\iter)] (\state[\timestep]) - \valfun[{\criticparams[][(\inneriter)]}] (\state[\timestep]) \right)^2$

            \ENDFOR

            \STATE $\criticparams[][(\iter)] \gets \temp[][{(\iters[\criticparams])}]$

            \STATE Estimate the IHVP via the Nystr\"om method:
$\esthvp[AC] \gets (\grad[\criticparams][2] \estcriticobj[][\actorparams] (\criticparams[][][\iter])^{-1} \grad[\criticparams] \estactorobj (\actorparams[][][\iter], \criticparams[][][\iter]))$       
            
            \STATE Calculate the hypergradient using 
            Equation~12 for the direct gradient and \Cref{cor:surrogate-gradient} for the implicit gradient:
            $\grad[\actorparams] \estactorobj[(\iter)] 
            \gets 
            \grad[\actorparams] \estactorobj(\actorparams[][][\iter], \criticparams[][][\iter]) 
            -
            \grad[\actorparams \criticparams] \, \estcriticobj (\actorparams[][][\iter], \criticparams[][][\iter]) \esthvp[AC]
            $ 
            
            \STATE Update the actor by following the hypergradient:
            $\actorparams[][(\iter + 1)] \gets \actorparams[][(\iter)] + \learnrate[\actorparams] \grad[\actorparams] \estactorobj[(\iter)]$
        \ENDFOR
    \ENDFOR
\ENDFOR
\end{algorithmic}
\end{algorithm}

%% file: sections/experiments.tex
\captionsetup[subfigure]{labelformat=simple}  
\renewcommand\thesubfigure{(\alph{subfigure})}

\section{Experiments}
\label{sec:experiments}

\newlength{\panelW}
\setlength{\panelW}{0.235\textwidth} 
\newcommand{\panelH}{3.15cm}         

\begin{figure*}[t]
  \centering
  \setlength{\tabcolsep}{4pt} 

  \begin{tabular}{@{} c c c !{\vrule width 0.6pt} c @{}}
    \subcaptionbox{Walker2d\label{fig:walker2d_Nystrom_vanilla}}[\panelW]{%
      \includegraphics[width=\linewidth,height=\panelH,keepaspectratio]{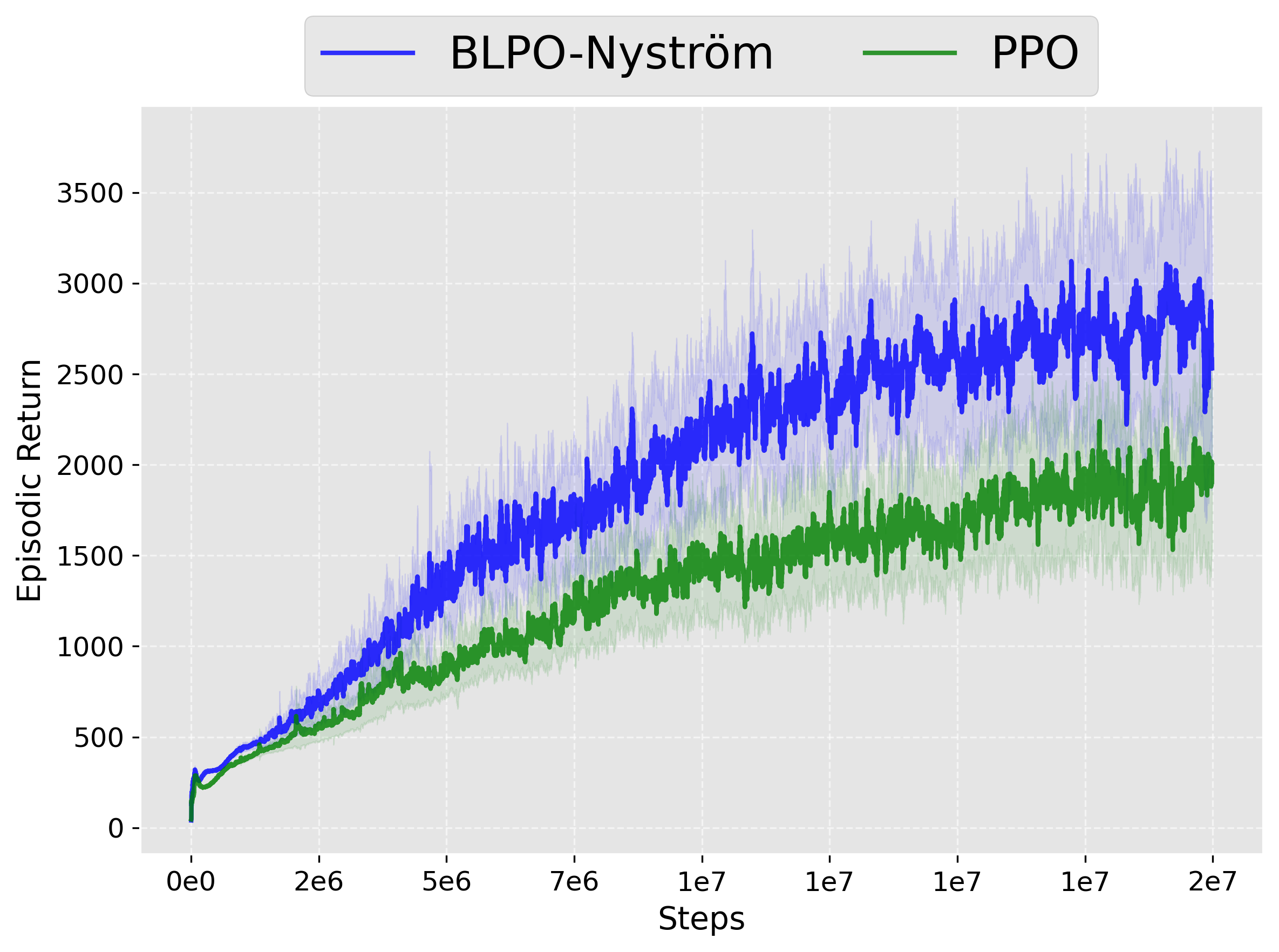}%
    } &
    \subcaptionbox{\scriptsize{Inverted Double Pendulum}\label{fig:inverted_double_Nystrom_vanilla}}[\panelW]{%
      \includegraphics[width=\linewidth,height=\panelH,keepaspectratio]{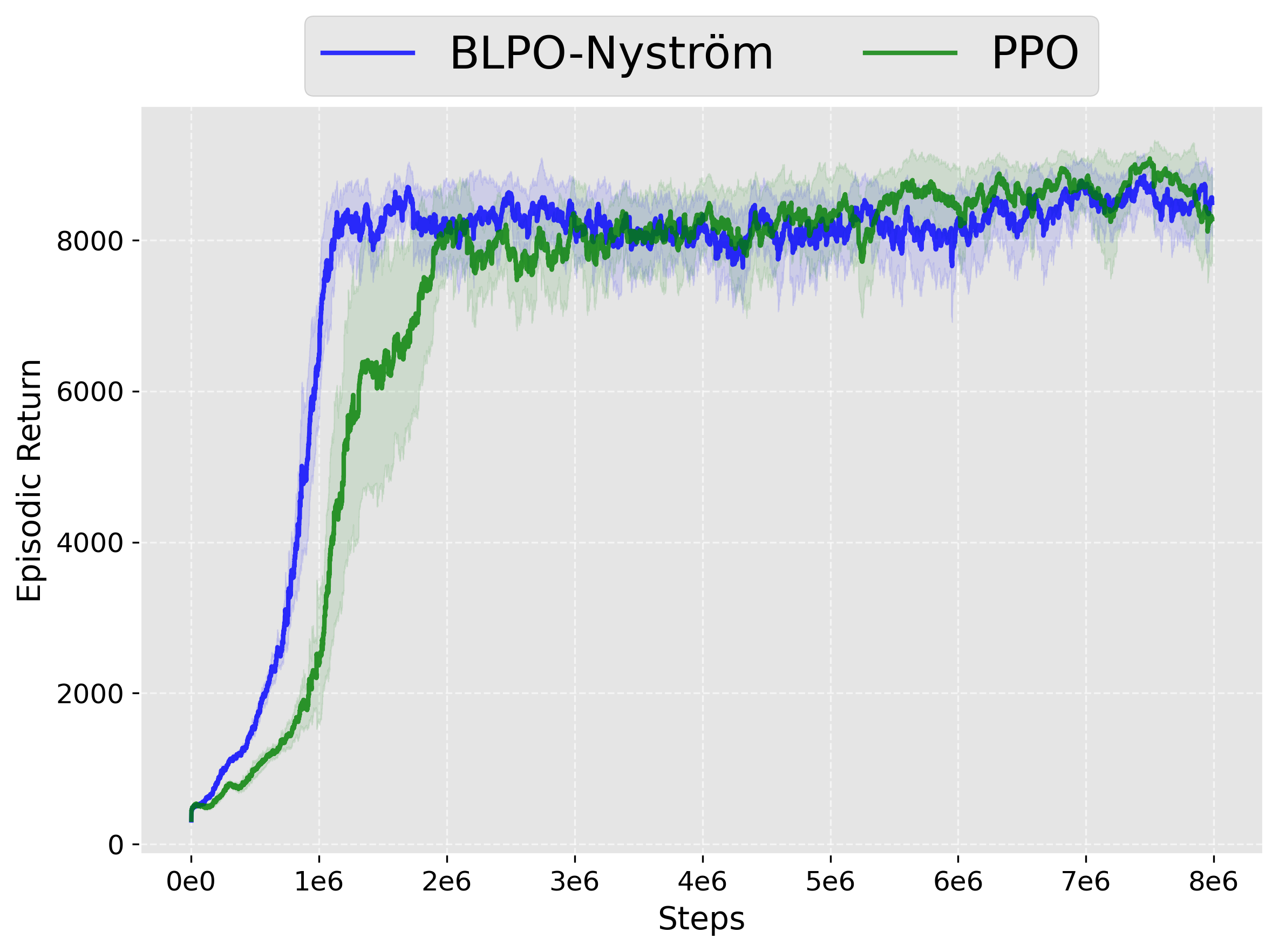}%
    } &
    \subcaptionbox{Hopper\label{fig:hopper_Nystrom_vanilla}}[\panelW]{%
      \includegraphics[width=\linewidth,height=\panelH,keepaspectratio]{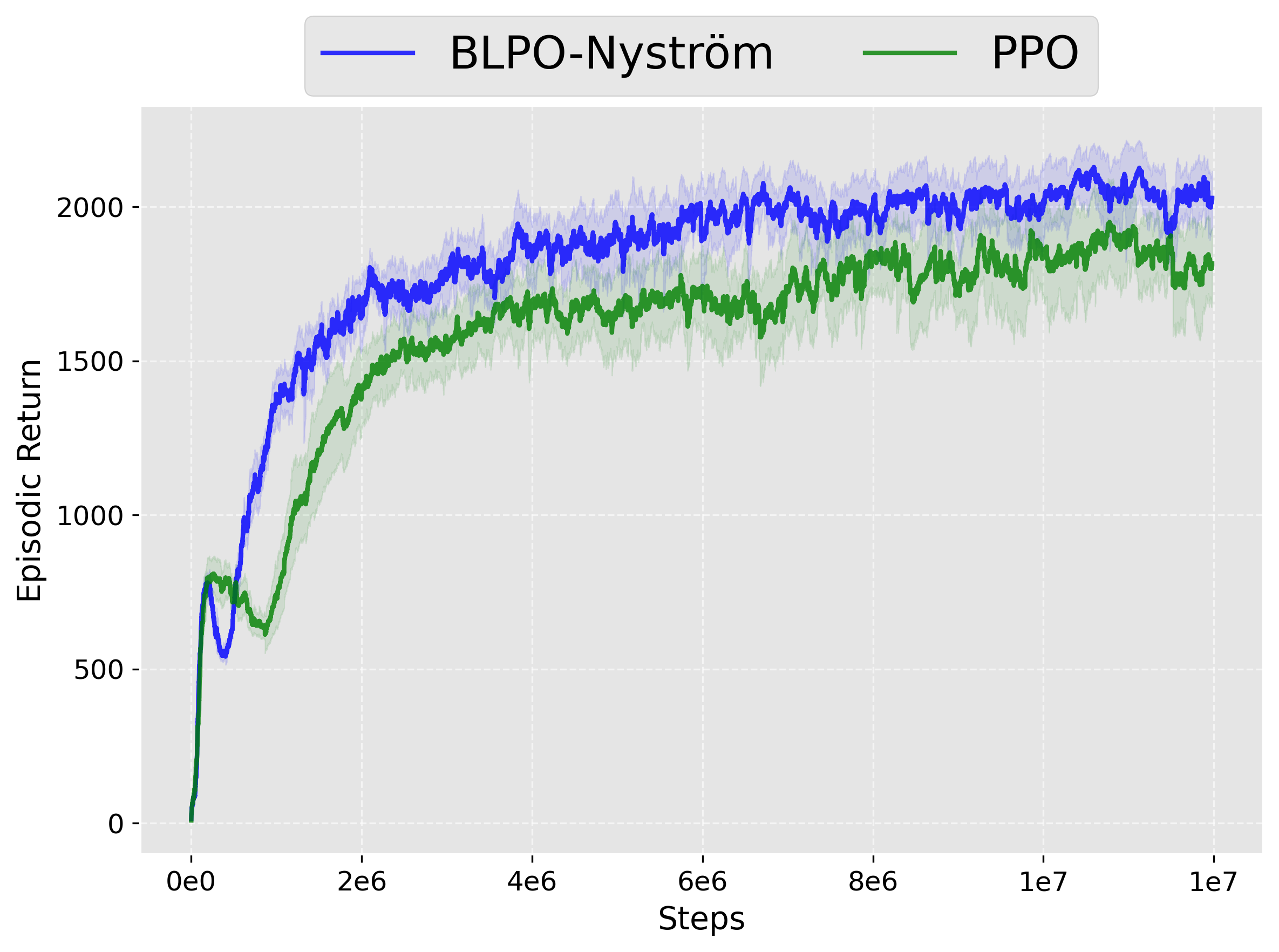}%
    } &
    \subcaptionbox{Cartpole\label{fig:discrete_cartpole}}[\panelW]{%
      \includegraphics[width=\linewidth,height=\panelH,keepaspectratio]{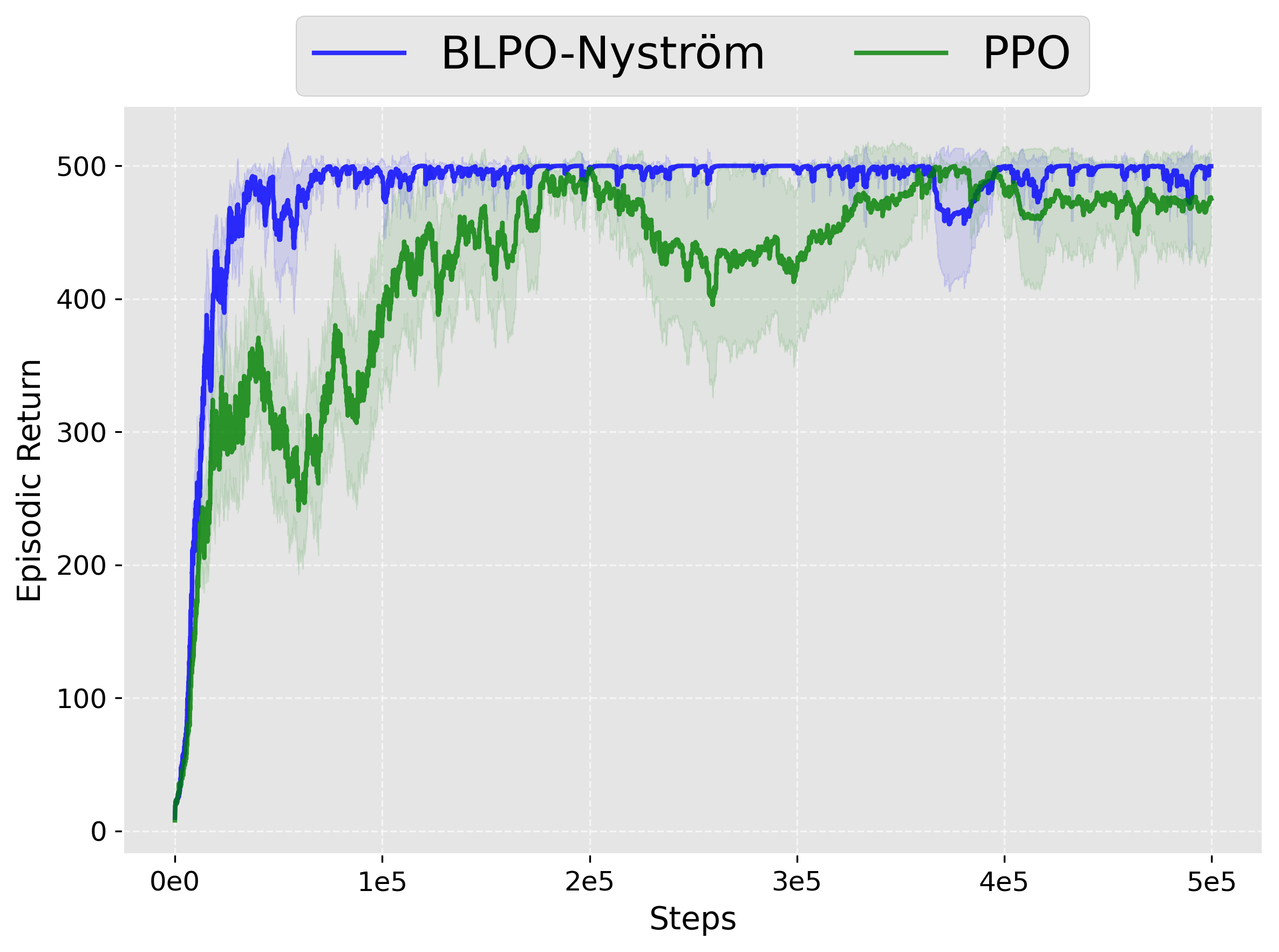}%
    } \\[0.35cm]

    \subcaptionbox{Inverted Pendulum\label{fig:inverted_pendulum_Nystrom_vanilla}}[\panelW]{%
      \includegraphics[width=\linewidth,height=\panelH,keepaspectratio]{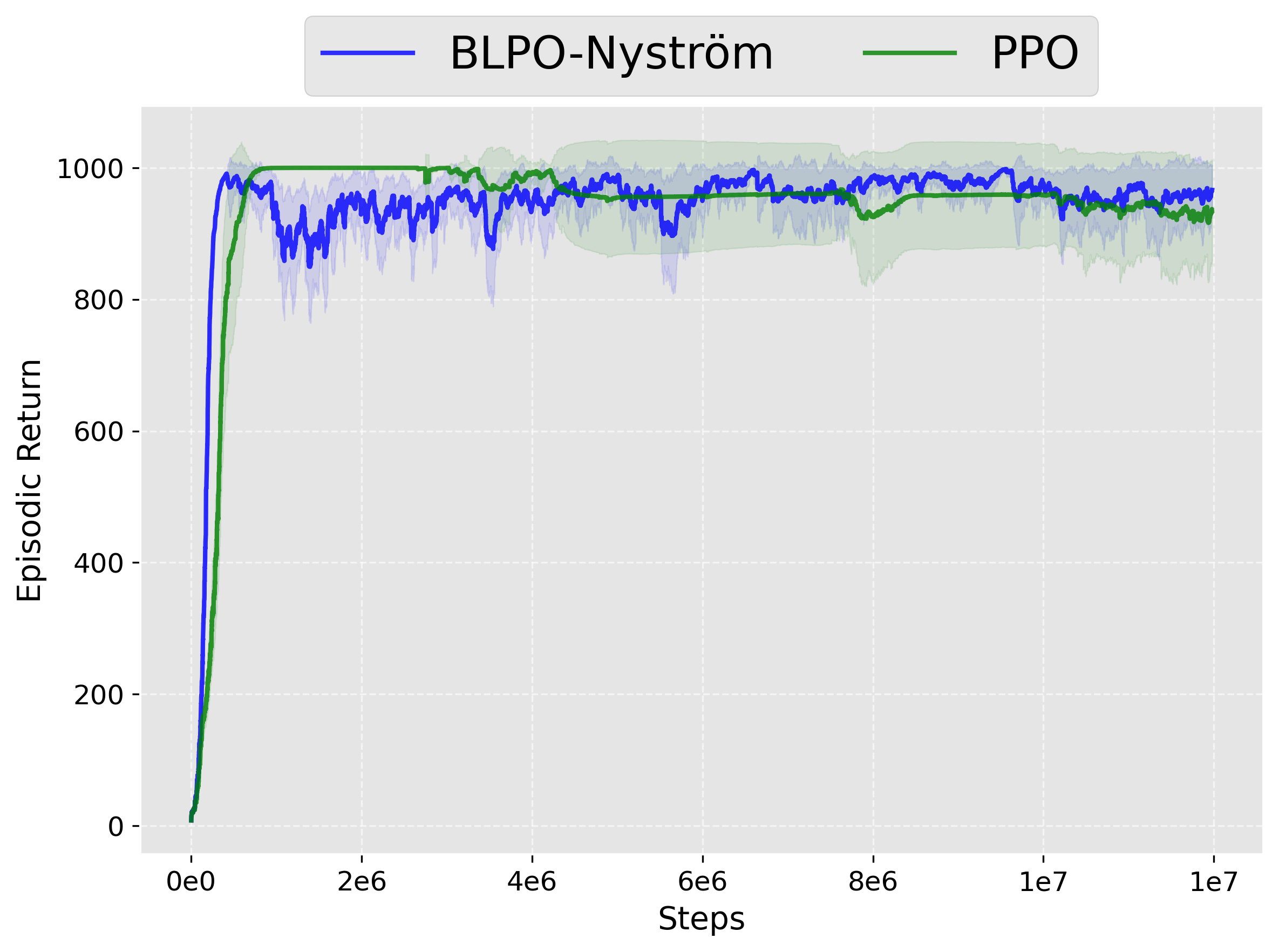}%
    } &
    \subcaptionbox{Humanoid Standup\label{fig:humanoid_Nystrom_vanilla}}[\panelW]{%
      \includegraphics[width=\linewidth,height=\panelH,keepaspectratio]{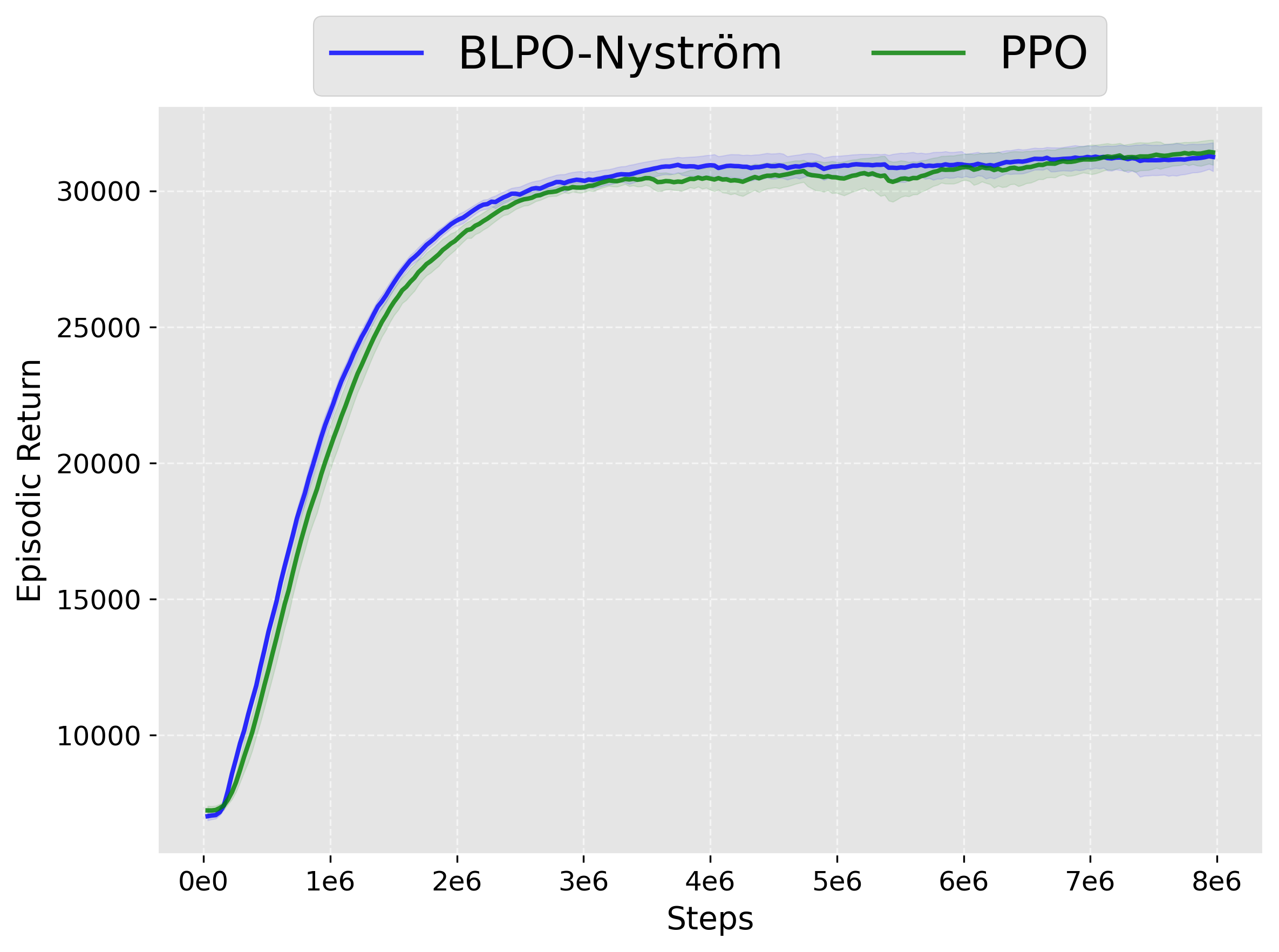}%
    } &
    \subcaptionbox{Pusher\label{fig:pusher_Nystrom_vanilla}}[\panelW]{%
      \includegraphics[width=\linewidth,height=\panelH,keepaspectratio]{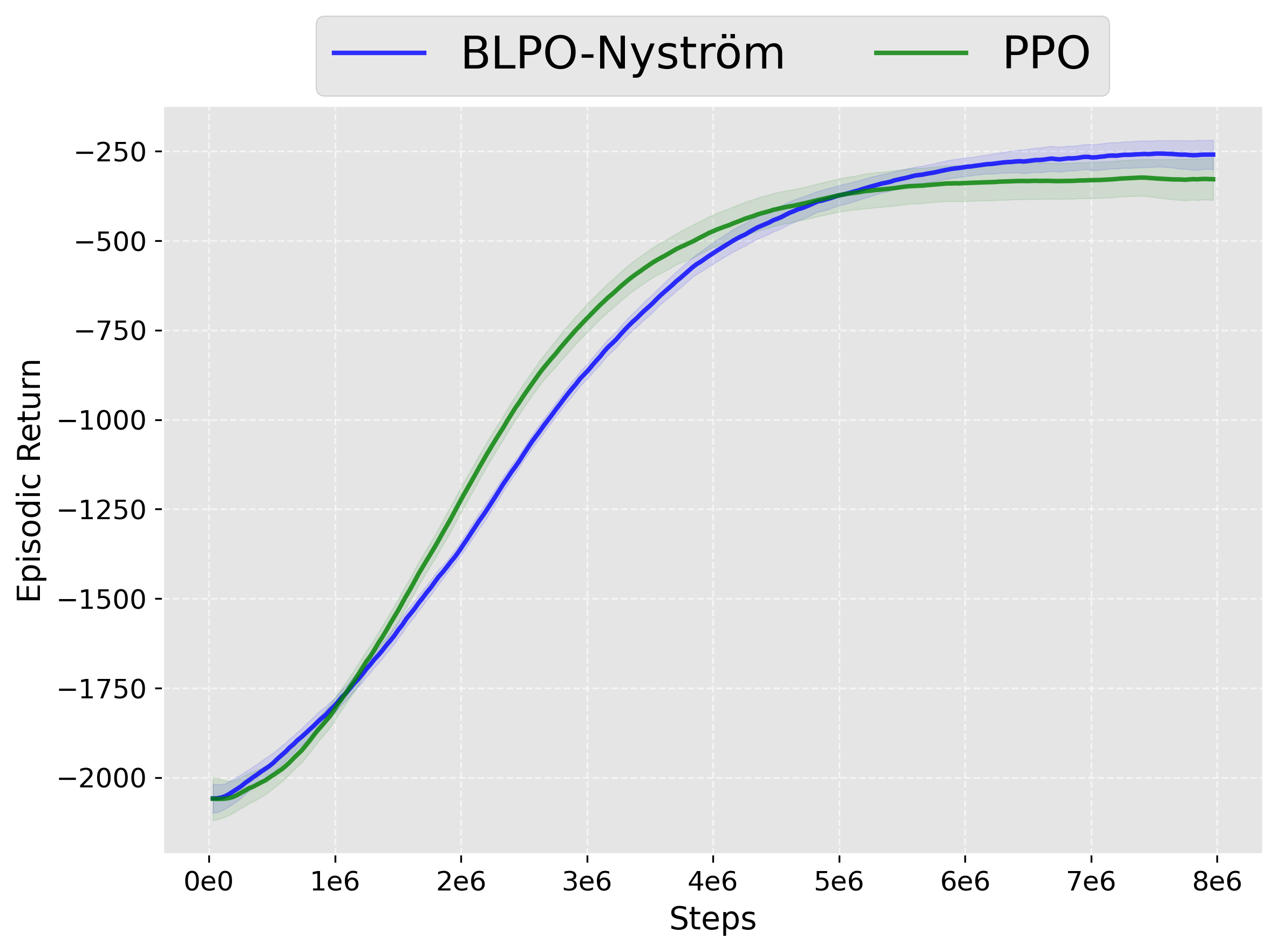}%
    } &
    \subcaptionbox{Acrobot\label{fig:discrete_acrobot}}[\panelW]{%
      \includegraphics[width=\linewidth,height=\panelH,keepaspectratio]{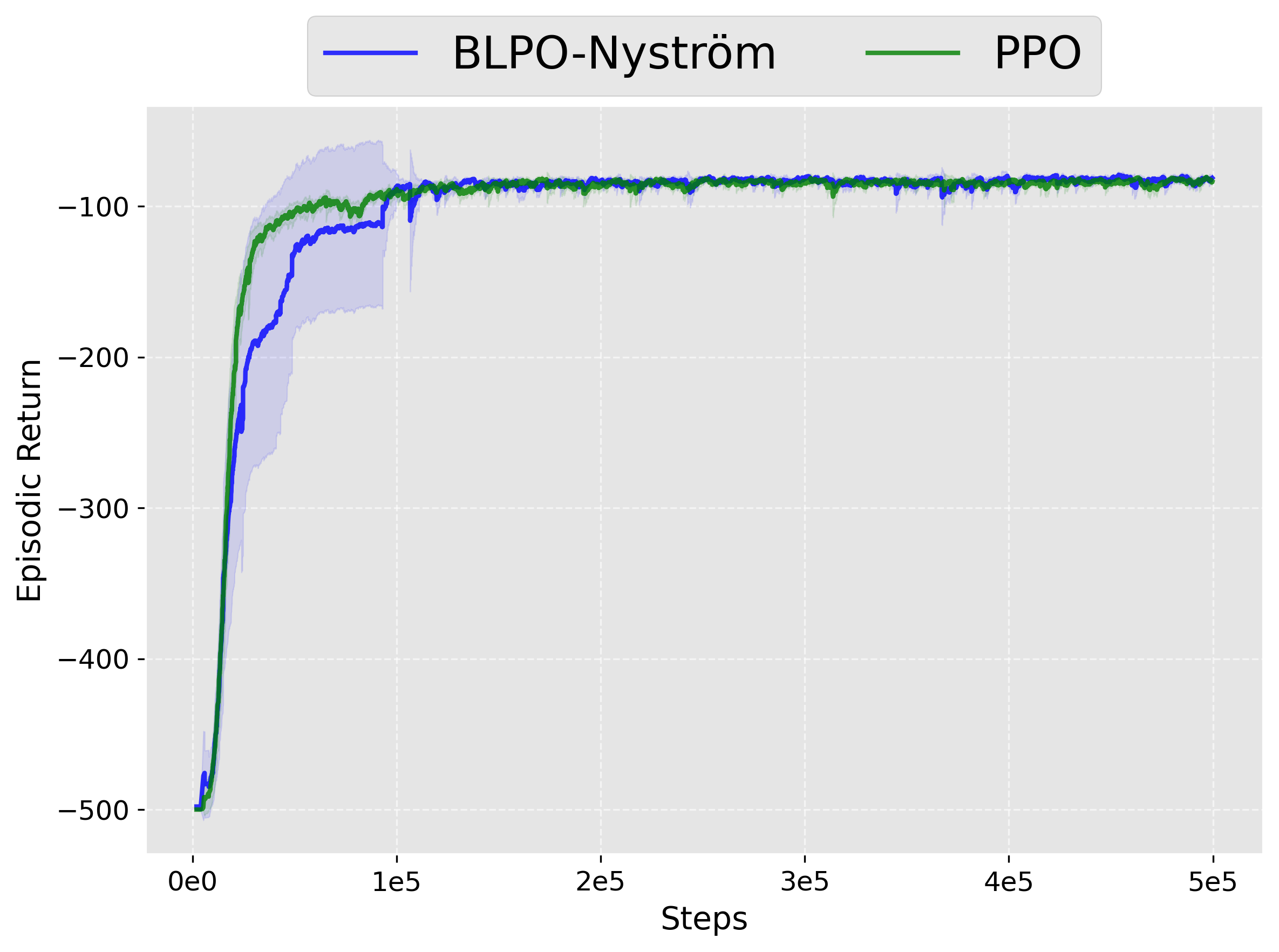}%
    } \\
  \end{tabular}

  \caption{
    In control tasks, \BLPO{} either outperforms PPO or performs comparably.
    The left three columns show continuous control experiments; the right column shows discrete control experiments.
  }
  \label{fig:brax_Nystrom_mixed}
\end{figure*}


We use as our baseline the PPO algorithm from PureJaxRL \cite{lu2022discovered}, with separate policy and value networks of fully-connected MLPs, each with two hidden layers of 64 units.
Our implementation of \BLPO---\BLPO{} Nystr\"om hereafter, specified in \Cref{alg:BLPO-Nystrom}---extends this implementation of PPO with a Nystr\"om-based hypergradient computation and nested critic updates. 
We also test another variant, \BLPO-CG, which uses CG to estimate the hypergradient instead of Nystr\"om's method.
In all cases, we calculate advantages using the Generalized Advantage Estimator (GAE) \cite{gae}.

We conduct experiments on discrete control tasks from Gymnax \cite{gymnax2022github} and continuous control tasks%
\footnote{In continuous control, we use a multivariate Gaussian to represent the policy network.}
from Brax \cite{brax2021github}.
All experiments were run using the default PureJaxRL PPO hyperparameters on a single RTX 3090 GPU.%
\footnote{See \appref{\Cref{apdx: experimental_details}} for additional experimental details.
Our code is available online at {\color{magenta} \url{https://github.com/Arnie-He/BLPO}}.}



\vspace{-2.5mm}
\paragraph{Heuristic}
The eigenspectrum of the Hessian has been found to have low empirical rank, with the bulk of the eigenspectrum clustered around zero with several large outliers \cite{sagun2016eigenvalues, ghorbani2019investigation}. 
As a result, we clip the IHVP to ensure that the inversion does not lead to an arbitrarily large gradient step, which would destabilize training.

\vspace{-2.5mm}
\paragraph{Performance}
In \Cref{fig:brax_Nystrom_mixed} we report 95\% confidence intervals around the average episodic returns across 15 seeds. 
We find that \BLPO{} outperforms PPO in most of our experiments and matches PPO in the rest. 
In Walker2D and Hopper, \BLPO{} learns a significantly better policy than PPO. 
Moreover, \BLPO{} converges with fewer samples than PPO in both Pendulum tasks. 
We also compare \BLPO, hereafter BLPO-Nystr\"om, to a variant that uses CG instead of Nystr\"om to estimate the hypergradient. 
We find that BLPO-Nystr\"om either outperforms or matches BLPO-CG in terms of episodic returns across all tasks.
We present these ablations, as well as nesting without any hypergradients {\color{magenta} \href{https://arxiv.org/abs/2505.11714}{here}}.

Performance parity on simpler tasks (e.g., Inverted Pendulum, Acrobot) reflects a ceiling effect: PPO already finds near-optimal solutions rapidly, rendering hypergradient information unnecessary.
Similarly, while Humanoid standup involves 376 parameters, its optimal configuration is a static hold with relatively non-chaotic dynamics.
BLPO's advantage emerges in complex, chaotic (Inverted Double Pendulum), and open-ended environments without a performance ceiling (Walker2d and Hopper), where actor-critic coupling is strong. 
In these settings, actor updates drastically shift the value landscape, causing PPO's performance, which mimics simultaneous updates, to degrade. 
BLPO's formulation, which allows the actor to anticipate the critic's response via the hypergradient, yields better policy updates.
In summary, BLPO should be the preferred choice when simulations are expensive, so that the reduced number of environment steps outweighs the per-step gradient overhead.

\vspace{-2.5mm}
\paragraph{Runtime Analysis}
Algorithms that rely on second-order information can run significantly slower than their first-order counterparts. 
On the simple control tasks, we find that \BLPO-Nyström is indeed slower than PPO.
In the more complex environments, however, the gap between all algorithms is smaller, as most of the computation time is spent on the simulator.
With regards to BLPO-CG, although the calculation of an \IHVP{} takes approximately 1.5 times a normal gradient step \cite{pearlmutter1994fast, dagréou2024howtocompute}, BLPO-CG is slower than BLPO-Nystr\"om  on all tasks, as poor conditioning necessitates more CG iterations.

\begin{figure}
    \centering
    \includegraphics[width=\linewidth]{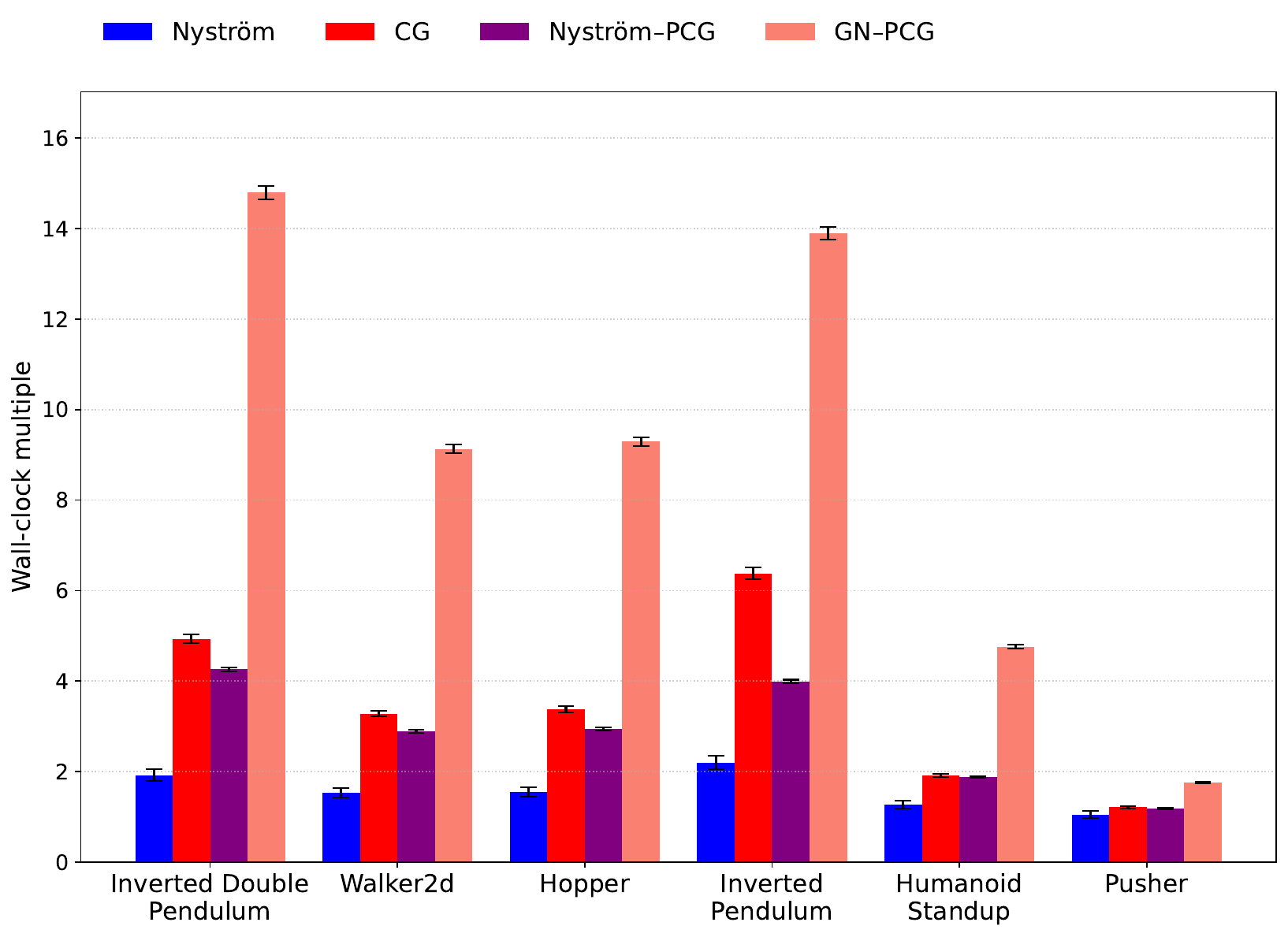}
    \caption{Wall-clock runtime of BLPO variants as a multiple of PPO's runtime (i.e., PPO corresponds to 1$\times$).}
    \label{fig:rel-runtime}
\end{figure}

We also experimented with BLPO variants using pre-conditioned conjugate gradient.%
\footnote{Empirically, we found 50 iterations maximized the performance of CG on in RL environments in our study, and we thus set the maximum number of CG iterations to 50.} 
Interestingly, the Nystr\"om method can itself be used as a preconditioner for conjugate gradient \cite{frangella2023randomized}.
We find that doing so improves the convergence time over unconditioned CG. 
We also use the Gauss-Newton matrix, a positive-semi-definite approximation of the Hessian; however, the additional overhead required to compute the Gauss-Newton matrix makes this method impractical. 
Our results, shown in \Cref{fig:rel-runtime}, demonstrate that the Nystr\"om method achieves the best balance of speed and performance.



%% file: sections/conc.tex
\section{Conclusion}
\label{sec:conc}

In this paper, we developed \BLO--Nystr\"om, an algorithm that solves a BLO using Nystr\"om's method to compute the hypergradient, and we established convergence rates for our algorithm to a 
local
stationary point of a non-convex strongly-convex BLO.
Our main contribution, \BLPO--Nystr\"om applies
these ideas to a BLO formulation of actor critic, which nests the critic's optimization inside the actor's, and follows the hypergradient of the actor.
A BLO formulation of AC is akin to a Stackelberg game, where it can be advantageous to be the leader: a Stackelberg leader (in our application, the actor) often attains greater rewards as the first-mover than she would in a simultaneous-move version of the same game \cite{blum2019computing}.

While the AC-BLO formulation is not new, \BLPO{} leverages the Nyström method to compute a low-rank approximation of the inverse Hessian-vector product, stabilizing the hypergradient computation.
Experimentally, we found that \BLPO{} consistently achieved stronger performance on a suite of discrete and continuous control tasks than PPO, the standard baseline. 
Future research should evaluate BLPO on a wider variety of tasks, such as Atari.
Future research is also needed to better evaluate its scalability to larger environments.
Successful scaling would render it applicable to other larger BLO, such as RLHF, meta-learning, and hyperparameter optimization.
This work takes a positive step towards paving the way for more efficient and scalable second-order optimization in deep learning.


%% file: sections/acknowledge.tex
\begin{acks}
This work was supported by the Office of Naval Research (ONR) grants N00014-22-1-2592 and N00014-24-1-2657 (institutional IDs GR5291318 and GR5250124, respectively).
We would also like to thank Jacob Makar-Limanov for investigating the low-rank structure of the Hessian in our experimental setup, and for his invaluable feedback on its implications for stabilizing the IHVP.
\end{acks}

%% file: appendix/Figures_and_Tables.tex
\clearpage
\section{Additional Figures and Tables}
\label{appx:figs_tables}
\subsection{Ablations}
\begin{figure*}[htbp]
  \centering
  \begin{subfigure}[b]{0.32\textwidth}
    \centering
    \includegraphics[width=\textwidth]{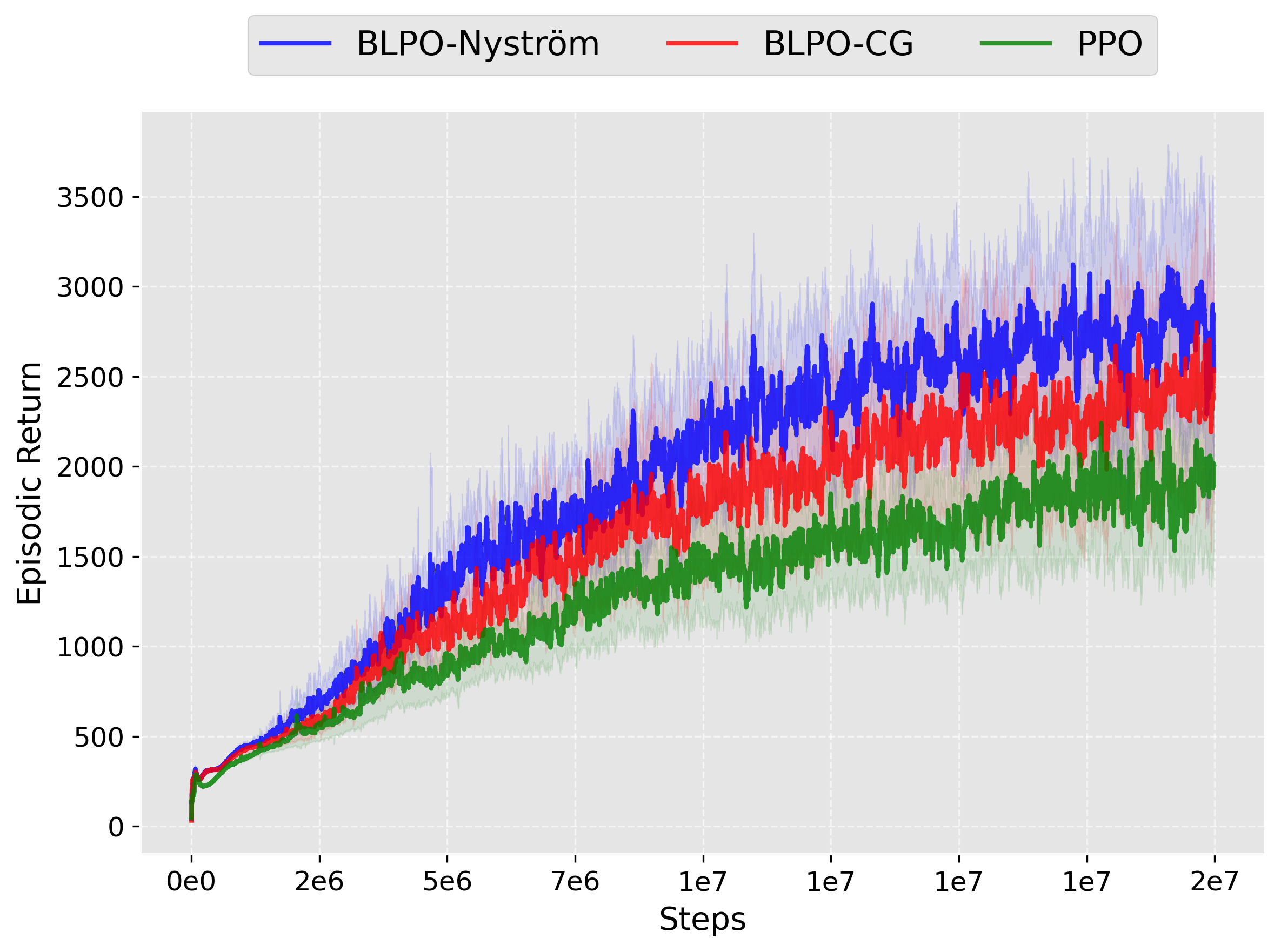}
    \caption{Walker2d}
    \label{fig:ablation_walker_cg}
  \end{subfigure}
  \hfill
  \begin{subfigure}[b]{0.32\textwidth}
    \centering
    \includegraphics[width=\textwidth]{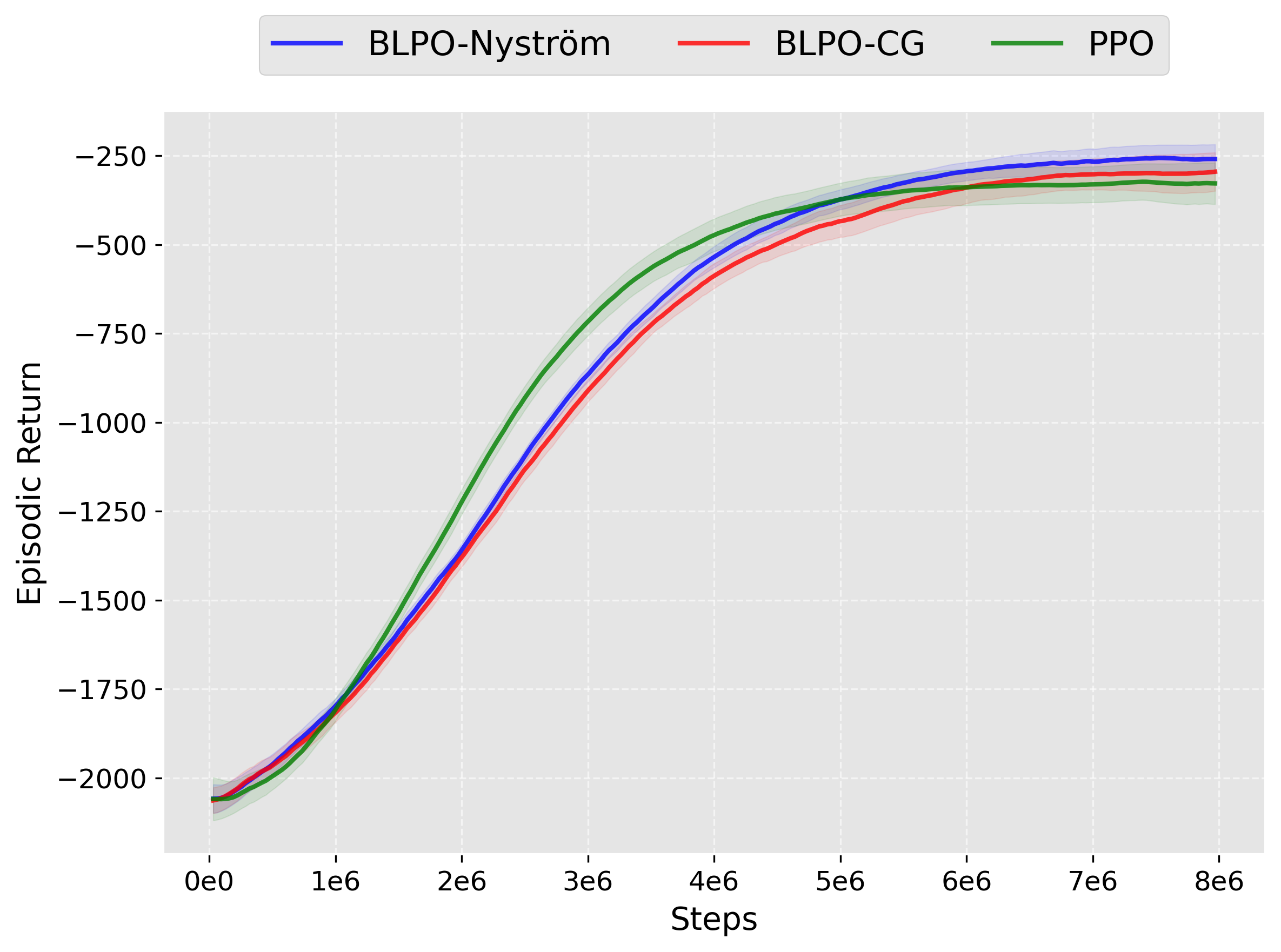}
    \caption{Pusher}
    \label{fig:ablation_inverted_double_Nystrom_cg}
  \end{subfigure}
  \hfill
  \begin{subfigure}[b]{0.32\textwidth}
    \centering
    \includegraphics[width=\textwidth]{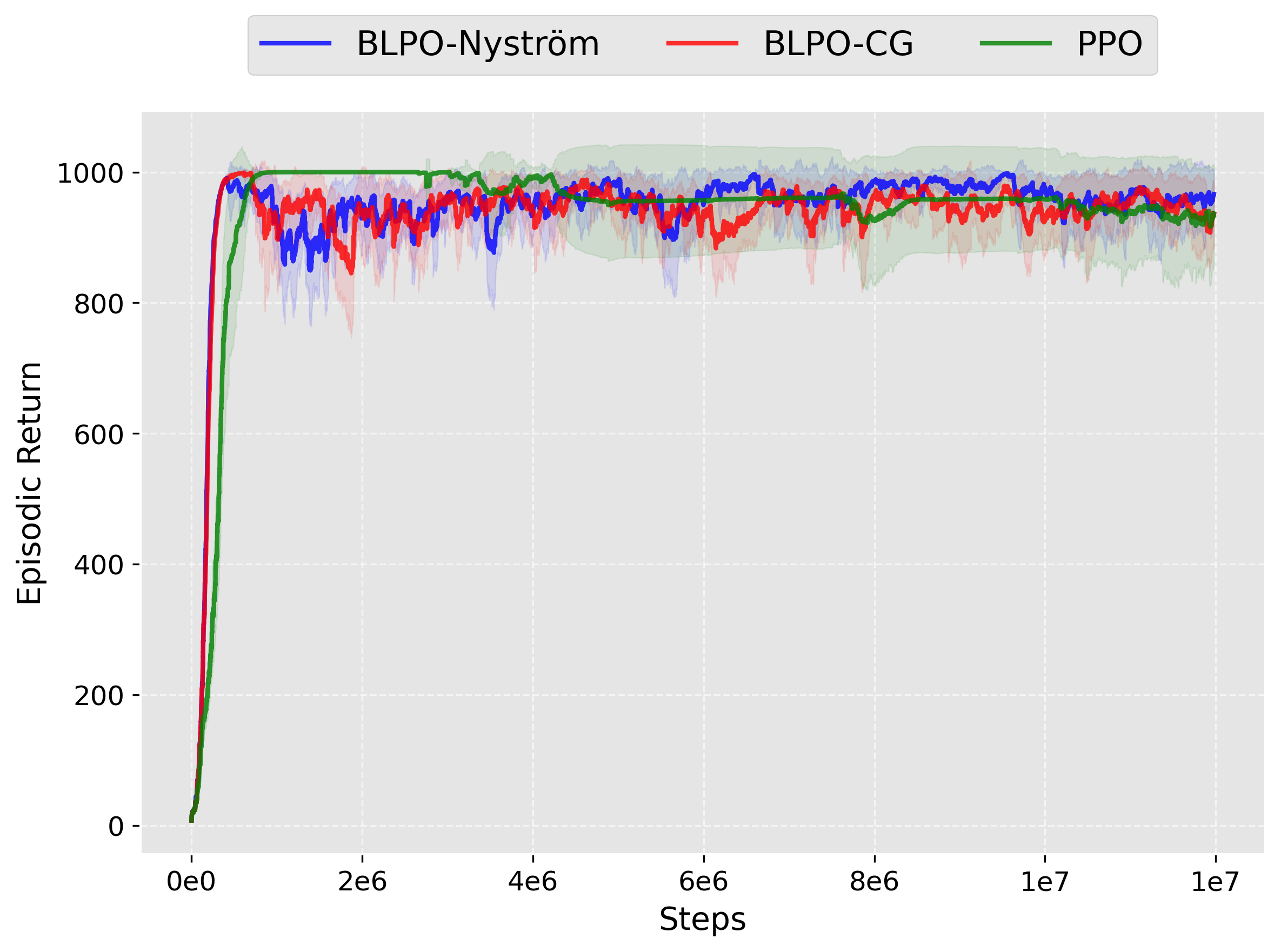}
    \caption{Inverted Pendulum}
    \label{fig:ablation_hopper_Nyström_cg}
  \end{subfigure}

  \caption{
    We conduct and ablations to show that the hypergradient with Nystr\"om method is key. Here we show examples where BLPO using conjugate gradient is more unstable and under performs the Nystr\"om method. Importantly, the Nystr\"om method never underperforms conjugate gradient in any of our other experiments. To control for speed we set the maximum number of CG iterations to 20. 
  }
  \label{fig:ablation-cg}
\end{figure*}

\begin{figure*}[htbp]
  \centering
  \begin{subfigure}[b]{0.32\textwidth}
    \centering
    \includegraphics[width=\textwidth]{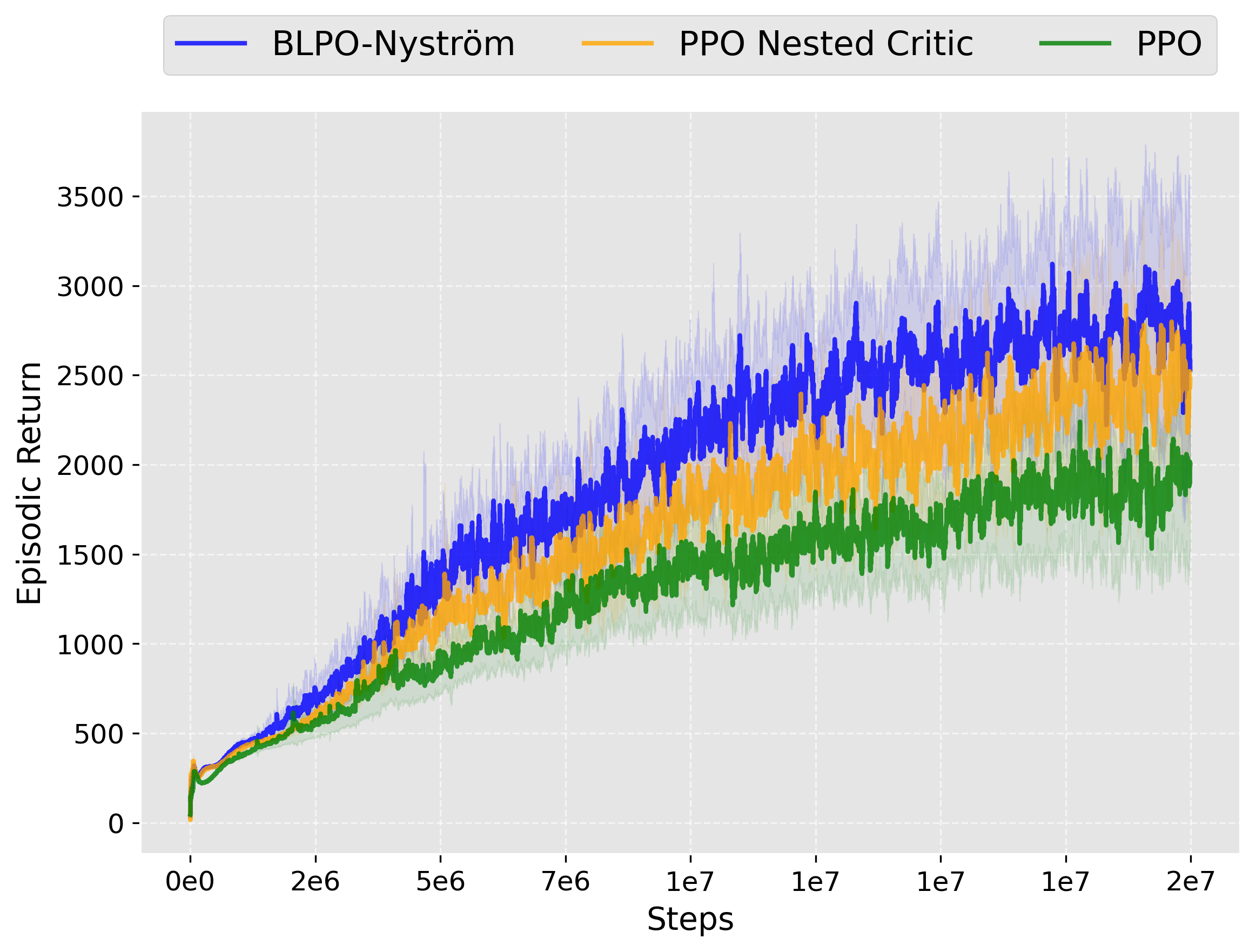}
    \caption{Walker2d}
    \label{fig:ablation_walker_nested}
  \end{subfigure}
  \hfill
  \begin{subfigure}[b]{0.32\textwidth}
    \centering
    \includegraphics[width=\textwidth]{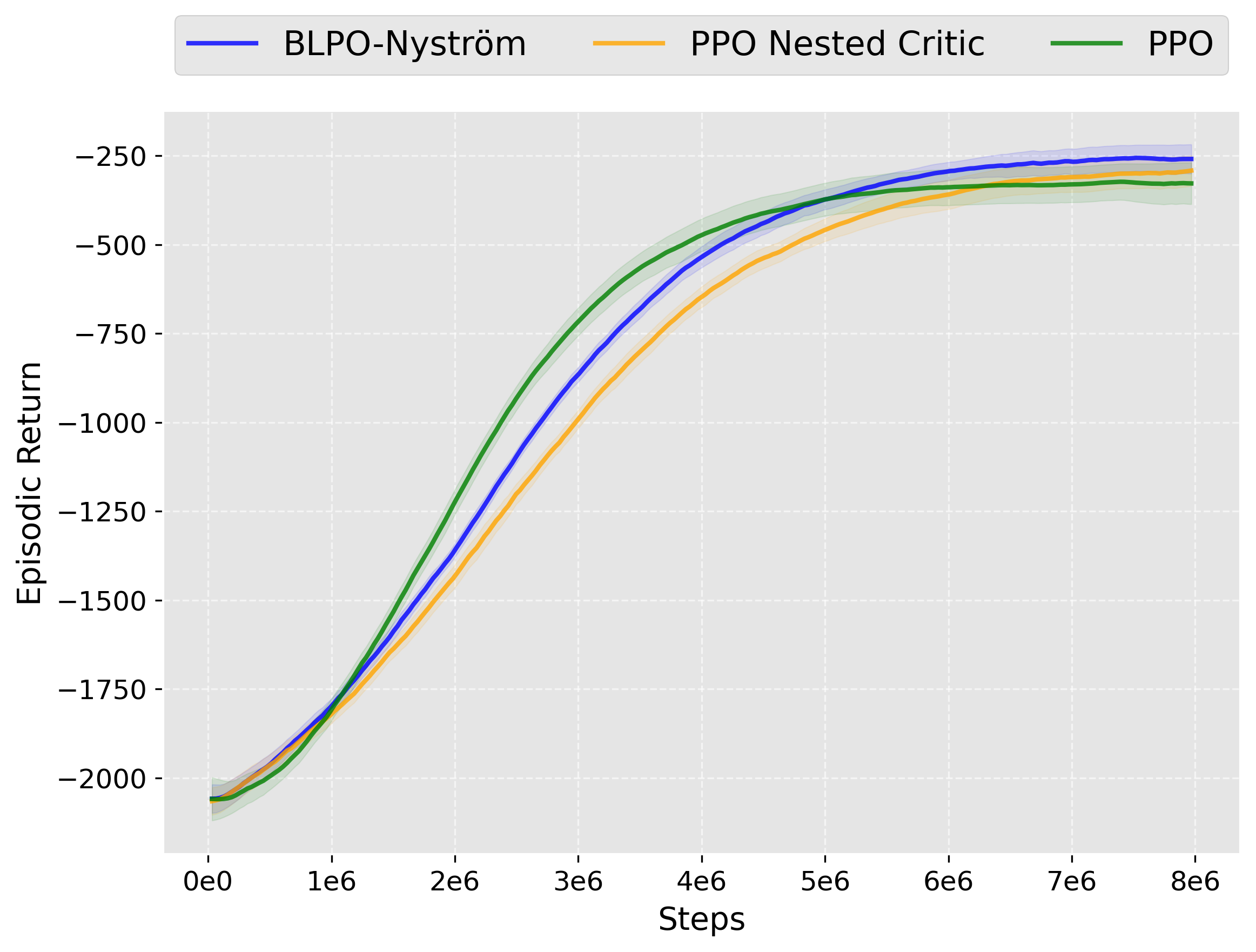}
    \caption{Pusher}
    \label{fig:ablation_inverted_double_Nystrom_nested}
  \end{subfigure}
  \hfill
  \begin{subfigure}[b]{0.32\textwidth}
    \centering
    \includegraphics[width=\textwidth]{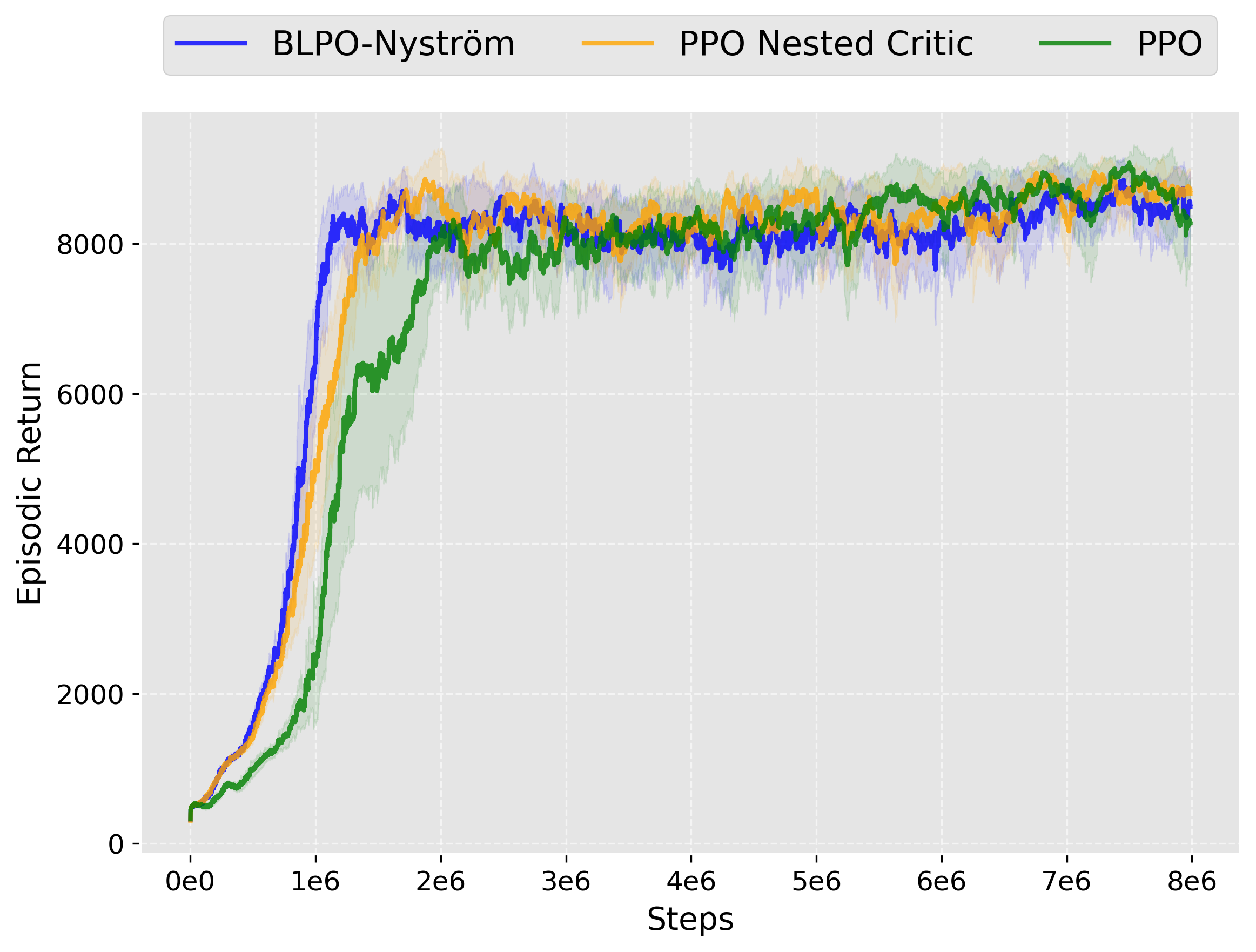}
    \caption{Inverted Double Pendulum}
    \label{fig:ablation_hopper_Nyström_nested}
  \end{subfigure}

  \caption{
    Here we show examples of where nesting the critic without the hypergradient is not enough. The Nystr\"om method never under performs nesting in any of our other experiments.   
  }
  \label{fig:ablation-nested}
\end{figure*}

\begin{figure*}[htbp]
  \centering
  \begin{subfigure}[b]{0.32\textwidth}
    \centering
    \includegraphics[width=\textwidth]{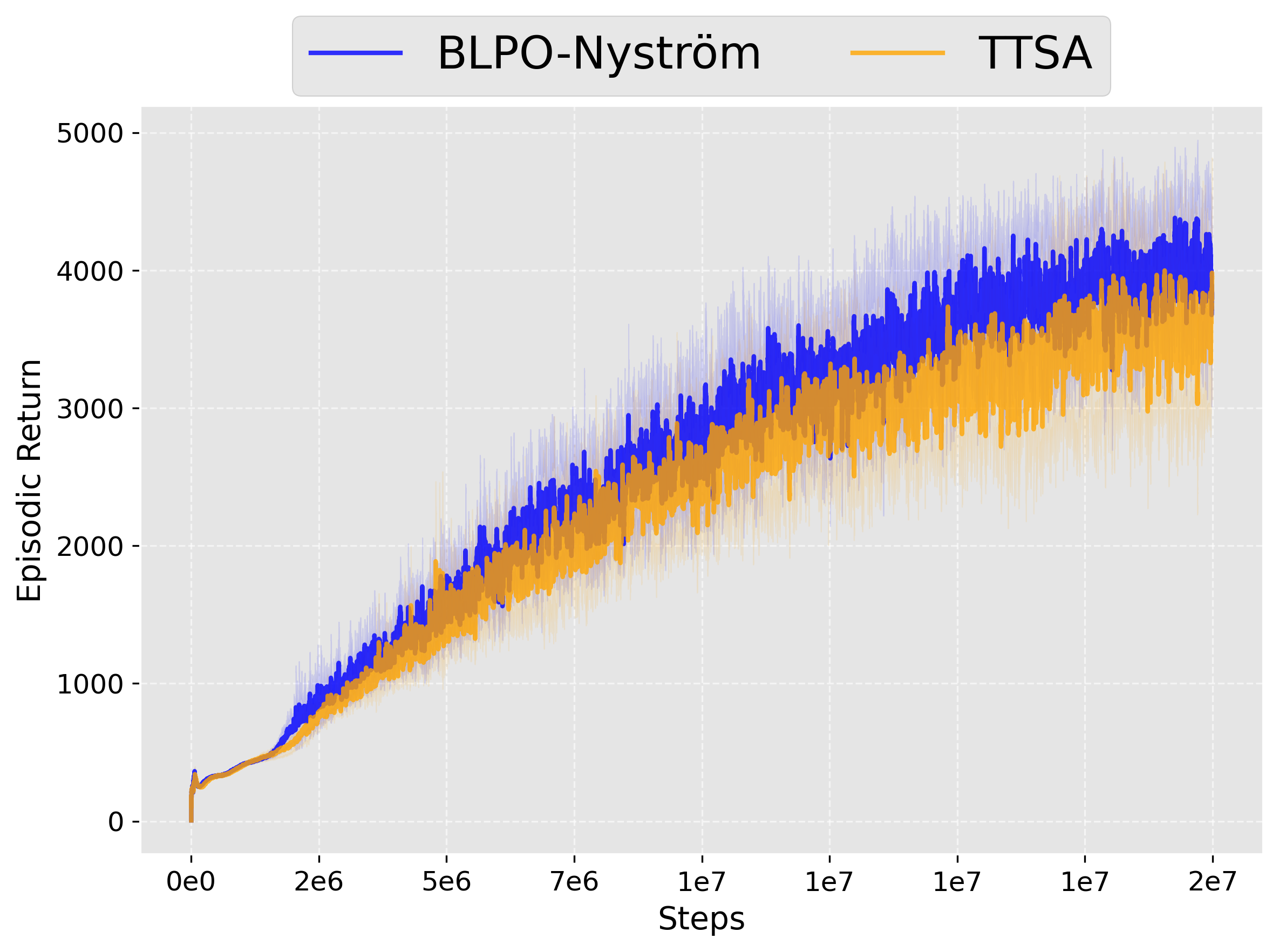}
    \caption{Walker2d}
    \label{fig:ablation_walker_ttsa}
  \end{subfigure}
  \hfill
  \begin{subfigure}[b]{0.32\textwidth}
    \centering
    \includegraphics[width=\textwidth]{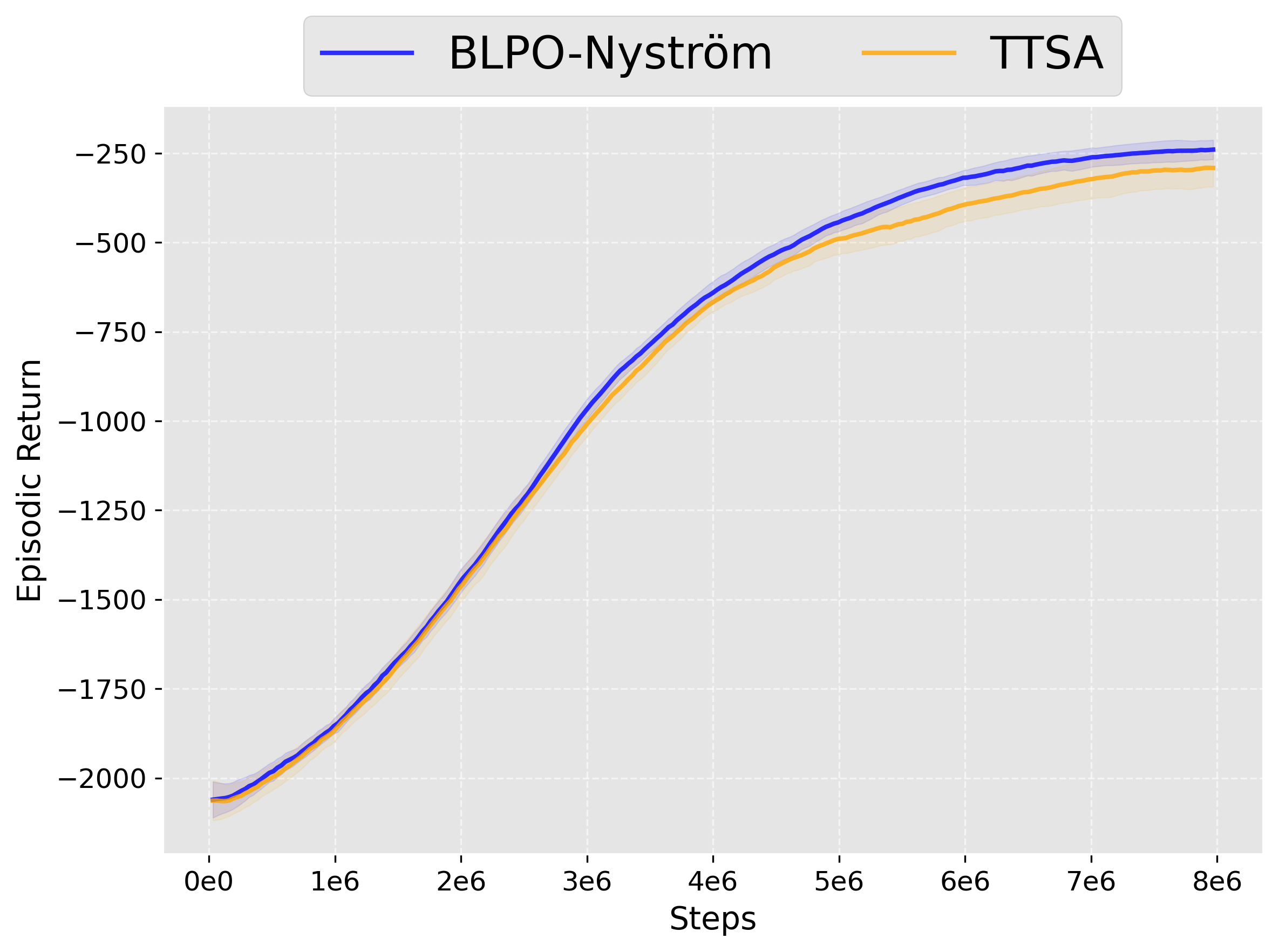}
    \caption{Pusher}
    \label{fig:pusher_ttsa}
  \end{subfigure}
  \hfill
  \begin{subfigure}[b]{0.32\textwidth}
    \centering
    \includegraphics[width=\textwidth]{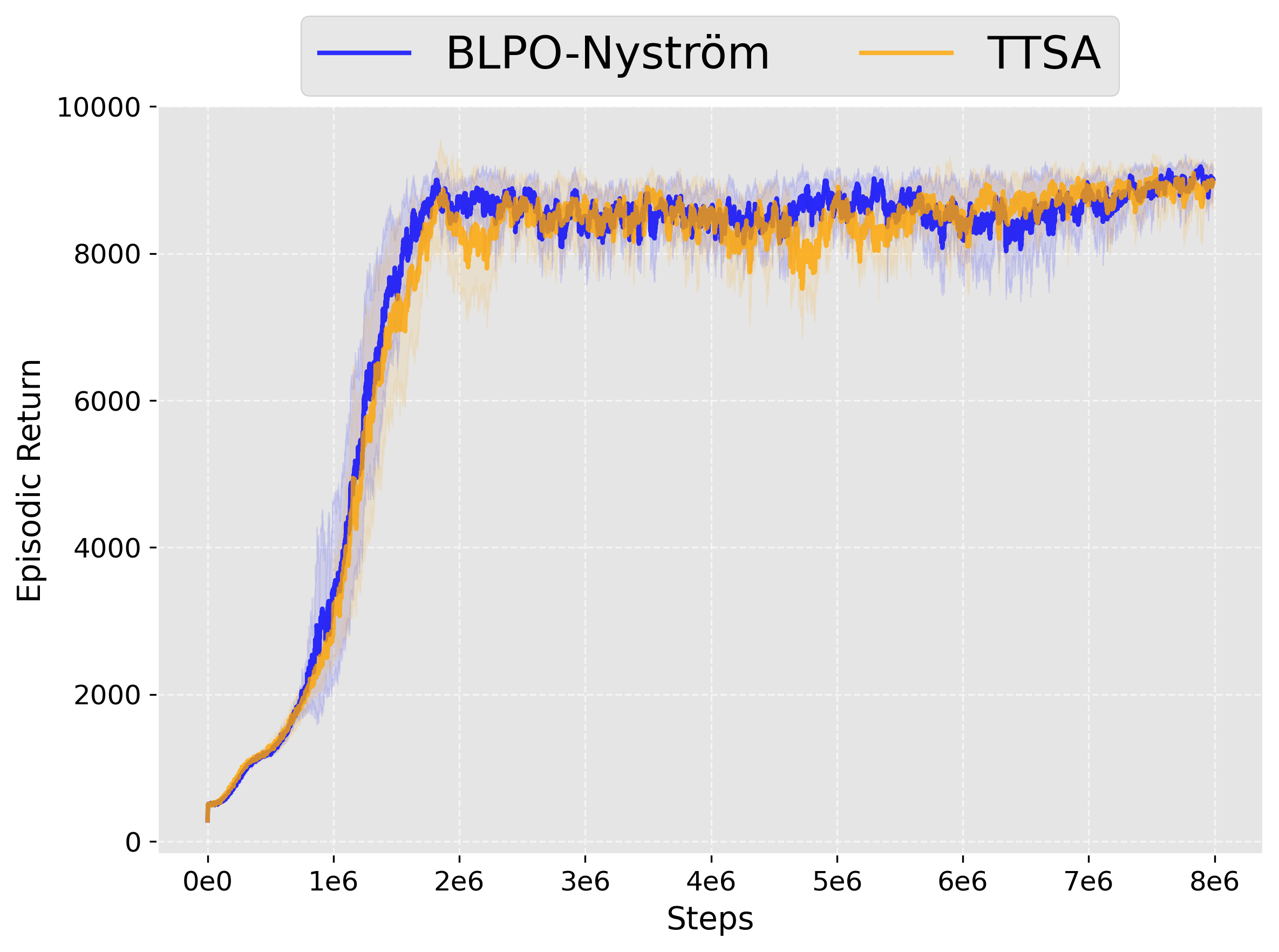}
    \caption{Inverted Double Pendulum}
    \label{fig:IDP_ttsa}
  \end{subfigure}
  \caption{
    Here we show that removing nesting, but having two separate learning rates for the actor and critic is not enough (two-time scale, TTSA).
  }
  \label{fig:ablation-nested}
\end{figure*}

%% file: appendix/Experimental_Details.tex
\clearpage
\section{Experimental Details}

\label{apdx: experimental_details}

We adapt our baseline PPO algorithm from the PureJaxRL\cite{lu2022discovered}, with separate policy network and value networks of fully-connected MLP with two hidden layers of 64 units. We use two separate. We extend the baseline PPO with our Nystrom based hypergradient and nest the critic updates. We use the same set of hyperparameters for all three of our algorithms (PPO, BL-Nystrom-PPO, and BL-CG-PPO) on all the environments. In practice, we sample columns for our Nystr\"om hypergradient uniformly at random.

The difference between our set-up for discrete control and continuous control is the number of paralel environments. For the ease of reproducibility, we provide a table of the hyperparameters we chose for all our algorithms in  \cref{table: hyperparam}. We run all our experiments with 15 seeds and take the average, and we shade the 95 confidence interval region.

Throughout our goal to build a fair and strong baseline and to further improve that with our proposed algorithm, we have had relied on \cite{shengyi2022the37implementation} for implementation details. We have deviated from all the 37 details of PPO discussed in this post from a few points, we will explain why, for the sake of fair comparison and easier reproducibility. 

\subsection{Value Function Loss Clipping}
We did not implement the value function loss clipping following the recommendations from \citet{shengyi2022the37implementation} which find that that value function loss clipping either does harm or no benefit to the overall performance of PPO.
PureJaxRL doesn't implement this function. Since our goal is to build a strong baseline instead of a high-fidelity reproduction, we ignore this trick. 

\subsection{Annealing the Learning Rate}
The original PPO paper implemented linear annealing on the learning rate, but \cite{andrychowicz2021what} discovered that it offers only a small performance boost. Since our nested algorithms have  different learning rates for the actor and critic we opt to simply not anneal the learning rates to keep the comparisons fair.



\subsection{Clipping and Bounding the hypergradient}

Inside $\grad[\actorparams] \criticobj[][\actorparams] (\criticparams)$ We clip $\log\policy[\actorparams](\action[\timestep] \mid \state[\timestep]) \adv[\timestep](\state[\timestep], \action[\timestep])$ in a similar way to the PPO actor objective with a separate $\epsilon$ threshold that we specify in the hyperparameter table as CLIP\_F. We also bound the entire hypergradient with IHVP\_BOUND times the norm of the first order actor gradient. These two hyperparameters are both held the same across the hypergradient algorithms(CG and Nystrom).

\begin{table}[t]
\caption{The hyperparamter table for all of our algorithms. Number in parathesis indicates the hyperparameter for discrete control vs continuous control }
\label{table: hyperparam}
\vskip 0.15in
\begin{center}
\begin{small}
\begin{sc}
\begin{tabular}{lcccr}
\toprule
Hyperparameter & PPO & BLPO-CG & BLPO-Nyström\\ 
\midrule
\textbf{NUM_ENVS} & 32(4) & 32(4) & 32(4) \\ 
\textbf{ROLLOUT_LEN} & 640(128) & 640(128) & 640(128) \\ 
\textbf{TOTAL_TIMESTEPS} & 8e6(5e5) & 8e6(5e5) & 8e6(5e5) \\
\textbf{NUM_MINIBATCHES} & 32(4) & 32(4) & 32(4) \\
\textbf{UPDATE_EPOCHS} & 4 & 4 & 4\\
\textbf{GAMMA} & 0.99 & 0.99 & 0.99\\
\textbf{GAE_LAMBDA} & 0.95 & 0.95 & 0.95\\
\textbf{CLIP_EPS} & 0.2 & 0.2 & 0.2 \\
\textbf{ENT_COEF} & 0.0 & 0.0 & 0.0 \\
\textbf{VF_COEF} & 0.5 & / & / \\ 
\textbf{ACTIVATION} & tanh & tanh & tanh \\
\textbf{ANNEAL_LR} & False (True) & / & / \\
\textbf{NORMALIZE_ENV} & True & True & True \\
\textbf{LR} & 2.5e-4 & / & / \\ 
\textbf{ACTOR_LR} & / & 2.5e-4 & 2.5e-4 \\
\textbf{CRITIC_LR} & / & 1e-3 & 1e-3 \\ 
\textbf{NESTED_UPDATES} & / & 3(10) & 3(10) \\ 
\textbf{IHVP_BOUND} & / & 1.0 & 1.0 \\
\textbf{CLIP_F} & / & 0.5 & 0.5 \\ 
\textbf{lambda_reg} & / & \textbf{0.0} & / \\ 
\textbf{MAX_CG_ITER} & / & \textbf{20} & / \\ 
\textbf{NYSTROM_RANK} & / & / & \textbf{5} \\
\textbf{NYSTROM_RHO} & / & / & \textbf{50} \\ 

\bottomrule
\end{tabular}
\end{sc}
\end{small}
\end{center}
\vskip -0.1in
\end{table}

%% file: appendix/additional_experiments.tex
\clearpage
\section{Additional Experiments}
\begin{figure*}[htbp]
  \centering
  \begin{subfigure}[b]{0.32\textwidth}
    \centering
    \includegraphics[width=\textwidth]{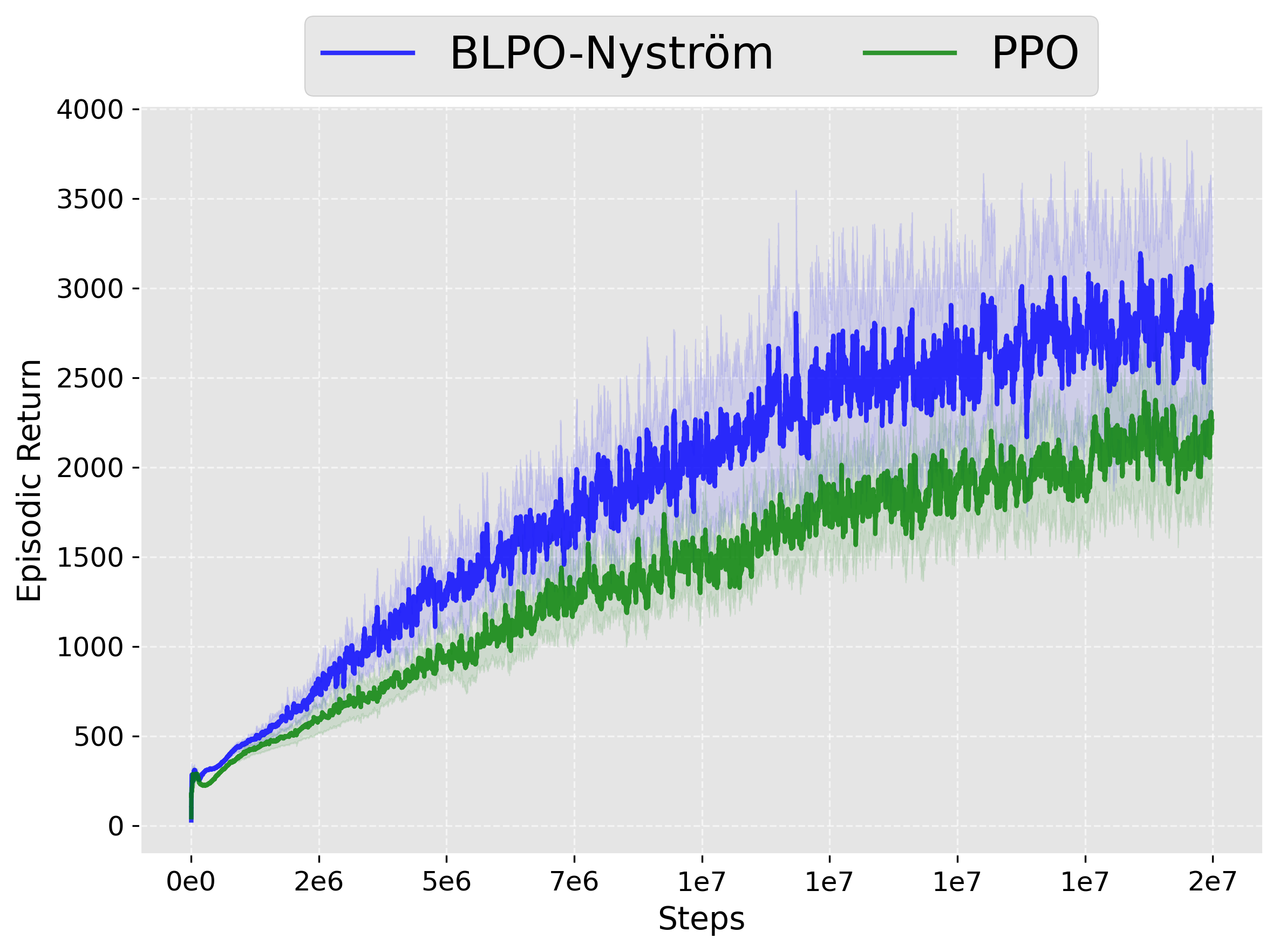}
    \caption{Walker2d}
    \label{fig:walker_3lr}
  \end{subfigure}
  \hfill
  \begin{subfigure}[b]{0.32\textwidth}
    \centering
    \includegraphics[width=\textwidth]{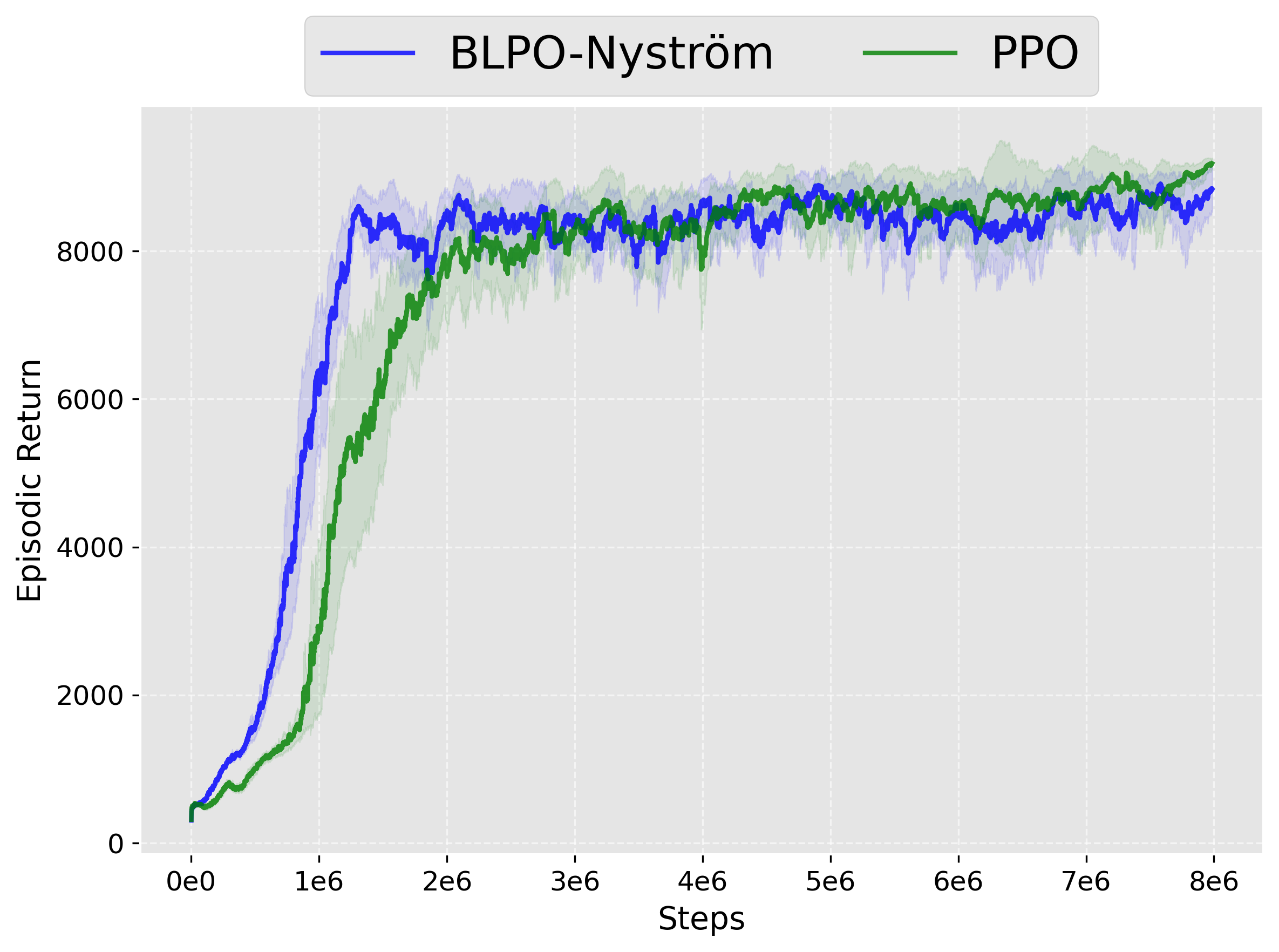}
    \caption{Inverted Double Pendulum}
    \label{fig:inverted_double_nystrom_vanilla}
  \end{subfigure}
  \hfill
  \begin{subfigure}[b]{0.32\textwidth}
    \centering
    \includegraphics[width=\textwidth]{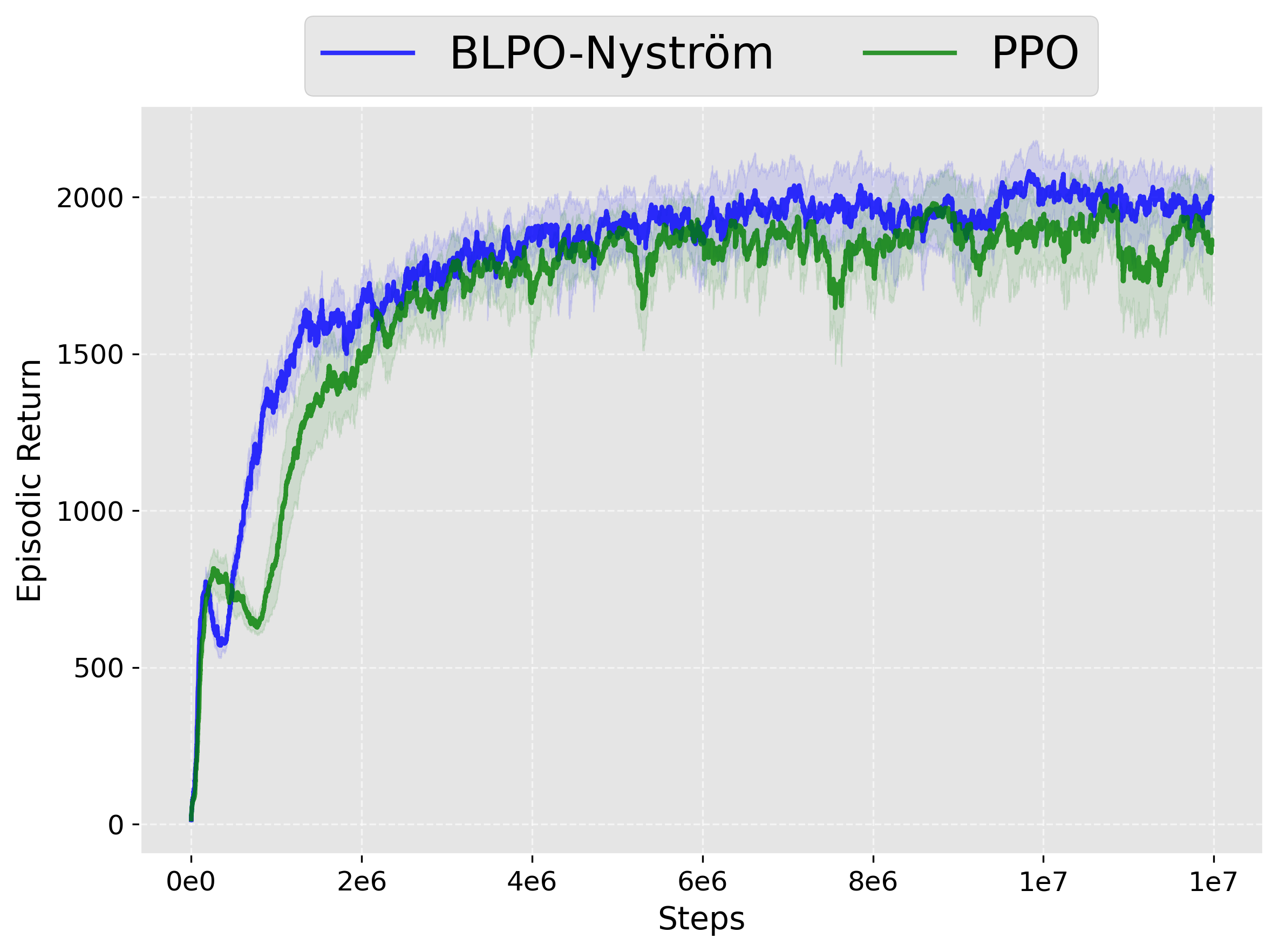}
    \caption{Hopper}
    \label{fig:hopper_3lr}
  \end{subfigure}
  \vspace{0.5cm}  

  \begin{subfigure}[b]{0.32\textwidth}
    \centering
    \includegraphics[width=\textwidth]{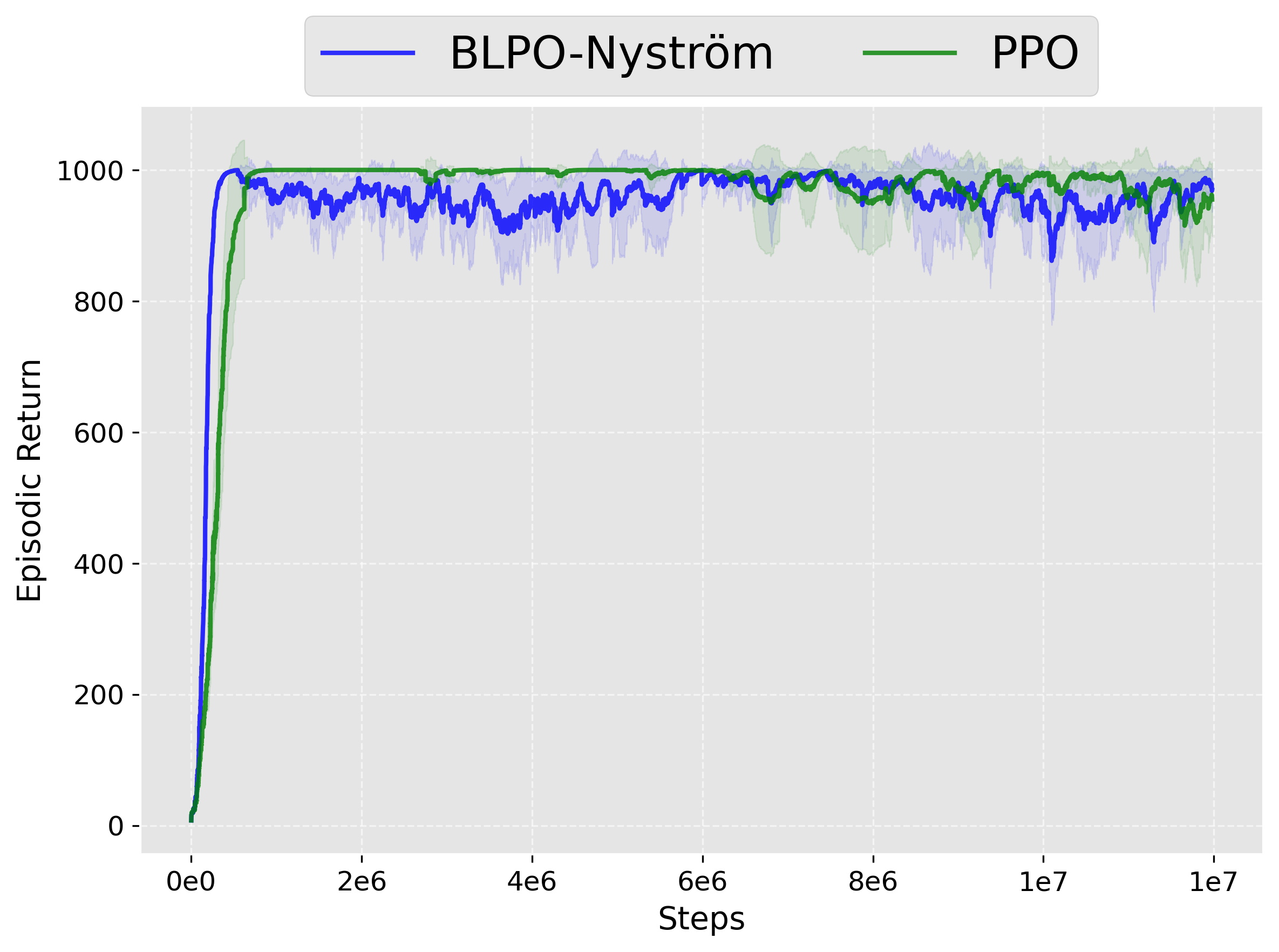}
    \caption{Inverted Pendulum}
    \label{fig:ip_3lr}
  \end{subfigure}
  \hfill
  \begin{subfigure}[b]{0.32\textwidth}
    \centering
    \includegraphics[width=\textwidth]{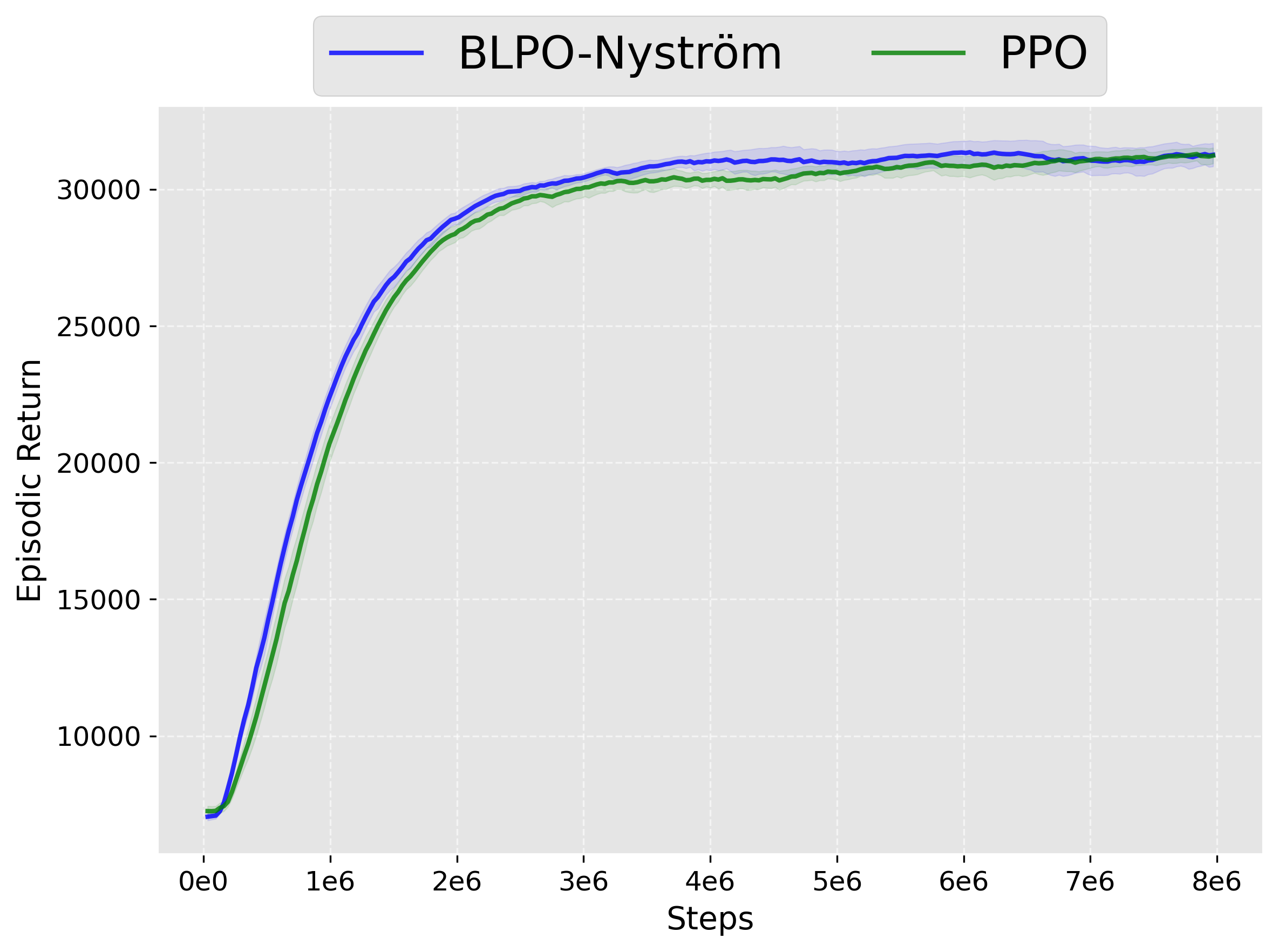}
    \caption{Humanoid Standup}
    \label{fig:humanoid_3lr}
  \end{subfigure}
  \hfill
  \begin{subfigure}[b]{0.32\textwidth}  
    \centering
    \includegraphics[width=\textwidth]{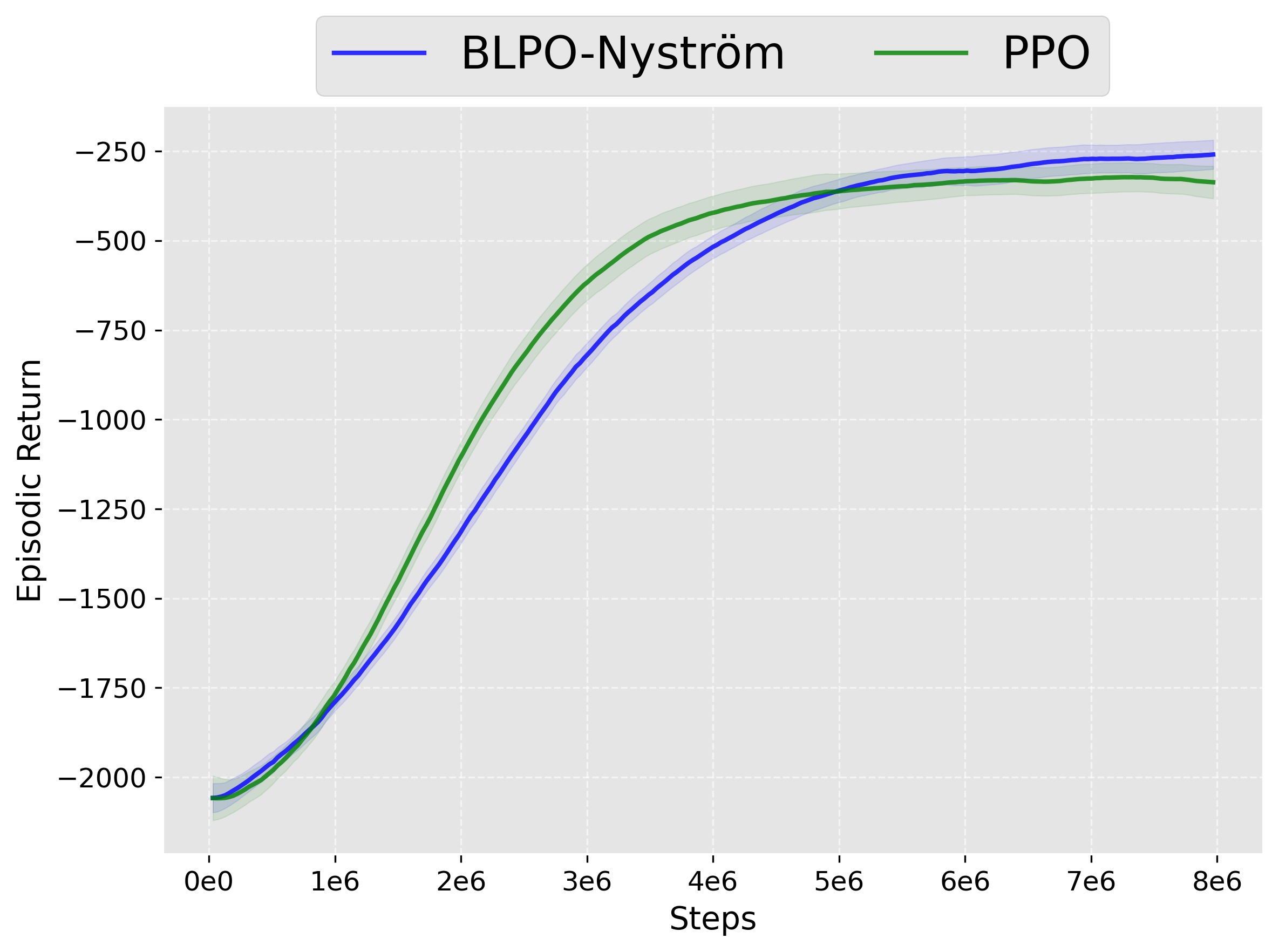}
    \caption{Pusher}
    \label{fig:pusher_3lr}
  \end{subfigure}

  \caption{
    We run additional experiments with a learning rate of 3e-4 as specified by PureJaxRL for continuous control. Our method still matches or outperforms base PPO. The hyperparameters are found in \cref{table:hyperparam_3lr}.
  }
  \label{fig:brax_3lr_3lr}
\end{figure*}

\begin{table}[t]
\label{table:hyperparam_3lr}
\caption{Hyperparameters for additional continuous control experiments}
\vskip 0.15in
\begin{center}
\begin{small}
\begin{sc}
\begin{tabular}{lcccr}
\toprule
Hyperparameter & PPO & BLPO-CG & BLPO-Nyström\\ 
\midrule
\textbf{NUM_ENVS} & 32 & 32 & 32 \\ 
\textbf{ROLLOUT_LEN} & 640 & 640 & 640 \\ 
\textbf{TOTAL_TIMESTEPS} & 8e6 & 8e6 & 8e6 \\
\textbf{NUM_MINIBATCHES} & 32 & 32 & 32 \\
\textbf{UPDATE_EPOCHS} & 4 & 4 & 4\\
\textbf{GAMMA} & 0.99 & 0.99 & 0.99\\
\textbf{GAE_LAMBDA} & 0.95 & 0.95 & 0.95\\
\textbf{CLIP_EPS} & 0.2 & 0.2 & 0.2 \\
\textbf{ENT_COEF} & 0.0 & 0.0 & 0.0 \\
\textbf{VF_COEF} & 0.5 & / & / \\ 
\textbf{ACTIVATION} & tanh & tanh & tanh \\
\textbf{ANNEAL_LR} & False & / & / \\
\textbf{NORMALIZE_ENV} & True & True & True \\
\textbf{LR} & 3e-4 & / & / \\ 
\textbf{ACTOR_LR} & / & 3e-4 & 3e-4 \\
\textbf{CRITIC_LR} & / & 1.2e-3 & 1.2e-3 \\ 
\textbf{NESTED_UPDATES} & / & 3 & 3 \\ 
\textbf{IHVP_BOUND} & / & 1.0 & 1.0 \\
\textbf{CLIP_F} & / & 0.5 & 0.5 \\ 
\textbf{lambda_reg} & / & \textbf{0.0} & / \\ 
\textbf{MAX_CG_ITER} & / & \textbf{20} & / \\ 
\textbf{NYSTROM_RANK} & / & / & \textbf{5} \\
\textbf{NYSTROM_RHO} & / & / & \textbf{50} \\ 

\bottomrule
\end{tabular}
\end{sc}
\end{small}
\end{center}
\vskip -0.1in
\end{table}

%% file: appendix/proofs.tex
\clearpage
\section{Omitted Proofs}
\label{appx:allproofs}
\subsection{Sketches and Warm-up}
\label{appx:warmup-proofs}
\input{appendix/proofs/sketch}
\input{appendix/proofs/warm_up_convergence}

\clearpage
\subsection{Convergence Proofs}
\label{appx:proofs}
\input{appendix/proofs/lemmas}
\input{appendix/proofs/convergence_proof}
\clearpage
\subsection{Reinforcement Learning Proofs}
\label{appx:rl_proofs}
\input{appendix/proofs/convex_rl}
\input{appendix/proofs/surrogate}

%% file: appendix/proofs/sketch.tex
\subsubsection{Proof Sketch \amy{why is this just a proof sketch, and not an actual proof?!} \arjun{You and NeurIPS suggest having a sketch of complex proofs} of the main result \Cref{thm:convergence}}

\Cref{thm:convergence} is structured around the smoothness properties of the hypergradient $\hyperobj$ and the strong convexity of $\innerobj$. First, we establish the gradient descent step:
\begin{align}
     \hyperobj(\outer[][][\iter+1]) &\leq \hyperobj(\outer[][][\iter]) + \langle \grad \hyperobj(\outer[][][\iter]), \outer[][][\iter+1] - \outer[][][\iter]\rangle + \frac{\hyperlip}{2}\Vert \outer[][][\iter+1] - \outer[][][\iter]\Vert^2 \\
     &\leq \hyperobj(\outer[][][\iter]) - \bigg(\frac{\learnrate[\outer]}{2} - \learnrate[\outer]^2 \hyperlip \bigg) \Vert \hyperobj(\outer[][][\iter])\Vert^2 + \bigg(\frac{\learnrate[\outer]}{2} + \learnrate[\outer]^2 \hyperlip \bigg)  \Vert \grad \hyperobj(\outer[][][\iter]) - \hat{\grad} \hyperobj(\outer[][][\iter])\Vert^2.
\end{align}

Then, using \Cref{lem:hyperbound} we are able to bound the following term with high probability:

\begin{align}
    \Ex [\Vert  \esthypergrad(\outer[][][\iter]) - \grad \hyperobj(\outer[][][\iter]) \Vert^2 ] \leq \Gamma  ( 1 - \learnrate[\inner] \strongconst)^{D}\Vert \inner[][0][\iter] - \inner[][][k] \Vert^2 + 2\firstlip^2\Psi^2,
\end{align}

where $\Psi$ is the error induced by the Nystr\"om method which holds with high probability using \Cref{lem:hvp-bound}. \Cref{ass:strongly-convex} establishes that $
\Gamma  ( 1 - \learnrate[\inner] \strongconst)^{\inneriters}\Vert \inner[][0][\iter] - \inner[][][k] \Vert \to 0$ as the number of inner iterations $\inneriters$ approaches $\infty$. We then use \cref{lem:combined-grad-bound} which incorporates warm-starts of the inner optimization to establish:

\begin{align}
\Vert \inner[][0][\iter] - \inner[][][k] \Vert^2 \leq \bigg( \frac{1}{2} \bigg)^k \Vert \inner[][0][0] - \inner[][][0] \Vert^2 + \sum^{\iter-1}_{i=0} \bigg(\frac{1}{2} \bigg)^{\iter-1-i}  \bigg[ 8 \condnum^2 \learnrate[\outer]^2 \firstlip^2\Psi^2 + 4 \condnum^2 \learnrate[\outer]^2 \Vert \hyperobj(\outer[][][i]) \Vert^2  \bigg].
\end{align}

Finally, using \cref{cor:hyperbound} and combining these above results and setting the appropriate constants, and telescoping to obtain:

\begin{align}
    \frac{1}{K}\sum^{K-1}_{\iter=0}  \Vert \hyperobj(\outer[][][\iter])\Vert^2 \leq \underbrace{\frac{64}{K} \bigg(\hyperlip (\hyperobj(\outer[][][0]) - \inf_{\outer} \hyperobj(\outer)) +  5\Vert \inner[][0][0] - \inner[][][0] \Vert^2\bigg)}_{\text{gradient descent error}} + \underbrace{10 \firstlip^2\Psi^2}_{\text{Nystr\"om error}} \leq \epsilon.
\end{align}

By choosing the number of inner iterations and appropriate column sampling regime for the Nystr\"om method yields our convergence result. 

%% file: appendix/proofs/warm_up_convergence.tex
\subsubsection{Warm-up}
Here we present a simplified version of our convergence result from \cref{thm:convergence}. In this case, we assume we have access to an inexact hypergradient: $\esthypergrad(\outer) = \grad\hyperobj(\outer) + \Psi$ and $\Vert \Psi \Vert \leq \sfrac{\epsilon}{2} $. Our main result then adjusts this proof to include the specifics of the Nystr\"om method.  
\begin{theorem}[Inexact Hypergradient]
    Assuming access to an inexact hypergradient oracle $\esthypergrad(\outer) = \grad\hyperobj(\outer) + \Psi$ and $\Vert \Psi \Vert \leq \sfrac{\epsilon}{2} $, and the inexact hypergradient is $\hyperlip$-Lipschitz continuous with $\hyperlip \in  \order(\condnum^3)$, then the number of gradient descent steps to achieve an $\epsilon$-stationary point is $\order(\frac{\condnum^3}{\sfrac{\epsilon}{2}})$.
\end{theorem}
\begin{proof}
Starting with the gradient descent rule:
\begin{align}
    \outer[][][\iter+1] &= \outer[][][\iter] - \learnrate[\outer] \esthypergrad(\outer[][][\iter]) \\
    &= \outer[][][\iter] - \learnrate[\outer][ \grad\hyperobj(\outer[][][\iter])+ \Psi].
\end{align}

Now we apply the fact that $\grad \hyperobj$ is $\hyperlip$-continuous, so we have:

\begin{align}
\hyperobj(\outer[][][\iter+1]) \leq \hyperobj(\outer[][][\iter])+ \langle \grad \hyperobj(\outer[][][\iter]), \outer[][][\iter+1] - \outer[][][\iter]\rangle + \frac{
\hyperlip}{2}\Vert \outer[][][\iter+1] - \outer[][][\iter] \Vert^2.
\end{align}

Combine this with $\outer[][][\iter+1] - \outer[][][\iter] = -\learnrate[\outer][\grad \hyperobj(\outer[][][\iter]) + \Psi]$. Now, applying the descent rule with the inexact gradient we obtain:

\begin{align}
\hyperobj(\outer[][][\iter+1]) &\leq \hyperobj(\outer[][][\iter]) + \langle \grad \hyperobj(\outer[][][\iter]), -\learnrate[\outer][\grad \hyperobj(\outer[][][\iter]) + \Psi]\rangle + \frac{
\hyperlip}{2}\Vert -\learnrate[\outer][\grad \hyperobj(\outer[][][\iter]) + \Psi]\Vert^2 \\
&= \hyperobj(\outer[][][\iter]) - \learnrate[\outer] \Vert \grad \hyperobj(\outer[][][\iter])\Vert^2 - \langle \learnrate \grad \hyperobj(\outer[][][\iter], \Psi\rangle + \frac{\hyperlip \learnrate[\outer]^2}{2} \Vert \grad \hyperobj(\outer[][][\iter]) + \Psi \Vert^2, 
\end{align}

which simplifies to by expanding and using the fact that $-2\langle a, b \rangle = -\Vert a \Vert^2 - \Vert b\Vert^2 + \Vert - b\Vert^2$

\begin{align}
    \hyperobj(\outer[][][\iter+1]) \leq \hyperobj(\outer[][][\iter]) - \bigg(\frac{\learnrate[\outer]}{2} - \learnrate[\outer]^2 \hyperlip\bigg) \Vert \grad \hyperobj(\outer[][][\iter]) \Vert^2 + \bigg(\frac{\learnrate[\outer]}{2} + \learnrate[\outer]^2 \hyperlip\bigg) \Vert \Psi\Vert^2. 
\end{align}

Choosing $\learnrate[\outer] \leq \frac{1}{8\hyperlip}$ 
\begin{align}
    \hyperobj(\outer[][][\iter+1]) \leq \hyperobj(\outer[][][\iter]) - \frac{3}{64\hyperlip}\Vert \grad \hyperobj(\outer[][][\iter]) \Vert^2+ \frac{5}{64\hyperlip}\Psi^2.
    \label{eq:easygrad}
\end{align}

Now, by the definition of telescoping over $\iter = 0,...,K-1$:

\begin{align}
    \sum^{K-1}_{k=0}[\hyperobj(\outer[][][\iter+1]) - \hyperobj(\outer[][][0])] = \hyperobj(\outer[][][K]) - \hyperobj(\outer[][][0]),
\end{align}

and back into \cref{eq:easygrad} we obtain:

\begin{align}
    \hyperobj(\outer[][][K]) - \hyperobj(\outer[][][0]) \leq -\frac{3}{64\hyperlip}\sum^{K-1}_{\iter=0} \Vert \grad \hyperobj(\outer[][][\iter])\Vert^2 + \frac{5}{64\hyperlip} \Vert\Psi\Vert^2.
\end{align}

Finally, rearranging to put $\grad \hyperobj(\outer[][][\iter])$ on the LHS, we obtain:

\begin{align}
    \frac{1}{K}\sum^{K-1}_{\iter=0} \Vert \grad \hyperobj(\outer[][][\iter])\Vert^2 \leq \underbrace{\frac{64\hyperlip}{K}\bigg(\hyperobj(\outer[][][0]) - \inf_{\outer} \hyperobj(\outer)\bigg)}_{\text{Gradient descent error}} + \underbrace{5 \Vert\Psi\Vert^2}_{\text{Inexact hypergradient error}} \leq \epsilon. 
\end{align}

For the gradient complexities, given that $\Vert \Psi \Vert \leq \sfrac{\epsilon}{2}$, it is sufficient to ensure that the gradient descent error is also less than $\sfrac{\epsilon}{2}$ to ensure that we achieve an $\epsilon$-stationary point.  Recall that $\hyperlip$ is $\order(\condnum^3)$, so the number of hypergradient iterations is $K \in \order(\frac{\condnum^3}{\sfrac{\epsilon}{2}})$.
\end{proof}




%% file: appendix/proofs/lemmas.tex
\subsubsection{Supporting Lemmas}
\warmstarts*  
\begin{proof}
Consider the optimality condition $\grad[\inner]\innerobj[][\outer](\inner)= 0$ and $\grad[\inner]\innerobj[][\outerpoint](\innerpoint)=0$. 
\begin{align}
    \grad[\inner]\innerobj[][\outer](\inner)- \grad[\inner]\innerobj[][\outerpoint](\innerpoint) &= 0\\
    \underbrace{\grad[\inner]\innerobj[][\outer](\inner)- \grad[\inner]\innerobj[][\outer](\innerpoint)}_{\Delta_1} + \underbrace{\grad[\inner]\innerobj[][\outer](\innerpoint) - \grad[\inner]\innerobj[][\outerpoint](\innerpoint)}_{\Delta_2} &= 0.
\end{align}

Combining \cref{ass:strongly-convex} with the fact that $\Delta_1 = -\Delta_2$, we obtain:
\begin{align}
    \langle \grad[\inner]\innerobj[][\outer](\inner)- \grad[\inner]\innerobj[][\outer](\innerpoint), \inner - \innerpoint \rangle &\geq \strongconst \Vert \inner - \innerpoint\Vert^2 \\\langle \Delta_1, \inner - \innerpoint \rangle &\geq \strongconst \Vert \inner - \innerpoint\Vert^2
    \\
    \langle -\Delta_2, \inner - \innerpoint \rangle &\geq \strongconst \Vert \inner - \innerpoint\Vert^2.
\end{align}

By \cref{ass:first-order-lip}:
\begin{align}
    \Vert \Delta_2 \Vert \leq \firstlip \Vert \outer - \outerpoint \Vert,
\end{align}

which we can combine, and then apply apply Cauchy-Schwartz to obtain:
\begin{align}
    \Vert \Delta_2 \Vert \cdot \Vert \inner - \innerpoint\Vert &\geq \strongconst \Vert \inner - \innerpoint\Vert^2 \\
    \implies \firstlip \Vert \outer - \outerpoint \Vert \cdot \Vert \inner - \innerpoint\Vert &\geq \strongconst \Vert \inner - \innerpoint\Vert^2.
\end{align}
Assuming $\Vert \inner - \innerpoint \Vert \neq 0$ we obtain the desired result.
\end{proof}

\begin{lemma}[Nystr\"om Bound]
\label{lem:nys-bound}
Suppose Assumptions \ref{ass:strongly-convex}, \ref{ass:first-order-lip}, \ref{ass:second-order-lip} and \ref{ass:grad-bound} and that $\Hessian$ is a rank-$\rank$ matrix.

Denote $\hvp[][*][\iter] = (\grad[\inner][2]\innerobj[][\outeriter](\inner)^{-1} \grad[\inner]\outerobj[](\outer[][][\iter],\inner)$
Select $\rank \geq \order \bigg(\frac{\truerank \log(\sfrac{1}{\highprob})}{\nystromerror^4} \bigg)$ columns with probability proportional to $\hessentry[ii]$, then the following bound holds with probability of at least $1-\highprob$:

\begin{align}
    \Ex[\Vert \hvp[][][\iter] - \hvp[][*][\iter] \Vert ] \leq \Psi,
    \label{eq:Psi}
\end{align}

where $\Psi = \frac{\gradbound}{\nystromconst} \frac{\lambda_{\rank+1} + \nystromerror \truerank \firstlip^2}{ \lambda_{\rank+1} + \nystromerror \truerank \firstlip^2   + \nystromconst}$.
\end{lemma}

\begin{proof}
We denote the true Hessian at step $\iter$ of $\grad[\inner][2]\innerobj[][\outeriter](\inner)$ as $ \Hessian[][][\iter]$ and the Nystr\"om low-rank approximation as $\Hessian[\rank][][\iter]$. We denote the following IHVPs:
\begin{align}
\hvp[][*][\iter] = (\Hessian[][][\iter] + \nystromconst \I)^{-1} \grad[\inner]\outerobj[](\outer[][][\iter],\inner) \quad \text{and}\quad  \hvp[][][\iter] = (\Hessian[\rank][][\iter] + \nystromconst \I)^{-1}  \grad[\inner]\outerobj[](\outer[][][\iter],\inner).
\end{align}
With \cref{ass:strongly-convex} we have that $\Hessian[][][\iter], \Hessian[\rank][][\iter]$ are positive semidefinite and invertible. Now, we start by bounding the difference in IHVPs:
\begin{align}
\Vert \hvp[][*][\iter] - \hvp[][][\iter] \Vert_2 &\leq \bigg \Vert (\Hessian[][][\iter] + \nystromconst \I)^{-1}  \grad[\inner]\outerobj(\outer[][][\iter], \inner) - (\Hessian[\rank][][\iter] + \nystromconst \I)^{-1}  \grad[\inner]\outerobj[](\outer[][][\iter],\inner) \bigg \Vert_2 \\
& \leq\bigg \Vert \grad[\inner]\outerobj[](\outer[][][\iter],\inner) \bigg \Vert (\Hessian[][][\iter] + \nystromconst \I)^{-1}  - (\Hessian[\rank][][\iter] + \nystromconst \I)^{-1}  \bigg \Vert_{\onorm}\\
&\leq \bigg\Vert \grad[\inner]\outerobj[](\outer[][][\iter],\inner) \bigg \Vert_2  \bigg( \frac{1}{\nystromconst} \frac{\Vert \Hessian[][][\iter] - \Hessian[\rank][][\iter] \Vert_{\onorm}}{\Vert \Hessian[][][\iter] - \Hessian[\rank][][\iter] \Vert_{\onorm} + \nystromconst} \bigg) \quad \text{Using theorem 1 of \cite{hataya2023Nystrom}}
\label{eq:hayata_thm1}.
\end{align}

Now, given that $\Hessian$ is symmetric and positive definite, it can be be factored into the quadratic form using via Choelesky decomposition implying that it is indeed a Gram matrix. With this in hand, we can utilize theorem 1 from \cite{Drineas2005}, to select columns with probability  to $ \{Pr_i\}^\innercard_{i=1}$ such that:

\begin{align}
    Pr_i = \frac{\hessentry[ii][2]}{\sum^{\truerank}_{i=1} \hessentry[ii][2]},
\end{align}

Requiring that $\rank \geq \order \bigg( \frac{\truerank \log(1/\highprob)}{\nystromerror^4} \bigg)$, then the following bound holds with probability of at least $1-\delta$ using theorem 2 of \cite{drineas2005Nystrom}:

\begin{align}
    \Ex [ \Vert \Hessian - \Hessian[\rank] \Vert ]_{\onorm} &\leq \Vert \Hessian -\Hessian[\rank][*] \Vert_{\onorm} + \nystromerror \sum^{\truerank}_{i=1} \hessentry[ii][2]  \\
    &\leq \lambda_{\rank+1} + \nystromerror \sum^{\truerank}_{i=1} \hessentry[ii][2],
    \label{eq:sum_diag_error}
\end{align}

where $\Hessian[\rank][*]$ is the best $\rank$-rank approximation, and $\lambda_{\rank+1}$ is the $\rank+1$ largest eigenvalue. With \cref{ass:second-order-lip} we can bound the largest eigenvalue of $\Hessian$ by $\firstlip$. Thus we can bound the sum in \cref{eq:sum_diag_error} from above by $\nystromerror \truerank \firstlip^2 $: 
\begin{align}
\label{eq:sum_diag}
    \Ex [ \Vert \Hessian - \Hessian[\rank] \Vert ]_{\onorm} &\leq \lambda_{\rank+1} + \nystromerror \truerank\firstlip^2.
\end{align}

Finally, combining \cref{eq:sum_diag} with \cref{eq:hayata_thm1} we obtain and applying \cref{ass:grad-bound} we obtain:
\begin{align}
  \Ex [ \Vert \hvp[][][\iter] - \hvp[][*][\iter] \Vert ]\leq \Psi \quad \text{with probability of a least } 1 - \highprob.
\end{align}
with $\Psi = \frac{\gradbound}{\nystromconst} \frac{\lambda_{\rank+1} + \nystromerror \truerank \firstlip^2}{ \lambda_{\rank+1} + \nystromerror \truerank \firstlip^2   + \nystromconst} $

\end{proof}

\begin{lemma}[IHVP Bound]
\label{lem:hvp-bound}
Suppose Assumptions \ref{ass:strongly-convex}, \ref{ass:first-order-lip}, \ref{ass:second-order-lip} and \ref{ass:grad-bound}, then
$$\Vert \hvp[][*][\iter] - \hvp[][*][\iter - 1] \Vert  \leq \bigg( \frac{\firstlip}{\strongconst + \nystromconst} + \frac{\secondlip \gradbound}{(\strongconst + \nystromconst)^2} \bigg)\bigg(1 + \frac{\firstlip}{\strongconst} \bigg)\Vert \hyperobj(\outer[][][\iter]) \Vert_2 $$ 
\end{lemma}

\begin{proof}
\begin{align}
    \Vert \hvp[][*][\iter] - \hvp[][*][\iter - 1] \Vert_2 &= \bigg \Vert  (\Hessian[][][\iter] + \nystromconst \I)^{-1}  \grad[\inner] \outerobj(\outer[][][\iter], \inner) - (\Hessian[][][\iter -1] + \nystromconst \I)^{-1} \grad[\inner] \outerobj(\outer[][][\iter-1], \inner)\bigg \Vert_2 \\
    &= \bigg \Vert ( \Hessian[][][k] + \nystromconst \I)^{-1} \grad[\inner] \outerobj(\outer[][][\iter], \inner) - \grad[\inner] \outerobj(\outer[][][\iter-1], \inner) \\
    &\quad + \grad[\inner]\outerobj(\outer[][][\iter-1], \inner) \bigg \Vert_2 \text{\arjun{ check ordering use op norm}} \\
    &\leq \bigg \Vert ( (\Hessian[][][k] + \nystromconst \I)^{-1}  \bigg \Vert_{\onorm} \bigg \Vert \grad[\inner] \outerobj(\outer[][][\iter], \inner) - \grad[\inner] \outerobj(\outer[][][\iter-1], \inner) \bigg \Vert_2 \\
    &\quad + \bigg \Vert (\Hessian[][][\iter] + \nystromconst \I)^{-1} - (\Hessian[][][\iter-1] + \nystromconst \I)^{-1} \bigg \Vert_{\onorm} \bigg \Vert \grad[\inner]\outerobj(\outer[][][\iter-1], \inner)\bigg \Vert_2 .
\end{align}

Using \cref{ass:strongly-convex}, we have that $\Vert (\Hessian + \nystromconst \I)^{-1}  \Vert_{\onorm} \leq \frac{1}{\strongconst + \nystromconst}$. Combining with \cref{ass:first-order-lip}.

\begin{align}
    \quad &\leq  \frac{\firstlip}{\strongconst + \nystromconst} \bigg \Vert \both[][][\iter] - \both[][][\iter-1] \bigg \Vert_2 \\
    &\quad + \gradbound \bigg \Vert (\Hessian[][][\iter] + \nystromconst \I)^{-1} - (\Hessian[][][\iter-1] + \nystromconst \I)^{-1} \bigg \Vert_{\onorm}.
    \label{eq:z1}
\end{align}

Now we use the fact $\bm{A}^{-1} - \bm{B}^{-1} = \bm{A}(\bm{B}-\bm{A})\bm{B}^{-1}$ we obtain:

\begin{align}
    \bigg \Vert (\Hessian[][][\iter] + \nystromconst \I)^{-1} - (\Hessian[][][\iter-1] + \nystromconst \I)^{-1} \bigg \Vert_{\onorm} &=  
    \bigg \Vert (\Hessian[][][\iter] + \nystromconst \I )^{-1} \bigg( (\Hessian[][][\iter-1] + \nystromconst \I) -  (\Hessian[][][\iter] + \nystromconst \I) \bigg ) (\Hessian[][][\iter-1] + \nystromconst \I)^{-1} \bigg \Vert_{\onorm}\\
    &= \frac{\secondlip}{(\strongconst + \nystromconst)^2} \Vert \both[][][\iter] - \both[][][\iter-1] \Vert_2.
    \label{eq:inverse_id}
\end{align}

Combining \Cref{eq:z1} and \Cref{eq:inverse_id} we obtain:
\begin{align}
    \Vert \hvp[][*][\iter] - \hvp[][*][\iter - 1] \Vert_2 &\leq \frac{\firstlip}{\strongconst + \nystromconst} \Vert \both[][][\iter] - \both[][][\iter-1] \Vert_2  +  \frac{\secondlip \gradbound}{(\strongconst + \nystromconst)^2} \Vert \both[][][\iter] - \both[][][\iter-1] \Vert_2 \\
    &= \bigg( \frac{\firstlip}{\strongconst + \nystromconst} + \frac{\secondlip \gradbound}{(\strongconst + \nystromconst)^2} \bigg) \Vert \both[][][\iter] - \both[][][\iter-1] \Vert_2 \\
    \Vert \hvp[][*][\iter] - \hvp[][*][\iter - 1] \Vert_2 &\leq \frac{\firstlip}{\strongconst + \nystromconst} \Vert \both[][][\iter] - \both[][][\iter-1] \Vert_2  +  \frac{\secondlip \gradbound}{(\strongconst + \nystromconst)^2} \Vert \both[][][\iter] - \both[][][\iter-1] \Vert_2 \\
    &\leq \bigg( \frac{\firstlip}{\strongconst + \nystromconst} + \frac{\secondlip \gradbound}{(\strongconst + \nystromconst)^2} \bigg)\bigg(1 + \frac{\firstlip}{\strongconst} \bigg)\Vert \outer[][][\iter] - \outer[][][\iter-1] \Vert_2 \text{\quad Using \cref{lem:warm-starts}} \\
    &\leq \bigg( \frac{\firstlip}{\strongconst + \nystromconst} + \frac{\secondlip \gradbound}{(\strongconst + \nystromconst)^2} \bigg)\bigg(1 + \frac{\firstlip}{\strongconst} \bigg)\Vert \hyperobj(\outer[][][\iter]) \Vert_2. \text{\quad \arjun{checked. this seems to be correct, but off by a small factor compared to Ji}}
\end{align}

\end{proof}

\begin{lemma}[Hypergradient Error Bound]
Suppose Assumptions \ref{ass:strongly-convex}, \ref{ass:first-order-lip}, \ref{ass:second-order-lip} and \ref{ass:grad-bound}. , then with probability of at least $1 - \highprob$
\arjun{checkpoint}

\begin{align}
    \Ex [\Vert  \esthypergrad(\outer[][][\iter]) - \grad \hyperobj(\outer[][][\iter]) \Vert^2 ] \leq \Gamma  ( 1 - \learnrate[\inner] \strongconst)^{D}\Vert \inner[][0][\iter] - \inner[][][k] \Vert^2 + 2\firstlip^2\Psi^2,
\end{align}

where 
\begin{align}
  \Gamma \doteq \bigg( \firstlip^2 + \frac{\secondlip^2 \gradbound^2}{(\strongconst+\nystromconst)^2} + 2\firstlip^2 \bigg(\frac{2\firstlip}{(\strongconst + \nystromconst)}+\frac{2\gradbound \secondlip}{(\strongconst+\nystromconst)^2}\bigg)^2 \bigg) ,
  \label{eq:big-constants}
\end{align}
and $\Psi$ is given in $\cref{eq:Psi}$.

\label{lem:hyperbound}
\end{lemma}

\begin{proof}
    \begin{align}
          \Vert  \esthypergrad(\outer[][][\iter]) - \grad \hyperobj(\outer[][][\iter]) \Vert^2 &\leq \Vert \grad[\outer] \outerobj(\outer[][][\iter], \inner[][][k]) - \grad[\outer] \outerobj(\outer[][][\iter],\inner[][\inneriters][\iter] \Vert^2 \\
         & \quad + \Vert \grad[\outer\inner] \innerobj( \inner[][\inneriters][\iter])\Vert \cdot \Vert \hvp[][*][\iter] - \hvp[][][\iter]\Vert^2 \\
         & \quad+ \Vert \grad[\outer\inner]\innerobj( \inner[][][k]) - \grad[\outer\inner]\innerobj(\inner[][\inneriters][\iter]) \Vert^2 \cdot \Vert \hvp[][*][\iter] \Vert^2 \\
         & \leq \firstlip^2 \Vert \inner[][][k] - \inner[][\inneriters][\iter]\Vert^2 + \firstlip^2\Vert \hvp[][*][\iter] - \hvp[][][\iter]\Vert^2 \\
         & \quad + \secondlip^2 \Vert \inner[][][k] - \inner[][\inneriters][\iter] \Vert^2 \cdot \Vert \hvp[][*][\iter] \Vert^2 \text{\quad \arjun{check that the mixed partial is bounded by L^2}} \\
         &\leq \firstlip^2 \Vert \inner[][][k] - \inner[][\inneriters][\iter]\Vert^2 + \firstlip^2\Vert \hvp[][*][\iter] - \hvp[][][\iter]\Vert^2 \\
         & \quad + \frac{\secondlip^2 \gradbound^2}{\strongconst^2} \Vert \inner[][][k] - \inner[][\inneriters][\iter] \Vert^2 \text{\quad since $\Vert \hvp[*] \Vert \leq \frac{\gradbound}{\strongconst}$}.
         \label{eq:hypergrad-diff}
    \end{align}

Now, we use $\inner[][\inneriters][\iter]$ to denote that our gradient descent learns an inexact $\inner[][][k]$ with $\numnestediters$ steps of gradient descent. This leads to an HVP $$\hvp[][\inneriters][\iter] = (\grad[\inner][2]\innerobj(\outer[][][\iter],\inner[][\inneriters][\iter])^{-1}\grad[\inner]\outerobj(\outer[][][\iter],\inner[][\inneriters][\iter]).$$ Additionally, we denote $ \hvp[][*][\iter]$ to be the true IHVP and $\hvp[][][\iter]$ to be the low-rank IHVP. We now apply the triangle inequality to these three IHVPs:

\begin{align}
    \Vert \hvp[][*][\iter] - \hvp[][][\iter] \Vert \leq   \Vert \hvp[][*][\iter] - \hvp[][\inneriters][\iter] \Vert +   \Vert \hvp[][][\iter] - \hvp[][\inneriters][\iter] \Vert.
    \label{eq:inexact-hvp}
\end{align}

First, we bound $ \Vert \hvp[][*][\iter] - \hvp[][\inneriters][\iter] \Vert$
\begin{align}
    \Vert \hvp[][*][\iter] - \hvp[][\inneriters][\iter] \Vert &= \bigg \Vert  (\Hessian[][][\iter] + \nystromconst \I)^{-1}  \grad[\inner]\outerobj(\outer[][][\iter], \inner[][][k]) - (\Hessian[][\inneriters][\iter] + \nystromconst \I)^{-1} \grad[\inner]\outerobj(\outer[][][\iter], \inner[][\inneriters](\outer)) \bigg \Vert_2 \\
    &\leq \bigg \Vert  (\Hessian[][][\iter] + \nystromconst \I)^{-1} \Vert_{\onorm} \cdot \bigg \Vert \grad[\inner]\outerobj(\outer[][][\iter], \inner[][][k]) - \grad[\inner]\outerobj(\outer[][][\iter], \inner[][\inneriters][\iter]) \bigg \Vert_2 \\
    &\quad + \bigg \Vert (\Hessian[][][\iter] + \nystromconst \I)^{-1} - (\Hessian[][\inneriters][\iter] + \nystromconst 
    \I)^{-1}\bigg \Vert_{\onorm} \cdot \bigg \Vert \grad[\inner]\outerobj(\outer[][][\iter], \inner[][\inneriters](\outer))  \bigg \Vert_2 \\
    &\leq  \bigg (\frac{\firstlip}{\strongconst+ \alpha}  + \frac{\gradbound \secondlip}{(\strongconst + \nystromconst)^2} \bigg) \bigg \Vert \inner[][][k] - \inner[][\inneriters][\iter] \bigg \Vert_2. \quad \text{\arjun{checked, correct!}}
\end{align}

Now, bound $\Vert \hvp[][][\iter] - \hvp[][\inneriters][\iter] \Vert$

\begin{align}
    \Vert \hvp[][][\iter] - \hvp[][\inneriters][\iter] \Vert &=  \bigg \Vert  (\Hessian[\rank][][\iter] + \nystromconst \I)^{-1}  \grad[\inner]\outerobj(\outer[][][\iter], \inner[][][k]) - (\Hessian[][\inneriters][\iter] + \nystromconst \I)^{-1} \grad[\inner]\outerobj(\outer[][][\iter], \inner[][\inneriters]) \bigg \Vert_2 \\
     &\leq \bigg \Vert  (\Hessian[\rank][][\iter] + \nystromconst \I)^{-1} \Vert_{\onorm} \cdot \bigg \Vert \grad[\inner]\outerobj(\outer[][][\iter], \inner[][][k]) - \grad[\inner]\outerobj(\outer[][][\iter], \inner[][\inneriters][\iter]) \bigg \Vert_2 \\
    &\quad + \bigg \Vert (\Hessian[\rank][][\iter] + \nystromconst \I)^{-1} - (\Hessian[][\inneriters][\iter] + \nystromconst \I)^{-1} \bigg \Vert_{\onorm} \cdot \bigg \Vert \grad[\inner]\outerobj(\outer[][][\iter], \inner[][\inneriters])  \bigg \Vert_2 \\
    &\leq \frac{\firstlip}{\strongconst + \nystromconst} + \gradbound \bigg \Vert \inner[][][k] - \inner[][\inneriters][\iter] \bigg \Vert_2 \bigg \Vert (\Hessian[\rank][][\iter] + \nystromconst \I)^{-1} - (\Hessian[][\inneriters][\iter] + \nystromconst \I)^{-1} \bigg \Vert_{\onorm}  \\ 
    &\leq \frac{\firstlip}{\strongconst + \nystromconst} \bigg \Vert \inner[][][k] - \inner[][\inneriters][\iter] \bigg \Vert_2  \\
    &\quad + \gradbound \bigg \Vert (\Hessian[\rank][][\iter] + \nystromconst \I)^{-1} - (\Hessian[][][\iter] + \nystromconst \I)^{-1}  + (\Hessian[][][\iter] + \nystromconst \I)^{-1} -  (\Hessian[][\inneriters][\iter] + \nystromconst \I)^{-1} \bigg \Vert_{\onorm} \\
    &\leq  \bigg (\frac{\firstlip}{\strongconst+ \alpha} + \frac{\gradbound \secondlip}{(\strongconst + \nystromconst)^2} \bigg)  \bigg \Vert \inner[][][k] - \inner[][\inneriters][\iter] \bigg \Vert_2 + \gradbound \bigg \Vert (\Hessian[\rank][][\iter] + \nystromconst \I)^{-1}-  (\Hessian[][][\iter] + \nystromconst \I)^{-1} \bigg \Vert_{\onorm}.
\end{align}

By applying  \cref{lem:hvp-bound}, we obtain with high probability:

\begin{align}
    \Ex [\Vert \hvp[][][\iter] - \hvp[][\inneriters][\iter] \Vert] \leq \bigg (\frac{\firstlip}{\strongconst+ \alpha}  + \frac{\gradbound \secondlip}{(\strongconst + \nystromconst)^2} \bigg)  \bigg \Vert \inner[][][k] - \inner[][\inneriters][\iter] \bigg \Vert_2 + \Psi.
\end{align}

Recombining \cref{eq:inexact-hvp}, we get with high probability:

\begin{align}
      \Ex[\Vert \hvp[][*][\iter] - \hvp[][][\iter] \Vert] \leq \bigg (\frac{2\firstlip}{\strongconst+ \alpha}  + \frac{2\gradbound \secondlip}{(\strongconst+\nystromconst)^2} \bigg) \bigg \Vert \inner[][][k] - \inner[][\inneriters][\iter] \bigg \Vert_2 + \Psi.
\end{align}

Recombining with \cref{eq:hypergrad-diff}, we obtain:

\begin{align}
   \Ex [ \Vert  \esthypergrad(\outer[][][\iter]) - \grad \hyperobj(\outer[][][\iter]) \Vert^2 ] &\leq \firstlip^2 \Vert \inner[][][k] - \inner[][\inneriters][\iter]\Vert^2  + \frac{\secondlip^2 \gradbound^2}{(\strongconst+\nystromconst)^2} \Vert \inner[][][k] - \inner[][\inneriters][\iter] \Vert^2 + \firstlip^2\Vert \hvp[][*][\iter] - \hvp[][][\iter]\Vert^2 \\
    & \leq \firstlip^2 \Vert \inner[][][k] - \inner[][\inneriters][\iter]\Vert^2  + \frac{\secondlip^2 \gradbound^2}{(\strongconst+\nystromconst^2)} \Vert \inner[][][k] - \inner[][\inneriters][\iter] \Vert^2 + \\
    & \quad+ 2\firstlip^2 \bigg (\frac{2\firstlip}{\strongconst+ \alpha}  + \frac{2\gradbound \secondlip}{(\strongconst+\nystromconst)^2} \bigg) \bigg \Vert \inner[][][k] - \inner[][\inneriters][\iter] \bigg \Vert^2 + 2\firstlip^2 \Psi^2 \\
    &\leq \bigg( \firstlip^2 + \frac{\secondlip^2 \gradbound^2}{(\strongconst+\nystromconst)^2} + 2\firstlip^2 \bigg(\frac{2\firstlip}{(\strongconst + \nystromconst)}+\frac{2\gradbound \secondlip}{(\strongconst+\nystromconst)^2}\bigg)^2 \bigg)  ( 1 - \learnrate[\inner] \strongconst)^{\frac{D}{2}}\Vert \inner[][0][\iter] - \inner[][][k] \Vert^2  \\
    & \quad + 2\firstlip^2 \Psi^2,
\end{align}

where the last line follows because of \cref{ass:strongly-convex}. Thus, we can use the gradient descent lemma of strongly convex functions learning rate $\learnrate[\inner]$ we have that $\Vert \inner[][][k] - \inner[][\inneriters][\iter] \Vert \leq \Gamma ( 1 - \learnrate[\inner] \strongconst)^{\frac{D}{2}}\Vert \inner[][0][\iter] - \inner[][][k] \Vert$ which holds with high probability.
    
\end{proof}

\begin{lemma}[Inner optimization error bound]

Suppose Assumptions \ref{ass:first-order-lip}, \ref{ass:second-order-lip}, \ref{ass:grad-bound}. 
Choose $\numnestediters \geq \ln{2(2+4\condnum^2\learnrate[\outer]^2 \Gamma)} \div \ln \frac{1}{1 - \learnrate[\inner]} = \Theta(\ln\condnum) = \Theta(\condnum)$, then:
\begin{align}
\Vert \inner[][0][\iter] - \inner[][][k] \Vert^2 \leq \bigg( \frac{1}{2} \bigg)^k \Vert \inner[][0][0] - \inner[][][0] \Vert^2 + \sum^{\iter-1}_{i=0} \bigg(\frac{1}{2} \bigg)^{\iter-1-i}  \bigg[ 8 \condnum^2 \learnrate[\outer]^2 \firstlip^2\Psi^2 + 4 \condnum^2 \learnrate[\outer]^2 \Vert \hyperobj(\outer[][][i]) \Vert^2  \bigg],
\end{align}
where $\Gamma, \Psi$ are defined in \cref{eq:big-constants}
\label{lem:combined-grad-bound}
\end{lemma}

\begin{proof}
Using a warm start for the inner loop we have $\inner[][0][\iter] = \inner[][\inneriters][\iter-1]$

\begin{align}
\Vert \inner[][0][\iter] - \inner[][][k] \Vert^2 &\leq 2\Vert \inner[][\inneriters][\iter-1] - \inner[][][\iter-1]\Vert^2 + 2\Vert \inner[][][\iter-1] - \inner[][][\iter-1]\Vert^ 2 \\
&\leq 2(1-\learnrate[\inner] \strongconst)^\numnestediters \Vert \inner[][0][\iter-1]- \inner[][][\iter-1] \Vert^2 + 2 \condnum^2 \learnrate[\outer]^2 \Vert \esthyperobj(\outer[][][\iter-1]) \Vert^2 \text{\quad \cite{ghadimi_approximation_2018} lemma 2.2}\\ 
&\leq 2(1-\learnrate[\inner] \strongconst)^\numnestediters \Vert \inner[][0][\iter-1]- \inner[][][\iter-1] \Vert^2 \\
&\quad+ 4\condnum^2 \learnrate[\outer]^2 \Vert \grad\hyperobj(\outer[][][\iter-1]) - \grad \esthyperobj(\outer[][][\iter-1])\Vert^2 + 4 \condnum^2 \learnrate[\outer]^2 \Vert \hyperobj(\outer[][][\iter-1]) \Vert^2 \\
&\leq \bigg(2(1-\learnrate[\inner] \strongconst)^D + 4\condnum^2 \learnrate[\outer]^2\Gamma(1 - \learnrate[\inner]\strongconst)^D\bigg) \Vert \inner[][0][\iter-1]- \inner[][][\iter-1] \Vert^2 \\
&\quad + 8 \condnum^2 \learnrate[\outer]^2 \firstlip^2\Psi^2 + 4 \condnum^2 \learnrate[\outer]^2 \Vert \hyperobj(\outer[][][\iter-1]) \Vert^2.  \text{ \quad \arjun{A5}}
\label{eq:to_scope}
\end{align}


\arjun{New D}
Now, using $\learnrate[\inner] = \frac{1}{\firstlip}$, and $\learnrate[\outer] \in \order( \frac{1}{\hyperlip})$ we determine the choice of $\inneriters$ inner iterations. Using the \Cref{eq:to_scope} to bound the strongly-convex gradient descent lemma by a constant $\frac{1}{2}$:
\begin{align}
    (2 + 4\condnum^2 \learnrate[\outer]^2\Gamma)(1 - \learnrate[\inner]\strongconst)^{\numnestediters} \leq \frac{1}{2}.
\end{align}

Simplifying the LHS, we obtain, and for a large value of $\condnum$ we obtain:

\begin{align}
    (2 + \order(\condnum^{-2}))(1 - \learnrate[\inner] \strongconst)^{\inneriters} &\leq \frac{1}{2} \\
     (1 - \learnrate[\inner] \strongconst)^{\inneriters}&\leq \frac{1}{4}\\
    \inneriters \log (1 - \learnrate[\inner] \strongconst) &\leq - \log(4) \\
    \inneriters &\geq \frac{\log(4)}{- \log(1 -\learnrate[\inner]\strongconst)} \\
    &\geq \frac{\log(4)}{-\log(1 - \sfrac{1}{\condnum})}\\
    &\geq \frac{\log(4)}{\sfrac{1}{\condnum}} = \condnum \log(4).
\end{align}

Thus, it is sufficient to choose $\inneriters \in \Theta(\condnum)$. Finally, we telescope \cref{eq:to_scope} to obtain the final result:

\arjun{check if off by a factor of 2 from $||a - b||^2 \leq a^2 + b^2$}

\begin{align}
\Vert \inner[][0][\iter] - \inner[][][k] \Vert^2 \leq \bigg( \frac{1}{2} \bigg)^k \Vert \inner[][0][0] - \inner[][][0] \Vert^2 + \sum^{\iter-1}_{i=0} \bigg(\frac{1}{2} \bigg)^{\iter-1-i}  \bigg[ 8 \condnum^2 \learnrate[\outer]^2 \firstlip^2\Psi^2 + 4 \condnum^2 \learnrate[\outer]^2 \Vert \hyperobj(\outer[][][i]) \Vert^2  \bigg].
\end{align}

\end{proof}

\begin{corollary}
\label{cor:hyperbound}
Combining \Cref{lem:combined-grad-bound} and \Cref{lem:hyperbound} we obtain with probability $1 = \highprob$:
\begin{align}
     &\Ex [\Vert  \esthypergrad(\outer[][][\iter]) - \grad \hyperobj(\outer[][][\iter]) \Vert^2 ] \\
     & \quad \leq \gdterm \bigg[  \bigg(\frac{1}{2}\bigg)^{\iter} \Vert \inner[][0][0] - \inner[][][0] \Vert^2 + \sum^{\iter-1}_{i=0} \bigg(\frac{1}{2}\bigg)^{\iter -1 - i} 8 \condnum^2 \learnrate[\outer]^2 \firstlip^2\Psi^2 + 4 \condnum^2 \learnrate[\outer]^2 \sum^{\iter-1}_{i=0}  \bigg(\frac{1}{2}\bigg)^{\iter - 1-i}  \Vert \grad \hyperobj(\outer[][][i])\Vert^2 \bigg] + 2\firstlip^2\Psi^2\end{align}
Where $\gdterm = \Gamma(1 - \learnrate[\inner]\strongconst)^\numnestediters$
\end{corollary}

%% file: appendix/proofs/convergence_proof.tex
\subsubsection{Main Result}
\convergence*

\begin{proof}
\label{proof:thm1}
The proof is adapted from \cite{ji2021bilevel} to take into account the error introduced by the Nystr\"om approximation.  The characterization $\hyperlip$ is adapted from lemma 2 of \cite{ji2021bilevel} which leverages the result from \cite{ghadimi_approximation_2018} lemma 2.2:

 \begin{align}
        \hyperlip &= \firstlip + \frac{2\firstlip^2 + \secondlip\gradbound^2}{\strongconst + \nystromconst} + \frac{2\secondlip\firstlip\gradbound + \firstlip^3}{(\strongconst + \nystromconst)^2} + \frac{\secondlip\firstlip^2\gradbound}{(\strongconst+\nystromconst)^3} \\
        &= \order(\condnum^3).
\label{eq:hyperlip}
\end{align}

Then, writing the gradient descent expression:
\begin{align}
     \hyperobj(\outer[][][\iter+1]) &\leq \hyperobj(\outer[][][\iter]) + \langle \grad \hyperobj(\outer[][][\iter]), \outer[][][\iter+1] - \outer[][][\iter]\rangle + \frac{\hyperlip}{2}\Vert \outer[][][\iter+1] - \outer[][][\iter]\Vert^2 \\
     &\leq \hyperobj(\outer[][][\iter]) - \learnrate[\outer] \langle \grad \hyperobj(\outer[][][\iter]), \hat{\grad}\hyperobj(\outer[][][\iter]) - \grad \hyperobj(\outer[][][\iter])\rangle - \learnrate[\outer] \Vert \grad \hyperobj(\outer[][][\iter]\Vert^2 + \learnrate[\outer]^2 \hyperlip \Vert \grad \hyperobj(\outer[][][\iter])\Vert^2 \\
     &\quad + \learnrate[\outer]^2\hyperlip \Vert \grad \hyperobj(\outer[][][\iter]) - \hat{\grad} \hyperobj(\outer[][][\iter])\Vert^2 \\
     &= \hyperobj(\outer[][][\iter]) - \bigg(\frac{\learnrate[\outer]}{2} - \learnrate[\outer]^2 \hyperlip \bigg) \Vert \hyperobj(\outer[][][\iter]\Vert^2 + \bigg(\frac{\learnrate[\outer]}{2} + \learnrate[\outer]^2 \hyperlip \bigg)  \Vert  \esthypergrad(\outer[][][\iter]) - \grad \hyperobj(\outer[][][\iter]))\Vert^2. 
\end{align}

Using \Cref{lem:combined-grad-bound} gives with probability of at least $1 - \order(\highprob)$: 
\begin{align}
      \hyperobj(\outer[][][\iter+1]) &\leq \hyperobj(\outer[][][\iter]) - \bigg(\frac{\learnrate[\outer]}{2} - \learnrate[\outer]^2 \hyperlip \bigg) \Vert \hyperobj(\outer[][][\iter]\Vert^2\\
      &\quad +  \bigg(\frac{\learnrate[\outer]}{2} + \learnrate[\outer]^2 \hyperlip \bigg) \gdterm \bigg[  \bigg(\frac{1}{2}\bigg)^{\iter} \Vert \inner[][0][0] - \inner[][][0] \Vert^2 \\
      &\quad + \sum^{\iter-1}_{i=0} \bigg(\frac{1}{2}\bigg)^{\iter -1 - i} 8 \condnum^2 \learnrate[\outer]^2 \firstlip^2\Psi^2 + 4 \condnum^2 \learnrate[\outer]^2 \sum^{\iter-1}_{i=0}  \bigg(\frac{1}{2}\bigg)^{\iter  -1 -  i}  \Vert \grad \hyperobj(\outer[][][i])\Vert^2 \bigg] \\
      &\quad +  \bigg(\frac{\learnrate[\outer]}{2} +\learnrate[\outer]^2 \hyperlip \bigg) \firstlip^2\Psi^2 \\
      \hyperobj(\outer[][][\iter+1])  &\leq \hyperobj(\outer[][][\iter]) - \bigg(\frac{\learnrate[\outer]}{2} - \learnrate[\outer]^2 \hyperlip \bigg) \Vert \hyperobj(\outer[][][\iter]\Vert^2 \\
      &\quad +  \bigg(\frac{\learnrate[\outer]}{2} + \learnrate[\outer]^2 \hyperlip \bigg) \gdterm \bigg(\frac{1}{2}\bigg)^{\iter} \Vert \inner[][0][0] - \inner[][][0] \Vert^2 \\
      & \quad + 4 \bigg(\frac{\learnrate[\outer]}{2} + \learnrate[\outer]^2 \hyperlip \bigg) \gdterm \condnum^2 \learnrate[\outer]^2 \sum^{\iter-1}_{i=0}  \bigg(\frac{1}{2}\bigg)^{\iter - 1- i}  \Vert \grad \hyperobj(\outer[][][i])\Vert^2 \\
      &\quad + \bigg(\frac{\learnrate[\outer]}{2} + \learnrate[\outer]^2 \hyperlip \bigg) \gdterm \sum^\iter_{i=1} \bigg(\frac{1}{2}\bigg)^{\iter - i} 8 \condnum^2 \learnrate[\outer]^2 \firstlip^2\Psi^2 + \bigg(\frac{\learnrate[\outer]}{2} + \learnrate[\outer]^2 \hyperlip \bigg) \firstlip^2\Psi^2. 
\end{align}

Telescoping $\iter$ from 0 to $K-1$ to  becomes: \arjun{check}

\begin{align}
 \bigg(\frac{\learnrate[\outer]}{2} - \learnrate[\outer]^2 \hyperlip \bigg) \sum^{K-1}_{\iter=0} \Vert \grad \hyperobj(\outer[][][\iter])\Vert^2 &\leq \hyperobj(\outer[][][0]) - \inf_{\outer} \hyperobj(\outer) 
 \\ &\quad +  \bigg(\frac{\learnrate[\outer]}{2} + \learnrate[\outer]^2 \hyperlip \bigg) \gdterm \Vert \inner[][0][0] - \inner[][][0] \Vert^2 \\
 & \quad + \bigg(\frac{\learnrate[\outer]}{2} + \learnrate[\outer]^2 \hyperlip \bigg) \gdterm \condnum^2 \learnrate[\outer]^2 \sum^{K-1}_{k=1} \sum^{k-1}_{j=0} \bigg(\frac{1}{2}\bigg)^{k-1-j} \Vert \hyperobj(\outer[][][j])\Vert^2 \\
 &\quad + \gdterm   \bigg(\frac{\learnrate[\outer]}{2} + \learnrate[\outer]^2 \hyperlip \bigg) \sum^{K-1}_{k=1} \sum^{k-1}_{j=0} \bigg(\frac{1}{2}\bigg)^{k-1-j} 8 \condnum^2 \learnrate[\outer]^2 \firstlip^2\Psi^2 \\
 &\quad + \bigg(\frac{\learnrate[\outer]}{2} + \learnrate[\outer]^2 \hyperlip \bigg) K \firstlip^2\Psi^2 .
\end{align}

Now, using the fact that $\sum^{K-1}_{k=1} \sum^{k-1}_{j=0} \bigg(\frac{1}{2}\bigg)^{k-1-j} \Vert \hyperobj(\outer[][][j])\Vert^2 \leq\sum^{K-1}_{\iter=0} \frac{1}{2^\iter} \sum^{K-1}_{k=0} \Vert \grad \hyperobj(\outer[][][\iter])\Vert^2 \leq 2 \sum^{K-1}_{\iter=0} \Vert \grad \hyperobj(\outer[][][\iter])\Vert^2$ we obtain:

\begin{align}
\bigg(\frac{\learnrate[\outer]}{2} - \learnrate[\outer]^2 \hyperlip \bigg) \sum^{K-1}_{\iter=0} \Vert \grad \hyperobj(\outer[][][\iter])\Vert^2 &\leq \hyperobj(\outer[][][0]) - \inf_{\outer} \hyperobj(\outer) 
 \\ &\quad +  \bigg(\frac{\learnrate[\outer]}{2} + \learnrate[\outer]^2 \hyperlip \bigg) \gdterm \Vert \inner[][0][0] - \inner[][][0] \Vert^2 \\
 & \quad + 2\bigg(\frac{\learnrate[\outer]}{2} + \learnrate[\outer]^2 \hyperlip \bigg) \gdterm \condnum^2 \learnrate[\outer]^2 \sum^{K-1}_{\iter=0}  \Vert \hyperobj(\outer[][][\iter])\Vert^2 \\
 &\quad + 16 \bigg(\frac{\learnrate[\outer]}{2} + \learnrate[\outer]^2 \hyperlip \bigg) \gdterm \condnum^2 \learnrate[\outer]^2 K \firstlip^2\Psi^2 \\
 &\quad + \bigg(\frac{\learnrate[\outer]}{2} + \learnrate[\outer]^2 \hyperlip \bigg) K \firstlip^2\Psi^2 \\ 
 &\leq \hyperobj(\outer[][][0]) - \inf_{\outer} \hyperobj(\outer) 
 \\ &\quad +  \bigg(\frac{\learnrate[\outer]}{2} + \learnrate[\outer]^2 \hyperlip \bigg) \gdterm \Vert \inner[][0][0] - \inner[][][0] \Vert^2 \\
 & \quad + \bigg(\frac{\learnrate[\outer]}{2} + \learnrate[\outer]^2 \hyperlip \bigg) 2\gdterm \condnum^2 \learnrate[\outer]^2 \sum^{K-1}_{\iter=0}  \Vert \hyperobj(\outer[][][\iter])\Vert^2 \\
 &\quad + \bigg(\frac{\learnrate[\outer]}{2} + \learnrate[\outer]^2 \hyperlip \bigg) K \firstlip^2\Psi^2(16\gdterm \condnum^2 \learnrate[\outer]^2 + 1) .
\end{align}

Rearranging gives:

\begin{align}
    \bigg( \frac{\learnrate[\outer]}{2} - \learnrate[\outer]^2 \hyperlip - (\learnrate[\outer]^3\condnum^{2} + \learnrate[\outer]^4 \condnum^2 \hyperlip) \gdterm \bigg) \sum^{K-1}_{\iter=0}  \Vert \hyperobj(\outer[][][\iter])\Vert^2 &\leq \hyperobj(\outer[][][0]) - \inf_{\outer} \hyperobj(\outer) 
 \\ &\quad +  \bigg(\frac{\learnrate[\outer]}{2} + \learnrate[\outer]^2 \hyperlip \bigg) \gdterm \Vert \inner[][0][0] - \inner[][][0] \Vert^2 \\
 &\quad + \bigg(\frac{\learnrate[\outer]}{2} + \learnrate[\outer]^2 \hyperlip \bigg) K \firstlip^2\Psi^2(16\gdterm \condnum^2 \learnrate[\outer]^2 + 1) .
\end{align}

Select $\numnestediters \in \order(\condnum)$ to ensure that $\gdterm \leq 1$, $\gdterm \condnum^2 \learnrate[\outer]^2 \leq \frac{1}{16} $, $(\learnrate[\outer]^2 \condnum^{2} + \learnrate[\outer]^3\condnum^{2} \hyperlip) \gdterm \leq \frac{1}{4}$ produces the following:

\begin{align}
\bigg( \frac{\learnrate[\outer]}{4} - \learnrate[\outer]^2 \hyperlip \bigg) \sum^{K-1}_{\iter=0}  \Vert \hyperobj(\outer[][][\iter])\Vert^2 &\leq \hyperobj(\outer[][][0]) - \inf_{\outer} \hyperobj(\outer) 
 \\ &\quad +  \bigg(\frac{\learnrate[\outer]}{2} + \learnrate[\outer]^2 \hyperlip \bigg) \Vert \inner[][0][0] - \inner[][][0] \Vert^2 \\
 &\quad + 2 \bigg(\frac{\learnrate[\outer]}{2} + \learnrate[\outer]^2 \hyperlip \bigg) K \firstlip^2\Psi^2 .
\end{align}

Now, setting $\learnrate[\outer] \leq \frac{1}{8\hyperlip}$ we obtain:

\begin{align}
    \frac{1}{K}\sum^{K-1}_{\iter=0}  \Vert \hyperobj(\outer[][][\iter])\Vert^2 &\leq \frac{64 \hyperlip (\hyperobj(\outer[][][0]) - \inf_{\outer} \hyperobj(\outer)) +  5\Vert \inner[][0][0] - \inner[][][0] \Vert^2}{K} + 10 \firstlip^2\Psi^2 .
\end{align}

Now we can calculate the iteration complexities, by splitting the the $\epsilon$-error across the gradient descent error and the Nystr\"om error:
\begin{align}
    \frac{1}{K}\sum^{K-1}_{\iter=0}  \Vert \hyperobj(\outer[][][\iter])\Vert^2 \leq \underbrace{\frac{1}{K}64 \bigg(\hyperlip (\hyperobj(\outer[][][0]) - \inf_{\outer} \hyperobj(\outer)) +  5\Vert \inner[][0][0] - \inner[][][0] \Vert^2\bigg)}_{\text{gradient descent error}} + \underbrace{10 \firstlip^2\Psi^2}_{\text{Nystr\"om error}} \leq \epsilon .
\end{align}

Recall that we have $\hyperlip$ is $\order(\condnum^3)$. We chose the number of inner iterations following \Cref{lem:combined-grad-bound} to be $\inneriters = \order(\condnum)$. Then, we can bound the gradient descent term by $\frac{\epsilon}{2}$, and the Nystr\"om error term by $\frac{\epsilon}{2}$ with probability of at least $1 - \order(\highprob)$. This can be achieved by first choosing $K = \order(\frac{\condnum^3}{\sfrac{\epsilon}{2}})$ number of outer-loop iterations. And then, by following the sampling regime from \Cref{lem:hvp-bound}, we sample 
$\rank \geq \order\bigg(\frac{\Hessdim \log (\sfrac{1}{\highprob})}{(\sfrac{\epsilon}{2})^4}\bigg)$ columns. With these choices we can achieve an $\epsilon$-accurate stationary point with probability of at least $1 - \order(\highprob)$. This yields complexities  of $\order(\frac{\condnum^3}{\sfrac{\epsilon}{2}})$ for $\outerobj$,  $\order(\frac{\condnum^4}{\sfrac{\epsilon}{2}})$ for $\innerobj$,  $\order(\frac{\condnum^3}{\sfrac{\epsilon}{2}})$ for the Jacobian vector-product and  $\order(\frac{\condnum^3}{\sfrac{\epsilon}{2}})$ for the Hessian vector-product.
\end{proof}

\sarnie{}{First Bound the $\frac{1}{K}\sum^{K-1}_{\iter=0}  \Vert \hyperobj(\outer[][][\iter])\Vert^2$ term by the sum of the gradient descent error and the Nystrom error, then bound the gradient descent error by $\epsilon$ with some conditions(K iteration, q columns ...). Then bound the Nystrom error with a threshold with a high probability. Then the entire term can be bounded by something with a high probability, and we say that epsilon equals to this thing.}

\begin{remark}[On the choice of hyperparameters]

The regularization parameter $\nystromconst$ plays a crucial role in balancing numerical stability and the fidelity of curvature information. 
As discussed by \citet{vicol_implicit_2022}, for an eigenvalue $\lambda$ of the Hessian $H$, the effect of regularization is to modify the inverse eigenvalue as $(\lambda + \nystromconst)^{-1}$. 
When $\lambda \gg \nystromconst$, this term behaves as $\lambda^{-1}$, preserving curvature information. 
Conversely, when $\nystromconst \gg \lambda$, the approximation becomes insensitive to low-curvature directions, effectively ignoring them.  
In practice, $\nystromconst$ is selected empirically. 
One practical advantage of the Nystr\"om method is its ability to operate effectively with smaller $\nystromconst$.
For example, we use $\nystromconst = 50$, compared to $\nystromconst = 500$ in \citet{zheng2022stackelberg} for continuous control RL, and $\nystromconst = 10,000$ in \citet{fiez_implicit_2020} for a GAN-based bilevel problem, both of which use CG.

We also note that in our sampling procedure, columns are drawn independently, so with sampling probability $\delta \in [0, 1)$, the hypergradient bound holds with probability $1 - O(\delta)$. While theoretical results \cite{drineas2005Nystrom} may suggest that more than $\Hessdim$ columns are needed for especially for small matrices, however ceases to be a limitation for large matrices. Furthermore, empirically, neural network Hessians have been observed to exhibit low-rank structure \cite{ghorbani2019investigation, sagun2016eigenvalues}.
The Nystr\"om method exploits this structure, by reducing the problem to inverting a 
$\rank \times \rank$ matrix.
In practice, $\rank = 5$ seems sufficient \cite{hataya2023Nystrom}, which is tractable on modern GPU hardware.
\Cref{alg:nystrom-blo} presents our approach to bilevel optimization with Nystr\"om hypergradients.
\end{remark}

%% file: appendix/proofs/convex_rl.tex
\convexrl*
\begin{proof}
First we re-arrange the loss function using the temporal difference loss:
\begin{align}
    \criticobj[][\actorparams](\criticparams) &= \frac{1}{2} \Ex_{\substack{\state \sim \statedist[{\policy[\actorparams]}][\initstatedist]\\\nextstate \sim \policy_{\actorparams}}} \left[ \left( \Vfunc[][{\policy[\actorparams]}] (\state) - \Vfunc[\criticparams] (\state) \right)^2 \right] \\
    &= \frac{1}{2} \Ex_{\substack{\state \sim \statedist[{\policy[\actorparams]}][\initstatedist]\\\nextstate \sim \policy_{\actorparams}}} \left[ \left( \criticparams[][\top]\featurevec(\state) - (\reward + \discount \criticparams[][\top]\featurevec(\nextstate)) \right)^2 \right] \\
    &= \Ex_{\substack{\state \sim \statedist[{\policy[\actorparams]}][\initstatedist]\\\nextstate \sim \policy_{\actorparams}}} [(\criticparams[][\top] \fdiff(\state, \nextstate) - \reward)^2] .
\end{align}
Taking the derivative with respect to $\criticparams$ we obtain:

\begin{align}
    \grad[\criticparams] \criticobj[\actorparams](\criticparams) = \Ex_{\substack{\state \sim \statedist[{\policy[\actorparams]}][\initstatedist]\\\nextstate \sim \policy_{\actorparams}}}[(\criticparams[][\top] \fdiff(\state, \nextstate)) \fdiff(\state, \nextstate)].
\end{align}

Computing the Hessian yields:

\begin{align}
    \grad[\criticparams][2] \criticobj = \Ex_{\substack{\state \sim \statedist[{\policy[\actorparams]}][\initstatedist]\\\nextstate \sim \policy_{\actorparams}}}[\fdiff(\state, \nextstate) \fdiff(\state, \nextstate)^{\top}].
\end{align}
By hypothesis, $\Ex_{\substack{\state \sim \statedist[{\policy[\actorparams]}][\initstatedist]\\\nextstate \sim \policy_{\actorparams}}}[\fdiff(\state, \nextstate) \fdiff(\state, \nextstate)^{\top}]$ is positive definite, so

\begin{align}
    \grad[\criticparams][2] \criticobj = \Ex_{\substack{\state \sim \statedist[{\policy[\actorparams]}][\initstatedist]\\\nextstate \sim \policy_{\actorparams}}}[\fdiff(\state, \nextstate) \fdiff(\state, \nextstate)^{\top}] \precsim \lambda_{\min} \I
\end{align}
Where $\lambda_{\min}$ is the smallest eigenvalue, proving that $\criticobj[\actorparams](\criticparams)$ is $\strongconst$-strong convex with $\strongconst = \lambda_{\min}$
\end{proof}

\begin{remark}
Requiring a $\Ex_{\substack{\state \sim \statedist[{\policy[\actorparams]}][\initstatedist]\\\nextstate \sim \policy_{\actorparams}}}[\fdiff(\state, \nextstate) \fdiff(\state, \nextstate)^{\top}]$ to be positive definite means that the features $\featurevec(\state)$ and discounted next state $\discount\featurevec(\nextstate)$ collectively have full column rank \cite{bhandari2018finite} and is a standard assumption in reinforcement learning with a linear function approximation \cite{vanroy}. In other words, the features have enough coverage over the state distribution to avoid degeneracies. In particular, there is no direction $x$ in parameter space such that $\fdiff(\state, \nextstate)^{\top}x=0$. In practice this means that the policy visits a sufficiently rich set of states to not be rank-deficient. 
\end{remark}

%% file: appendix/proofs/surrogate.tex
\surrogate*
\if{0}
\sarnie{}{
    $\grad[\actorparams] \valfun[][{\policy[\actorparams]}](\state) =$
    \begin{align}
        \sum_{x \in \mathcal{S}} \sum_{k=0}^{\infty} \Pr(\state \to x, k, \policy) \sum_{\action} \grad[\actorparams] \policy[\actorparams] (\action \mid \state) \Qfunc[][{\policy[\actorparams]}] (\state, \action)
    \end{align} 
    \amy{you cannot sum over states or actions. both are assumed to be infinite.}
    \amy{what does your notation $\state \to x$ mean? it is undefined. but we do have notation for the visit/occupancy probability distribution, in case you need it: $\statedist[{\policy[\actorparams]}][\initstatedist]$.}
    is a result of the general policy gradient theorem. 
    In practice during a rollout, this gradient can be estimated by $\frac{1}{T^{'}}\sum_{\state[']=\state} \grad[\actorparams] \log \policy[\actorparams] (\action \mid \state[']) \Qfunc[][{\policy[\actorparams]}] (\state[s'], \action)$.
}

\sarjun{}{
\begin{align}
    \sum_{x \in \mathcal{S}} \sum_{k=0}^{\infty} \Pr(\state \to x, k, \policy) 
\end{align}
Is the probability of transitioning from state $\state$ to $x$ in $k$ steps under the current policy. 

\begin{align}
    &\grad[\actorparams] \valfun[][{\policy[\actorparams]}](\state[0]) \quad \text{arbitrary $\state[0] \sim \statedist[{\policy[\actorparams]}][\initstatedist]$} \\
    &=  \sum_{\x \in \mathcal{S}} \sum_{k=0}^{\infty} \Pr(\state[0] \to \x, k, \policy) \sum_{\action} \grad[\actorparams] \policy[\actorparams] (\action \mid \x) \Qfunc[][{\policy[\actorparams]}] (\x, \action) \\
    &= \sum_{\x} \eta(\x) \sum_{\action} \grad[\actorparams] \policy[\actorparams] (\action \mid \x) \Qfunc[][{\policy[\actorparams]}] (\x, \action) \\
    &= \sum_{\x^{\prime}} \eta(\x^\prime) \sum_{\x} \frac{\eta(\x)}{\sum_{\x^{\prime}}\eta(\x^{\prime})} \sum_{\action} \grad[\actorparams] \policy[\actorparams] (\action \mid \x) \Qfunc[][{\policy[\actorparams]}] (\x, \action) \\
    &= \sum_{\x^{\prime}} \eta(\x^\prime) \sum_{\x} \statedist[{\policy[\actorparams]}][\initstatedist](\x)
    \sum_{\action} \grad[\actorparams] \policy[\actorparams] (\action \mid \x) \Qfunc[][{\policy[\actorparams]}] (\x, \action) \\ 
    & \propto \sum_{\x} \statedist[{\policy[\actorparams]}][\initstatedist](\x)
    \sum_{\action} \grad[\actorparams] \policy[\actorparams] (\action \mid \x) \Qfunc[][{\policy[\actorparams]}] (\x, \action)\\
    &= \sum_{\x} \statedist[{\policy[\actorparams]}][\initstatedist](\x)
    \sum_{\action}  \policy[\actorparams] (\action \mid \x) \frac{\grad[\actorparams] \policy[\actorparams] (\action \mid \x)}{ \policy[\actorparams] (\action \mid \x)} \Qfunc[][{\policy[\actorparams]}] (\x, \action)\\
    &= \Ex_{\substack{\x \sim \statedist[{\policy[\actorparams]}][\initstatedist] \\ \action \sim \policy[\actorparams] (\x)}} \left[ \grad[\actorparams] \log \policy[\actorparams] (\action \mid \x)\Qfunc[][{\policy[\actorparams]}] (\x, \action) \right] \\
&\amy{where is $\state[0]$ in this last equation?}\\ &\arjun{it is not there because we only care about the successive states} \notag
\end{align} 

So the final expression becomes

\begin{align}
   &\Ex_{\substack{\state \sim \statedist[{\policy[\actorparams]}][\initstatedist] \\ \action \sim \policy[\actorparams] (\state)}} \left[ \left( \valfun[][{\policy[\actorparams]}] (\state) - \valfun[\criticparams] (\state) \right) \grad[\actorparams] \valfun(\state[0] = \state)\right] \\
   &= \Ex_{\substack{\state \sim \statedist[{\policy[\actorparams]}][\initstatedist]}} \left[ \left( \valfun[][{\policy[\actorparams]}] (\state) - \valfun[\criticparams] (\state) \right)
   \bigg( \Ex_{\substack{\traj \mid \state[0]=\state}}[
   \grad[\actorparams] \log \policy[\actorparams] (\action \mid \state) \Qfunc[][{\policy[\actorparams]}] (\state, \action)] \bigg) \right] 
\end{align}
\amy{we believe 33 is correct}
}

\begin{proof}

\amy{NEW. N.B. the notation ${\state'} \sim \histdist[{\policy[\actorparams]}][\state]$ means that $s'$ is drawn from the visit/occupancy distribution with $s_0 = s$. defined on p. 2.}

{\small
\begin{align}
\grad[\actorparams] \criticobj[][\actorparams] (\criticparams)
&= \frac{1}{2} \grad[\actorparams] \Ex_{\substack{\state \sim \statedist[{\policy[\actorparams]}][\initstatedist]}}  \left[ \left(  \Vfunc[][{\policy[\actorparams]}] (\state) - \Vfunc[\criticparams] (\state) \right)^2 \right] \\
&= \Ex_{\state \sim \statedist[{\policy[\actorparams]}][\initstatedist]} \left[ \left( \valfun[][{\policy[\actorparams]}] (\state) - \valfun[\criticparams] (\state) \right) \grad[\actorparams] \left( \valfun[][{\policy[\actorparams]}] (\state) - \valfun[\criticparams] (\state) \right) \right] \\
&= \Ex_{\state \sim \statedist[{\policy[\actorparams]}][\initstatedist]} \left[ \left( \valfun[][{\policy[\actorparams]}] (\state) - \valfun[\criticparams] (\state) \right) \grad[\actorparams] \valfun[][{\policy[\actorparams]}] (\state) \right]
\end{align}

Now, by the policy gradient theorem \cite{sutton1999policy},
\begin{align}
\grad[\actorparams] \valfun[][{\policy[\actorparams]}] (\state)
\propto& \Ex_{\substack{{\state'} \sim \histdist[{\policy[\actorparams]}][\state] \\ \action \sim \policy[\actorparams] (\state')}} \left[ \grad[\actorparams] \log \policy[\actorparams] (\action \mid \state') \, \Qfunc[][{\policy[\actorparams]}] (\state', \action) \right]
\end{align}
}
\end{proof}


\begin{corollary}
This \amy{this what?! ``THIS" alone is not enough information.} is estimated by
\begin{align}
    &\sum^{\trajlength}_{t=0}  \left( \valfun[][{\policy[\actorparams]}] (\state[t]) - \valfun[\criticparams] (\state[t]) \right)
    \bigg( \sum^{\trajlength-1-t}_{k=0} \discount^{k} (\grad[\actorparams] \log \policy[\actorparams] (\action[t+k] \mid \state[t+k]) \Qfunc[][{\policy[\actorparams]}] (\state[t+k], \action[t+k])) \bigg)
\end{align}
\end{corollary}
\fi

\begin{proof}
\begin{align}
\grad[\actorparams] \criticobj[][\actorparams] (\criticparams)
&= \frac{1}{2} \grad[\actorparams] \Ex_{\substack{\state \sim \statedist[{\policy[\actorparams]}][\initstatedist]}}  \left[ \left(  \Vfunc[][{\policy[\actorparams]}] (\state) - \Vfunc[\criticparams] (\state) \right)^2 \right] \\
&= \Ex_{\state \sim \statedist[{\policy[\actorparams]}][\initstatedist]} \left[ \left( \valfun[][{\policy[\actorparams]}] (\state) - \valfun[\criticparams] (\state) \right) \grad[\actorparams] \left( \valfun[][{\policy[\actorparams]}] (\state) - \valfun[\criticparams] (\state) \right) \right] \\
&= \Ex_{\state \sim \statedist[{\policy[\actorparams]}][\initstatedist]} \left[ \left( \valfun[][{\policy[\actorparams]}] (\state) - \valfun[\criticparams] (\state) \right) \grad[\actorparams] \valfun[][{\policy[\actorparams]}] (\state) \right].
\end{align}

Now, applying the log-derivative trick yields:
\begin{align}
\grad[\actorparams] \valfun[][{\policy[\actorparams]}] (\state)
\propto& 
\Ex_{{\traj[][\state]} \sim {\histdist[{\policy[\actorparams]}][\state]}} \left[ \grad[\actorparams] \log \histdist[{\policy[\actorparams]}][\state] (\traj[][\state]) \, \return[{\traj[][\state]}] \right].
\end{align}

Finally, by the policy gradient theorem \cite{sutton1999policy},
\begin{align}
\Ex_{{\traj[][\state]} \sim {\histdist[{\policy[\actorparams]}][\state]}} \left[ \grad[\actorparams] \log \histdist[{\policy[\actorparams]}][\state] (\traj[][\state]) \, \return[{\traj[][\state]}] \right]
=& \Ex_{\substack{{\state'} \sim \statedist[{\policy[\actorparams]}][\state] \\ \action \sim \policy[\actorparams] (\state')}} \left[ \grad[\actorparams] \log \policy[\actorparams] (\action \mid \state') \, \Qfunc[][{\policy[\actorparams]}] (\state', \action) \right].
\end{align}
\end{proof}


\begin{corollary}
Given $\numsamples$ sampled trajectories  of length $\timesteps$, $\{ (\state[0][1], \action[0][1], \reward[0][1], \state[1][1]$, $\ldots$, $\reward[\timesteps-1][0])$, $\ldots$, $(\state[0][\numsamples], 
\action[0][\numsamples], \reward[0][\numsamples], \state[1][\numsamples], \ldots, \reward[\timesteps-1][\numsamples]) \}$,
the gradient of the critic's objective function with respect to the actor's parameters can be estimated by:

\arjun{why does the batch start at 1, and the trajectory start at 0?} \amy{you can change this if you care to, but doing so might introduce bugs.}
\begin{align}
    &\frac{1}{\numsamples T} \sum^{\numsamples}_{m=1} \sum^{\trajlength - 1}_{t=0} \left( \valfun[][{\policy[\actorparams]}] (\state[t][\sample]) - \valfun[\criticparams] (\state[t][\sample]) \right)
    \left( \frac{1}{T-t} \sum^{\trajlength - 1 -t}_{k=0} \grad[\actorparams] \log \policy[\actorparams] (\action[t+k][\sample] \mid \state[t+k][\sample]) \, \Qfunc[][{\policy[\actorparams]}] (\state[t+k][\sample], \action[t+k][\sample]) \right).
\end{align}
\end{corollary}

%% file: appendix/nystrom_details.tex
\section{The Nystr\"om Method}
\label{appx:nystrom}

Proposed by \citet{hataya2023Nystrom}, the Nystr\"om method produces a low-rank approximation of the \IHVP{}.
Assume a $\Hessdim$-dimensional positive (semi)definite Hessian denoted by $\Hessian$ and a low-rank approximation of $\Hessian$ denoted by $\Hessian[\rank]$, for $\rank \ll \Hessdim$.
By selecting a set of $\randi$ indices of size $\rank$ at random, we obtain the Nystr\"om approximation
%
    $\Hessian[\rank] = \Hessian[[:, \randi]] \Hessian[[\randi, \randi]][\dagger] \Hessian[[:, \randi]][\top]$,
where $\Hessian[[:, \randi]] \in \R^{\Hessdim \times \rank}$ is a matrix of select columns of $\Hessian$ at indices $\randi$;
$\Hessian[[\randi, \randi]] \in \R^{\rank \times \rank}$ is a matrix of select rows of $\Hessian[[:, \randi]]$ at indices $\randi$;
and $\Hessian[[\randi, \randi]][\dagger]= \bm{U}\bm{\Lambda}^{-1}\bm{U}^{\top}$ is the Moore-Penrose pseudoinverse of $\Hessian[[\randi, \randi]]$, where $\bm{U}$ are the eigenvectors and $\bm{\Lambda}$  are the eigenvalues of $\Hessian[[\randi, \randi]]$.

For some small regularization constant $\nystromconst > 0$, we define the $\nystromconst$-regularized \IHVP{} as $(\Hessian[\rank] + \nystromconst \I)^{-1}$, where $\I$ is the $p$-dimensional identity matrix.%
\footnote{Incorporating the small perturbation $\nystromconst \I$ into this \IHVP{} is equivalent to optimizing a proximal regularization of the inner objective: i.e., $\innerobj[][\outer] (\inner) + \nicefrac{\nystromconst}{2}\Vert \inner - \arg \min_\inner \innerobj[][\outer] (\inner) \Vert$ \cite{lorraine2020optimizing}.}
This $\nystromconst$-regularized \IHVP{} decomposes as follows:
    $(\Hessian[\rank] + \nystromconst \I)^{-1} = (\Hessian[[:, \randi]] \Hessian[[\randi, \randi]][\dagger] \Hessian[[:, \randi]][\top] + \nystromconst \I) ^{-1}$.
Next, applying the Woodbury matrix identity,
    $(A + CBC^{\top})^{-1}=A^{-1} - A^{-1}C(B^{-1}+ C^{\top}A^{-1}C)^{-1}C^{\top}A^{-1}$, 
%
and setting $A = \nystromconst \I$, $B = \Hessian[[\randi, \randi]][\dagger]$, and $C = \Hessian[[:, \randi]]$ yields:
\begin{align*}
    (\Hessian[\rank] + \nystromconst \I)^{-1}  = \frac{1}{\nystromconst} \I - \frac{1}{\nystromconst^2}\Hessian[[:, \randi]] \bigg( \Hessian[[\randi:\randi]] + \frac{1}{\nystromconst} \Hessian[[:, \randi]][\top] \Hessian[[:, \randi]] \bigg)^{-1} \Hessian[[:, \randi]][\top].
\end{align*}
This final expression allows us to approximate the \IHVP{} $\hvp$ by $\esthvp = (\Hessian[\rank] + \nystromconst \I)^{-1} \grad[\inner] \outerobj (\outer, \innerest)$.

\input{sections/algo}
The regularization parameter $\nystromconst$ plays a crucial role in balancing numerical stability and the fidelity of curvature information. 
As discussed by \citet{vicol_implicit_2022}, for an eigenvalue $\lambda$ of the Hessian $H$, the effect of regularization is to modify the inverse eigenvalue as $(\lambda + \nystromconst)^{-1}$. 
When $\lambda \gg \nystromconst$, this term behaves as $\lambda^{-1}$, preserving curvature information. 
Conversely, when $\nystromconst \gg \lambda$, the approximation becomes insensitive to low-curvature directions, effectively ignoring them.  
In practice, $\nystromconst$ is selected empirically. 
One practical advantage of the Nystr\"om method is its ability to operate effectively with smaller $\nystromconst$.
For example, we use $\nystromconst = 50$, compared to $\nystromconst = 500$ in \citet{zheng2022stackelberg} for continuous control RL, and $\nystromconst = 10,000$ in \citet{fiez_implicit_2020} for a GAN-based bilevel problem, both of which use CG.

Notably, \citet{lorraine2020optimizing} observed that using the identity matrix, which is equivalent to choosing $\nystromconst \gg \firstlip$ and then rescaling the IHVP, matched the performance of approximate IHVPs computed via CG.
We find that the Nystr\"om method offers an attractive balance, where regularization remains modest without sacrificing stability.


%% file: appendix/pseudocode.tex
\clearpage
\section{Pseudocode}
\label{appx:psudocode}
We provide the pseudo code of the relevant baseline algorithms in this section.

\input{sections/algo}
A general actor critic framework \cref{alg:a2c}, and PPO \cref{alg:ppo}.

\begin{algorithm}[hb]
\caption{Actor Critic \cite{sutton2018reinforcement}} 
\label{alg:a2c}
\begin{algorithmic}
    \STATE Initialize actor network $\policy[\actorparams]$ and critic network $\Vfunc[\criticparams]$, parametrized respectively by $\actorparams[0]$, $\criticparams[0]$.
    \STATE \textbf{Input:} Hyperparameters and environment
    \FOR{$k = 0, 1, 2, \ldots$}
    \STATE Collect environmental trajectories $\Traj[k]$ with the current policy $\policy[\actorparams[k]]$.
    \STATE Compute the advantage $\adv[k]$ bootstrapped with current value function $\Vfunc[\criticparams[k]]$ using any method of advantage estimation.
    
        \STATE Update the actor by maximizing the policy gradient objective: \\ $\actorparams[k+1] = \actorparams[k] + \grad_{\actorparams} \frac{1}{T}\ \sum_{\timestep=0}^{\trajlength} \log \policy[\actorparams] (\action[t] | \state[t]) \adv[k](\state[t], \action[t])$
        \STATE Update the critic by minimizing the gap between bootstrapped targets and current value estimation for one step: \\ 
        $\criticparams[k+1] = \criticparams[k] -  \grad[\criticparams] \frac{1}{T}\sum_{\timestep=0}^{\trajlength}(\Qfunc[k](\state[t], \action[t]) - \Vfunc[\criticparams])^2$,
        where $\Qfunc[k](\state[t], \action[t]) = \Vfunc[\criticparams[k]](\state[t]) + \adv[k] (\state[t], \action[t])$
    \ENDFOR
\end{algorithmic}
\end{algorithm}

\begin{algorithm}[h]
\caption{PPO \cite{SpinningUp2018}}
\label{alg:ppo}
\begin{algorithmic}
    \STATE Initialize actor network $\policy[\actorparams]$ and critic network $\Vfunc[\criticparams]$, parametrized respectively by $\actorparams[0]$, $\criticparams[0]$.
    \STATE \textbf{Input:} Hyperparameters and environment
    \FOR{$k = 0, 1, 2, \ldots$}
    \STATE Collect environmental trajectories $\Traj[k]$ with the current policy $\policy[\actorparams[k]]$.
    \STATE Compute the normalized advantage $\adv[k]$ bootstrapped with current value function $\Vfunc[\criticparams[k]]$ using the Generalized Advantage Estimator 
    \FOR{all $n$ epochs}
    \FOR{all $m$ mini-batches created by collecting $\Traj[k]$ and $\adv[k]$}
        \STATE Update the actor by maximizing the $\epsilon$-clipped objective for one step: \\ $\actorparams[k+1] = \actorparams[k] + \grad[\actorparams] \frac{1}{|\Traj[k]|T}\sum_{\traj \in \Traj[k]} \sum_{\timestep=0}^{\trajlength}\text{ clip} \left( \frac{\policy[\actorparams] (\action[t] | \state[t])}{\policy[\actorparams[k]](\action[t] | \state[t])}, \epsilon \right)\adv[k](\state[t], \action[t])$
        \STATE Update the critic by minimizing the gap between GAE targets and current value estimation for one step: \\ 
        $\criticparams[k+1] = \criticparams[k] -  \grad[\criticparams] \frac{1}{|\Traj[k]|T}\sum_{\traj \in \Traj[k]} \sum_{\timestep=0}^{\trajlength}((\Vfunc[\criticparams[k]](\state[t]) + \adv[k] (\state[t], \action[t])) - \Vfunc[\criticparams])^2$
    \ENDFOR
    \ENDFOR
    \ENDFOR
\end{algorithmic}
\end{algorithm}


%% file: appendix/toy.tex
\section{Toy Examples}
\label{appd:toy}

\subsection{Toy Problem}
We adopt and extend the toy problem from \cite{zheng2022stackelberg}. In this case, we consider a single-step MDP where where $\actorparams \in [-1,1]$ is the decision variable. The reward is given by $\reward = \frac{-1}{5} \actorparams[][2]$ with . The critic is assumed to be a linear function approximation $\Vfunc[\criticparams](\actorparams) = \criticparams\actorparams$ with $\criticparams \in [-1,1]$. As in standard actor-critic, the actor seeks to maximize rewards, while the critic seeks to approximate the rewards generated by the actor. Therefore, the actor objective is $\actorobj(\actorparams, \criticparams) = \Vfunc[\criticparams](\actorparams) = \criticparams \actorparams$ and the critic objective is $\criticobj(\actorparams, \criticparams) =  \Ex_{\actorparams}[\reward(\actorparams) - \Vfunc[\criticparams](\actorparams)]^2$. Assuming the critic minimizes the mean-squared error of the sample action generated by the current actor, the critic objective becomes $\criticobj(\actorparams, \criticparams) =  (\reward(\actorparams) - \Vfunc[\criticparams](\actorparams))^2 = (\criticparams \cdot \actorparams + \frac{1}{5} \actorparams[][2])^2$. In our toy experiment we set the base learning rates to be $5e-2$, four times faster for the critic in TTS, and a regularizer of 0.3 for the regularized Stackelberg dynamics. 

\subsection{Conjugate Gradient vs Nystr\"om}
\label{appx:mlp}
We build small 2-layer MLP with ReLU activations. We randomly generate 100 neural networks: the batch size is sampled from $\{8,16,32,64\}$, the input dimension is sampled from $\{32,64, 128\}$, hidden layer size sampled from $\{8,16,32\}\}$ and output dimension sampled from $\{4,8,16\}$. The data $\X \sim \calN(0,\I)$ and the labels $\Y \sim 0.5 \calN(0,I)$. For calculating the regularized IHVP, we set $\nystromconst=0.01$, and sample a probe vector $u \sim \calN(0,\I)$. We run CG for 30 iterations and all computations are done to 64-bit precision on an AMD EPYC 7B12 cpu. 